\documentclass[twoside]{article}

\usepackage[accepted]{aistats2023}

\usepackage[round]{natbib}

\usepackage{amsthm, amsmath, amssymb}
\usepackage{mathtools}

\allowdisplaybreaks

\usepackage[utf8]{inputenc} 
\usepackage[T1]{fontenc}    

\usepackage{xparse}					
\usepackage{lipsum}					
\usepackage{xcolor}
\usepackage{comment}

\usepackage{siunitx}				
\usepackage[shortcuts]{extdash} 
\usepackage{anyfontsize}		
\usepackage{flushend}				

\usepackage{amsfonts}       
\usepackage{nicefrac}       
\usepackage{amsmath}				
\usepackage{mathtools}			
\usepackage{amssymb}				
\usepackage{bm}							
\usepackage{bbm}						

\usepackage{hyperref}

\usepackage{cleveref}
\crefname{equation}{Eqn.}{Eqns.}
\Crefname{equation}{Eqn.}{Eqns.}
\crefname{table}{Tab.}{Tabs.}
\Crefname{table}{Tab.}{Tabs.}
\crefname{figure}{Fig.}{Figs.}
\Crefname{figure}{Fig.}{Figs.}
\crefname{algorithm}{Alg.}{Algs.}
\Crefname{algorithm}{Alg.}{Algs.}
\crefname{proposition}{Prop.}{Proposition}
\Crefname{proposition}{Props.}{Propositions}
\crefname{section}{Section}{Sections}
\Crefname{section}{Section}{Sections}
\crefname{appendix}{Appendix}{Appendixes}
\Crefname{appendix}{Appendix}{Appendixes}
\crefname{assumption}{Assumption}{Assumption} 
\crefname{theorem}{Theorem}{Theorems} 
\crefname{definition}{Def.}{Definitions}
\crefname{corollary}{Corollary}{Corollary}
\crefformat{pluralequation}{#2Eqns.\,(#1)#3}
\Crefformat{pluralequation}{#2Eqns.\,(#1)#3}
\crefrangeformat{pluralequation}{Eqns.\,(#3#1#4) to~(#5#2#6)}
\Crefrangeformat{pluralequation}{Eqns.\,(#3#1#4) to~(#5#2#6)}
\crefmultiformat{pluralequation}{Eqns.\,(#2#1#3)}{ and~(#2#1#3)}{, (#2#1#3)}{ and~(#2#1#3)}
\Crefmultiformat{pluralequation}{Eqns.\,(#2#1#3)}{ and~(#2#1#3)}{, (#2#1#3)}{ and~(#2#1#3)}
\usepackage{xr}							

\usepackage{tikz}						
\usetikzlibrary{decorations.pathreplacing}
\usetikzlibrary{decorations.markings}
\usetikzlibrary{decorations.text}
\usetikzlibrary{patterns}
\usetikzlibrary{positioning}
\usetikzlibrary{shapes,arrows}
\usetikzlibrary{arrows.meta}
\usetikzlibrary{calc}
\usetikzlibrary{intersections}
\usetikzlibrary{plotmarks}
\usetikzlibrary{spy,backgrounds}
\usepackage{pgfplots}				
\usepgflibrary{plotmarks}
\usepgfplotslibrary{units}
\usepgfplotslibrary{fillbetween}
\pgfplotsset{compat=1.15}
\usetikzlibrary{pgfplots.groupplots}
\usetikzlibrary{external}
\usepackage{adjustbox}			
\usepackage{wrapfig}

\usepackage{booktabs}				
\usepackage{multirow}				
\usepackage{makecell}				

\usepackage{enumitem}				

\usepackage{caption}
\usepackage{placeins}  

\usepackage{url}						
\makeatletter
\let\UrlSpecialsOld\UrlSpecials
\def\UrlSpecials{\UrlSpecialsOld\do\/{\Url@slash}\do\_{\Url@underscore}}%
\def\Url@slash{\@ifnextchar/{\kern-.11em\mathchar47\kern-.2em}%
    {\kern-.0em\mathchar47\kern-.08em\penalty\UrlBigBreakPenalty}}
\def\Url@underscore{\nfss@text{\leavevmode \kern.06em\vbox{\hrule\@width.3em}}}
\makeatother								
\usepackage{algorithm2e}
\usepackage{listings} 

\definecolor{white}{rgb}{1.0,1.0,1.0}
\definecolor{brightred}{rgb}{1.0,0.1,0.1}
\definecolor{brightblue}{rgb}{0.0,0.0,0.8}
\definecolor{darkblue}{rgb}{0.0,0.0,0.5}
\definecolor{darkgreen}{rgb}{0.0,0.3,0.0}
\definecolor{brightgreen}{rgb}{0.0,0.8,0.0}
\definecolor{darkblack}{rgb}{0.0,0.0,0.0}
\definecolor{grey}{rgb}{0.3,0.3,0.3}

\long\def\comment#1{}  

\newcommand{\disT}{\textstyle}

\newcommand{\undersets}[2]{\underset{\hbox to 0pt{\hss{\scriptsize #1}\hss}}{#2}}
\newcommand{\oversets}[2]{\overset{\hbox to 0pt{\hss{\scriptsize #1}\hss}}{#2}}

\newcommand{\FF}{{\mathcal F}}

\newcommand{\HH}{{\mathcal H}}

\newcommand{\LL}{{\mathcal L}}

\newcommand{\NN}{{\mathcal N}}

\newcommand{\RRR}{\mathbb{R}} 
\newcommand{\Scal}{{\mathcal S}}  
\newcommand{\TT}{{\mathcal T}}
\newcommand{\TTt}{\tilde{\TT}}

\newcommand{\EEE}[2]{\mathbb{E}_{#1}\big\{#2\big\}}

\newcommand{\diag}{\mathrm{diag}}

\def\Vhrulefill{\leavevmode\leaders\hrule height 0.7ex depth \dimexpr0.4pt-0.7ex\hfill\kern0pt}

\def\One{\mathbb{I}}

\def\Amat{A}

\def\WVecZh{\tilde{\mathbf{W}}^{0}_h}
\def\WVecZOne{\tilde{\mathbf{W}}^{0}_1}
\def\WVecZH{\tilde{\mathbf{W}}^{0}_H}

\def\given{\vert} 

\newcommand{\muVec}{\boldsymbol{\mu}}

\newcommand{\muT}{\tilde{\boldsymbol{\mu}}}

\newcommand{\nuVec}{\boldsymbol{\nu}}
\newcommand{\nuVect}{\tilde{\boldsymbol{\nu}}}

\newcommand{\qn}{q^{(n)}}

\newcommand{\classk}{k}

\newcommand{\subc}{c}

\newcommand{\xVec}{\mathbf{x}}
\newcommand{\zVec}{\mathbf{z}}
\newcommand{\xVecN}{\mathbf{x}^{\hspace{0.5pt}(n)}}

\newcommand{\pT}{p_{\Theta}}
\newcommand{\qPhi}{q_{\Phi}}

\newcommand{\qPhiEnc}{\qPhi^{\mathrm{enc}}}
\newcommand{\pTPrior}{\pT^{\mathrm{prior}}}
\newcommand{\pTDec}{\pT^{\mathrm{dec}}}

\NewDocumentCommand{\y}{O{}O{}}{y_{#2}^{\,#1}{}}
\NewDocumentCommand{\yVec}{O{}}{\vec{y}_{\vphantom{c}}^{\,#1}{}}

\NewDocumentCommand{\Wgen}{O{}O{}}{\mathcal{W}_{#1#2}}

\NewDocumentCommand{\Rgen}{O{}O{}}{\mathcal{R}_{#1#2}}

\NewDocumentCommand{\W}{O{}O{}}{W_{#1#2}}

\NewDocumentCommand{\R}{O{}O{}}{R_{#1#2}}

\NewDocumentCommand{\Sc}{O{\subc}O{}}{s_{#1}^{#2}}

\NewDocumentCommand{\Igenc}{O{\subc}O{}}{\mathcal{I}_{#1}^{#2}}

\NewDocumentCommand{\Ic}{O{\subc}O{}}{I_{#1}^{#2}}

\NewDocumentCommand{\tk}{O{\classk}O{}}{t_{#1}^{#2}}

\NewDocumentCommand{\tVec}{O{}}{\vec{t}_{\vphantom{\classk}}^{\,#1}}

\NewDocumentCommand{\eps}{O{}}{\epsilon_{\textnormal{\tiny $#1$}}}

\NewDocumentCommand{\epst}{O{}}{\tilde{\epsilon}_{\textnormal{\tiny $#1$}}}

\renewcommand{\d}{\mathrm{d}}

\newcommand{\DKL}[2]{D_{\mathrm{KL}}\big(#1 \, \Vert \, #2\big)}

\newcommand{\Wall}{W}
\newcommand{\Mall}{M}

\newcommand{\sVec}{\mathbf{s}}

\newcommand{\muVect}{\tilde{\boldsymbol{\mu}}}
\newcommand{\zVect}{\tilde{\mathbf{z}}}
\newcommand{\zVecT}{\tilde{\mathbf{z}}}

\newcommand{\etaVec}{\boldsymbol{\eta}}
\newcommand{\etaVecT}{\boldsymbol{\eta}^{\mathrm{T}}}
\newcommand{\TVec}{\mathbf{T}}

\newcommand{\bVec}{\mathbf{b}}
\newcommand{\cVec}{\mathbf{c}}

\DeclareMathOperator{\Tr}{Tr}

\newcommand{\sigmaT}{\tilde{\sigma}}
\newcommand{\sigT}{\tilde{\sigma}}

\newcommand{\sigTVec}{\tilde{\boldsymbol{\sigma}}}

\newcommand{\alphaT}{\tilde{\alpha}}

\newcommand{\xiVec}{\boldsymbol{\xi}}
\newcommand{\Acal}{{\cal A}}

\newcommand{\JJT}{{\cal J}^{\mathrm{T}}}
\newcommand{\IIT}{{\cal I}^{\mathrm{T}}}

\newcommand{\vVec}{\boldsymbol{v}}
\newcommand{\gVec}{\boldsymbol{g}}
\newcommand{\gVecA}{\boldsymbol{g}^{\mathrm{A}}}
\newcommand{\gVecB}{\boldsymbol{g}^{\mathrm{B}}}
\newcommand{\gA}{g^{\mathrm{A}}}
\newcommand{\gB}{g^{\mathrm{B}}}
\newcommand{\thetaVec}{\boldsymbol{\theta}}

\newcommand{\del}[1]{\frac{\partial}{\partial{}#1}}

\newcommand{\Del}[2]{\frac{\partial{}\,#1}{\partial{}#2}}

\newtheorem{theorem}{Theorem}[]
\newtheorem{corollary}{Corollary}[]

\newtheorem{definition}{Definition}
\newtheorem{assumption}{Assumption}

\makeatletter
\def\blfootnote{\gdef\@thefnmark{}\@footnotetext}
\makeatother

\makeatletter
\hypersetup{pdftitle={The ELBO of Variational Autoencoders
Converges to a Sum of Entropies}, pdfauthor={Simon Damm, Dennis Forster, Dmytro Velychko, Zhenwen Dai, Asja Fischer, Jörg Lücke}}
\makeatother

\begin{document}

\runningauthor{Damm, Forster, Velychko, Dai, Fischer, Lücke}

\twocolumn[

\aistatstitle{The ELBO of Variational Autoencoders
Converges to a Sum of Entropies}

\vspace{-1ex}
\aistatsauthor{ Simon Damm$^{*}$ \And Dennis Forster$^{\dag}$ \And  Dmytro Velychko }
\aistatsaddress{ \normalsize Ruhr University Bochum, Germany  \And  \normalsize Frankfurt University of \\ Applied Sciences, Germany \And University of Oldenburg, Germany}
\vspace{-1ex}

\aistatsauthor{ Zhenwen Dai \And Asja Fischer \And Jörg Lücke$^{*}$ }
\aistatsaddress{ 
    \normalsize Spotify, London, UK \And Ruhr University Bochum, Germany \vspace{-0ex}
    \And University of Oldenburg, Germany %
    } ]
\begin{abstract}
The central objective function of a variational autoencoder (VAE) is its variational lower bound (the ELBO). Here we show that for standard (i.e., Gaussian) VAEs the ELBO converges to a value given by the sum of three entropies: the (negative) entropy of the prior distribution, the expected (negative) entropy of the observable distribution, and the average entropy of the variational distributions (the latter is already part of the ELBO). Our derived analytical results are exact and apply for small as well as for intricate deep networks for encoder and decoder. Furthermore, they apply for finitely and infinitely many data points and at any stationary point (including local maxima and saddle points). The result implies that the ELBO can for standard VAEs often be computed in closed-form at stationary points while the original ELBO requires numerical approximations of integrals. As a main contribution, we provide the proof that the ELBO for VAEs is at stationary points equal to entropy sums. 
Numerical experiments then show that the obtained analytical results are sufficiently precise also in those vicinities of stationary points that are reached in practice. 
Furthermore, we discuss how the novel entropy form of the ELBO can be used to analyze and understand learning behavior. More generally, we believe that our contributions can be useful for future theoretical and practical studies on VAE learning as they provide novel information on those points in parameters space that optimization of VAEs converges to.
\end{abstract}

\section{INTRODUCTION}
\blfootnote{$^{*}$joint main contributions}
\blfootnote{$^{\dag}$Large parts of the research were conducted while the author was affiliated with the University of Oldenburg, Germany.}
Variational autoencoders \citep[VAEs;][]{KingmaWelling2014,RezendeEtAl2014} have emerged as a popular choice for probabilistic generative modeling, a sub-field of unsupervised (deep) learning.
VAEs are in their most common form defined as latent variable models %
that learn %
representations of the data ${\xVec \sim p(\xVec)}$, with $\xVec$ $\in$ $\RRR^D$, in a usually low(er) dimensional latent space $\zVec$ $\in$ $\RRR^H$.
In contrast to conventional autoencoders, the mappings between data and latent space are stochastic. 
The (by far) most common choice for prior distribution $p(\zVec)$,
encoder $\qPhi(\zVec \given \xVec)$ and
decoder $\pT(\xVec \given \zVec)$ are Gaussian distributions, where deep neural networks (DNNs) are used to define Gaussian means and (optionally) covariances.
Ideally, maximum likelihood estimation of model parameters~$\Theta$ would be deployed with %
\begin{equation}
    \label{eq:Likelihood}
    p_\Theta(\xVec) = \int p(\zVec) \pT(\xVec \given \zVec) \d \zVec
\end{equation}
approximating the target data distribution $p(\xVec)$. However, since the likelihood in \cref{eq:Likelihood} 
is not tractable for complex decoder distributions, the \emph{evidence lower bound} (ELBO) is optimized instead. 
The ELBO is also known as {\em variational lower bound} or {\em variational free energy} \citep[][]{NealHinton1998}.

Given a set of $N$ data points $\xVecN$, $n \in \{1,\dots,N\}$, the ELBO in dependence of encoder and decoder parameters (i.e., $\Phi$ and $\Theta$, respectively) is given by%
\begin{align}
    \FF(\Phi,\Theta) = \,
    & \overbrace{   \frac{1}{N}\sum_{n} \int q_{\Phi}\!(\zVec\, \given\, \xVecN) \log\!\big( \pT(\xVecN\, \given\, \zVec) \big) \,\mathrm{d}\zVec}^{S_\mathrm{rec}(\Phi,\Theta)} \notag\\
    & \hspace{3mm}
    \underbrace{ - \, \frac{1}{N}\sum_{n} \DKL{q_{\Phi}\!(\zVec\, \given\, \xVecN)}{p(\zVec)}}_{S_\mathrm{reg}(\Phi,\Theta)} \enspace,%
\label{EqnELBO}
\end{align}
where $\DKL{q(\zVec)}{p(\zVec)}$ denotes the Kullback-Leibler divergence between two distributions $q$ and $p$.
The first ELBO term, $S_\mathrm{rec}(\Phi,\Theta)$, is usually referred to as reconstruction score and the second $S_\mathrm{reg}(\Phi,\Theta)$ as regularization score \citep{kingma2019introduction}.
For all encoder parameters $\Phi$ and decoder parameters $\Theta$ the ELBO is smaller or equal to the log-likelihood $\LL(\Theta)=\frac{1}{N}\sum_n \log\big(\pT(\xVecN)\big)$.

While the ELBO has proven to be an exceptionally successful learning objective, it is, like the log-likelihood itself, usually not analytically tractable for VAEs (and neither for many other models).
The intractability of the bound for VAEs stems from 
$S_\mathrm{rec}(\Phi,\Theta)$ in \cref{EqnELBO}: 
Because of (potentially very intricate) DNN non-linearities of standard VAEs, the integrals cannot be solved analytically. A central research challenge for training VAEs is therefore the development of efficient
methods to approximate intractable integrals of the ELBO. Indeed, the suggestion of efficient methods to estimate gradients of 
\cref{EqnELBO} using sampling and reparametrization \citep[][]{KingmaWelling2014,RezendeEtAl2014} has played the key role in the establishment of the field of VAE research.

Because of the potentially complex DNNs deployed in VAEs few exact theoretical results may be expected especially in realistic settings, i.e., for real and finite data sets and convergence to local optima or saddle points. Therefore maybe unexpectedly, we here provide an exact analytical result for all standard VAEs. Concretely, we show that 
at convergence the ELBO is given by a sum of the entropies of those distributions defining a VAE.
In order to make our contribution
more precise let us properly define a vanilla Gaussian VAE, which we here term \textbf{VAE--1}.
\begin{definition}[VAE--1; VAE with component-wise equivalent decoder variances] 
\label{def:VAE-1} 
Consider a VAE with standard normal prior\vspace{-1ex}
\begin{align*}
p(\zVec) &= \NN(\zVec; \mathbf{0}, \One)  \enspace,
\intertext{and Gaussian encoder and decoder given, respectively, by}
\qPhi(\zVec \given \xVec) &= \NN\big(\zVec; \nuVec_{\Phi}(\xVec), \TT_{\Phi}(\xVec)\big)  \enspace,  \\
\pT(\xVec \given \zVec) &= \NN(\xVec; \muVec_{\Theta}(\zVec),\sigma^2\One) \enspace.
\end{align*}
Let the encoder covariance  $\TT_{\Phi}(\xVec)$  
be diagonal and let encoder mean $\nuVec_{\Phi}(\xVec)$, covariance  $\TT_{\Phi}(\xVec)$,
and decoder mean $\muVec_{\Theta}(\zVec)$
be parameterized by DNNs, i.e.,
\begin{align}
    \begin{split}
        \nuVec_{\Phi}(\xVec) &= \mathrm{DNN}_{\nu}(\xVec; V) \enspace,\\
        \TT_{\Phi}(\xVec) &= \diag\left(\tau_1^2(\xVec), \dots, \tau_H^2(\xVec)\right)\enspace,\\
        \muVec_{\Theta}(\zVec) &=  \mathrm{DNN}_{\mu}(\zVec; W)
    \end{split}
\label{EqnParametrization}
\end{align}
with $\left(\tau_1^2(\xVec), \dots, \tau_H^2(\xVec)\right)^\mathrm{T} = \mathrm{DNN}_{\tau}(\xVec; T)$.
By $\Theta=(W,\sigma^2)$ we denote the set of all parameters of the decoder, and by $\Phi=(V,T)$ all parameters of the encoder. 
We hereby assume that the respective parameter sets $W,V$, and $T$ include all weight matrices and biases of the DNNs.
\end{definition}

The results we will derive are exploiting properties of the ELBO (\cref{EqnELBO}) at stationary points,
i.e., of those points in parameter space where the ELBO
reaches an extremum (local or global optima) or a saddle point. 
As a consequence, the presented results are applicable, e.g., for
gradient-based optimization techniques.
Throughout the paper, we refer with `at convergence' to the stationary points but we remark that VAE parameters will in practice only
reach the vicinity of these points (due to finite learning rates and stochasticity; we elaborate on this in \cref{app:Experiments}).

For VAE--1 as defined above, the final result is sufficiently concise to be stated here initially (before we discuss its derivation, related work, and more general VAEs later on):
At all stationary points of the ELBO
(\cref{EqnELBO})
we have
\begin{align}
\begin{split}
    \FF(\Phi,\Theta) &=
    \frac{1}{N}\sum_{n=1}^N \HH[\qPhi(\zVec\vert\xVecN)] \\
    &\quad- \HH[p(\zVec)] -\HH[\pT(\xVec\,|\,\zVec)] 
\end{split}
\label{EqnThreeEntropiesIntro}
\intertext{
where $\HH[p(\cdot)]$ denotes the entropy of a distribution $p(\cdot)$. 
That is, the ELBO is, at convergence, given by the entropies of encoder, prior, and decoder distribution. 
Using the closed-form expressions for Gaussian entropies, the ELBO is consequently closed-form and solely depends on %
the variance parameters of encoder and decoder:
}
\begin{split}
  \FF(\Phi,\Theta)
    &=%
    \frac{1}{2N}\sum_{n=1}^N\sum_{h=1}^{H}\log\!\big( 2 \pi e \tau_h^2(\xVecN; \Phi)  \big) 
    \\ &\quad-\,\frac{H}{2}\log(2\pi e)  -\,\frac{D}{2}\log(2\pi e \sigma^2) \enspace .
\end{split}
\label{EqnThreeEntropiesIntroExp}
\end{align}
No other parameters are required to compute the bound, and in particular no knowledge about (or passes through) the decoder DNN is needed. 
Considering \cref{EqnThreeEntropiesIntroExp}, we also remark that the expression does not contain any approximations of any integrals. 
Using just the data and the learned parameters, the ELBO can at convergence be computed exactly.
We stress that the closed-form expression (\cref{EqnThreeEntropiesIntroExp}) does {\em not} replace the original bound as a learning objective (i.e., it can not be used analogously to \cref{EqnELBO} for parameter optimization).
Importantly, however, \cref{EqnThreeEntropiesIntro} (and later
discussed generalizations) implies that during learning the ELBO converges to a sum of three entropies.

\section{RELATED WORK}
In order to understand and improve learning algorithms, points in parameter space representing (potentially locally) optimal solutions are of high interest.
For VAEs there are several lines of work in this respect which investigate 
ELBO optimization \citep{hoffman2016elbo,mescheder2017adversarial,DaiEtAl2018,LucasEtAl2019,shekhovtsov2022vae}.
Work by \citet{DaiEtAl2018} and \citet{LucasEtAl2019}, for instance, highlight the connections of linear (Gaussian) VAEs to (robust) principal component analysis \citep[PCA;][]{TippingBishop1999,XuEtAl2010,candes2011robust,HoltzmanEtAl2020} and
the crucial role of covariances of encoder and decoder, respectively, that help to circumvent undesirable maxima in the optimization landscape of VAEs. 
\citet{DaiEtAl2018} derived meaningful insights on optima of the  
ELBO based on tractable special cases. 
In addition, \cite{LucasEtAl2019} link spurious local maxima to posterior collapse and derive a measure to quantify this phenomenon.
\cite{shekhovtsov2022vae} study the approximation gap between likelihood and ELBO for exponential family VAEs, and demonstrate that
ELBO-based learning is subject to an inductive bias towards the (rather restricted) consistent set.

The here reported results, in contrast, closely relate the ELBO objective of VAEs to entropies.
The relation between the integrals of the ELBO and entropies has been of interest previously \citep{LuckeHenniges2012}.
However, the main results of that previous work have used (A)~the limit case of infinitely many data points, (B)~assumed a perfect match of variational and
full posterior distributions, and (C)~required convergence to global optima.
The work also discussed relaxations of these relatively unrealistic assumptions \citep[][Sec.\,6]{LuckeHenniges2012}.
But those relaxations made use of properties of sparse coding like models trained using expectation maximization.
While being related, the previous results are, therefore, not applicable to the training of VAEs. Furthermore, work in parallel to this study 
considers elementary generative models with distributions in the exponential family 
\citep[][]{lucke2022convergence} but no deep models such as VAEs are treated.
We will later make use of that work for more complex VAEs
in which DNNs also parameterize decoder variances. 
Results of \citep[][]{lucke2022convergence} in principle also allow
treatments of still less conventional VAEs, e.g., VAEs
defined using distributions such as Gamma, Bernoulli, continuous Bernoulli, Beta or Categorical distributions. However, for each distribution or combination of distributions, a parameterization condition has to be verified analytically. \Cref{sec:appProofPropThree} shows the verification for the most complex Gaussian VAEs treated here. In this context the Appendix also provides a brief discussion of the conditions for non-Gaussian distributions. 
While the result of \cref{EqnThreeEntropiesIntroExp} is generalizable, 
VAEs with fixed decoder variances are straight-forward examples for
VAEs not converging to entropy sums. But we remark that learning $\sigma^2$ is usually beneficial \citep{LucasEtAl2019,rybkin21a}.

Finally, a related but different line of research discusses how {\em other} learning objectives for deep models can be defined that are closed-form by definition. Examples are contributions by \citet{KingmaDhariwal2018} or \citet[][]{OordEtAl2016}. In contrast, we here investigate the standard (in general intractable) learning objective of VAEs and show its analytical tractability at convergence.

\section{THE ELBO AT CONVERGENCE}
\label{SecVarBounds}

For the derivation of our main results
we require a more explicit, yet standard form of the decoder DNN:

\begin{assumption}
\label{assumption:linear_mapping}
    The network $\mu_\Theta(\zVec) = \mathrm{DNN}_\mu(\zVec, W)$
    is a %
    composition
    of %
    linear mappings followed by point-wise
    non-linear functions. Concretely, $\muVec_\Theta(\zVec)$ is given by 
    \begin{align*}
    W^{L} \Scal\big( W^{L-1} \Scal( \cdots 
    \Scal(W^{0}\zVec + \bVec^{0}) 
    \cdots) + \bVec^{L-1}\big) + \bVec^L,
    \end{align*}
    where $W^{l}$ denotes the weight matrix of layer $l$ and $\bVec^l$ denotes its bias terms. The point-wise non-linearities $\Scal(\cdot)$ can (but do not have to) differ from layer to layer. 
\end{assumption}

Throughout this paper we presume \cref{assumption:linear_mapping} to be fulfilled.
We do not assume any specific architecture for the encoder networks $\mathrm{DNN}_\nu$ and $\mathrm{DNN}_\tau$.

\subsection{A Reparameterized VAE}
\label{SecReparameterization}
Before we derive the result for VAE--1 presented in \cref{EqnThreeEntropiesIntro}, let us consider a different VAE
which can be regarded as a reparametrization of VAE--1 
and which we refer to as \textbf{VAE--2}.
\begin{definition}[VAE--2, VAE with component-wise equivalent decoder variances \textit{and} learnable prior covariance]
\label{def:VAE-2} 
Consider a VAE with a parameterized prior
\begin{align*}
\pT(\zVect) &= \NN(\zVect; \mathbf{0}, \Amat) \enspace, \\ 
\intertext{and Gaussian encoder and decoder given by}
\qPhi(\zVect \given \xVec) &= \NN\big(\zVect; \nuVect_{\Phi}(\xVec), \TTt_{\Phi}(\xVec)\big) \enspace,  \\
\pT(\xVec\,|\,\zVect) &= \NN(\xVec; \muVect_{\Theta}(\zVect),\sigma^2\One) \enspace
\end{align*}
where $\nuVect_{\Phi}(\xVec)$ and $\TTt_{\Phi}(\xVec)$ are parametrized analogously to VAE--1 (\cref{def:VAE-1}). In contrast to VAE--1, the prior's covariance matrix
is now given by the diagonal matrix ${\Amat = \diag\left(\alpha_1^2, \dots, \alpha_H^2\right)}$. Furthermore, we assume for the decoder DNN, $\muVect_{\Theta}(\zVect)$, that the columns of the weight matrix $\tilde{W}^{0}$ 
are of unit length, i.e, for ${\tilde{W}^{0} = (\WVecZOne,\ldots,\WVecZH)}$ we demand  %
\begin{equation}
    \forall h \in \{1, \dots, H\}:
    \big(\WVecZh\big)^{\mathrm{T}}\WVecZh = 1\,. \enspace \label{EqnWConstraint}
\end{equation}
We refer to such a VAE as VAE--2.
\end{definition}
We will later see that VAE--2 can indeed parametrize the same distributions as VAE--1.
The advantage of \mbox{VAE--2} compared to VAE--1 is that it is of a form for which 
the variance parameters of its prior distribution can be learned (which will be exploited below).

The stationary points are %
those points in parameter space for which the derivatives w.r.t.\ all parameters (parameters $\Phi$ and $\Theta$) vanish.
In particular, this implies for VAE--2: %
\begin{equation}
\frac{\d}{\d \sigma^2} \FF(\Phi,\Theta) = 0  \quad \text{and} \quad \frac{\d}{\d \alpha^2_h} \FF(\Phi,\Theta) = 0\ \forall h \enspace. \label{EqnStatPoints}
\end{equation}

The derivatives w.r.t.\ the DNN parameters $W,V$, and $T$ are also zero at stationary points
(for the parameters $\tilde{W}^{0}$ a derivative with Lagrange multipliers is zero). However, we will 
only use \cref{EqnStatPoints} in the sequel.
We can now proceed to proof the following theorem:

\begin{theorem}
\label{theo:VAE-2}
Given a VAE--2 as in \cref{def:VAE-2} that satisfies \cref{assumption:linear_mapping}.
At all stationary points the %
ELBO $\FF(\Phi,\Theta)$
of VAE--2 is then equal to
\begin{align}
&\frac{1}{N}\sum_{n=1}^N \HH[\qPhi(\zVect\given\xVecN)] - \HH[\pT(\zVect)] -\HH[\pT(\xVec\,|\,\zVect)]    \nonumber \\
\begin{split}
     =\ &\frac{1}{N}\sum_{n=1}^N \frac{1}{2} \log\!\big(\det( 2\pi e \TTt_{\Phi}(\xVecN) \big) \\ 
    &- \frac{1}{2} \log\!\big(\det(2\pi e \Amat     ) \big) 
    - \frac{D}{2} \log\!\big( 2\pi e \sigma^2 \big).
\end{split}
\label{EqnPropOne}
\end{align}
\end{theorem}

\begin{proof}
We can rewrite the standard formulation of the %
ELBO (\cref{EqnELBO}) to consist of three terms:
\begin{align}
\FF(\Phi,\Theta) &= \FF_1(\Phi,\Theta) + \FF_2(\Phi,\Theta) + \FF_3(\Phi)\,, \  \text{with}\nonumber\\
\FF_1(\Phi,\Theta) &= \frac{1}{N}\sum_{n} \int \qn_{\Phi}\!(\zVec) \log\!\big( \pT(\zVec) \big)\hspace{0.5ex} \mathrm{d}\zVec \enspace,\nonumber\\
\FF_2(\Phi,\Theta) &= \frac{1}{N}\sum_{n} \int \qn_{\Phi}\!(\zVec) \log\!\big( \pT(\xVecN\,|\,\zVec) \big) \mathrm{d}\zVec \enspace,\nonumber\\
\FF_3(\Phi) &= -\, \frac{1}{N}\sum_{n} \int \qn_{\Phi}\!(\zVec) \log\!\big( \qn_{\Phi}\!(\zVec) \big) \mathrm{d}\zVec \enspace,\nonumber
\end{align}
where we dropped the `tilde' for $\zVec$ in the proof.

First consider $\FF_2(\Phi,\Theta)$ and observe that the logarithm of the prefactor in $\pT(\xVec\,|\,\zVec)$ evaluates to $-\frac{D}{2}\log(2\pi\sigma^2)$,\linebreak thus resembles the entropy of the Gaussian distribution (up to a constant factor). Hence, we can re-express $\FF_2(\Phi,\Theta)$ as follows:
\begin{align}
&\frac{1}{N}\sum_n \Bigg( - \frac{1}{2\sigma^2} \int \qn_{\Phi}\!(\zVec)\, \| \xVecN - \muVec_{\Theta}(\zVec) \|^2 \mathrm{d}\zVec \nonumber \\ 
 &\quad - \frac{D}{2} \log\!\big( 2\pi\sigma^2 \big) - \frac{D}{2} + \frac{D}{2} \Bigg) \nonumber\\
    =&\, \frac{D}{2} \left( 1 - \frac{1}{N D\sigma^2} \sum_n \int \qn_{\Phi}\!(\zVec)\, \| \xVecN - \muVec_{\Theta}(\zVec) \|^2 \mathrm{d}\zVec \right) \nonumber\\
    &\quad - \frac{D}{2} \log( 2\pi e \sigma^2  )  \enspace  ,
\label{ProofA}
\end{align}
where the last term is now the negative entropy of a Gaussian (where `$e$' is Euler's number).

At stationary points it applies that \mbox{$\frac{\mathrm{d}}{\mathrm{d}\sigma^2}\FF(\Phi,\Theta)=0$}. Only $\FF_2(\Phi,\Theta)$ depends on $\sigma^2$, which implies \mbox{$\frac{\mathrm{d}}{\mathrm{d}\sigma^2}\FF_2(\Phi,\Theta)=0$}. In virtue of \cref{ProofA}, the derivative has a specific structure given by:
\begin{align}
    0 &= \frac{\mathrm{d}}{\mathrm{d}\sigma^2}\FF_2(\Phi,\Theta) \nonumber\\
    \,&=\, \frac{D}{2} \big(\frac{1}{N\! D\sigma^4} \sum_n\hspace{-0.5ex}\int\hspace{-0.5ex} \qn_{\Phi}\!(\zVec) \| \xVecN - \muVec_{\Theta}(\zVec) \|^2 \mathrm{d}\zVec\big) - \frac{D}{2\sigma^2} \nonumber\\
    &= - \frac{D}{2\sigma^2} \big( 1 - \frac{1}{N\! D\sigma^2} \sum_n\hspace{-0.5ex}\int\hspace{-0.5ex} \qn_{\Phi}\!(\zVec) \| \xVecN - \muVec_{\Theta}(\zVec) \|^2 \mathrm{d}\zVec  \big). \nonumber%
\intertext{As $\frac{D}{2\sigma^2}$ is greater zero, it follows that:}
    & 1 - \frac{1}{N\! D\sigma^2}  \sum_n\hspace{-0.5ex}\int\hspace{-0.5ex} \qn_{\Phi}\!(\zVec) \| \xVecN - \muVec_{\Theta}(\zVec) \|^2 \mathrm{d}\zVec = 0 \enspace. \nonumber
\end{align} 
We recognize the integral to be the first term of \cref{ProofA}, i.e., the term not depending on the Gaussian entropy.
We can thus conclude that at stationary points of $\FF(\Phi,\Theta)$ it applies that
\begin{equation}
\FF_2(\Phi,\Theta) =  
- \frac{D}{2} \log\!\big( 2\pi e \sigma^2  \big) = -\HH[\pT(\xVec\,|\,\zVec)] \enspace.
\end{equation}
Next we consider the term $\FF_1(\Phi,\Theta)$ of the ELBO.
Analogous to $\FF_2(\Phi,\Theta)$, we observe that
the logarithm of the prefactor is similar to the entropy of a Gaussian (this time with diagonal covariance)
and we rewrite $\FF_1(\Phi,\Theta)$ as
\begin{align}
\begin{split}
    &%
    \frac{1}{N}\sum_n \! \Big(\!\! - \!\!\sum_h \frac{1}{2\alpha_h^2} \int \!\! \qn_{\Phi}\!(\zVec) z_h^{2} \mathrm{d}\zVec %
    - \frac{1}{2} \sum_h \log\!\big( 2\pi\alpha_h^2 \big)\!\Big) =
    \\ &\frac{1}{2} \sum_h \! \Big(1\!-\!\frac{1}{N\! \alpha_h^2} \sum_n \! \int \!\! \qn_{\Phi}\!(\zVec) z_h^{2} \mathrm{d}\zVec  \Big) - \frac{1}{2} \sum_h\log\!\big( 2\pi e \alpha_h^2 \big)\! \enspace,
\end{split}
\label{ProofBB}
\end{align}
\vspace{-3.2ex}\\
where the last term is the negative entropy of the prior. At stationary points \cref{EqnStatPoints} applies, and we obtain:
\begin{align*}
0 &= \frac{\mathrm{d}}{\mathrm{d}\alpha_h^2}\FF_1(\Phi,\Theta) \\
    &=\, \frac{1}{2}\! \sum_{h'} \frac{\delta_{hh'}}{N\!\alpha_h^4} \sum_n \! \int \!\!  \qn_{\Phi}\!(\zVec)  z_{h'}^{2} \mathrm{d}\zVec   - \frac{1}{2}\! \sum_{h'}\! \frac{\delta_{hh'}}{\alpha_h^2} \nonumber\\
	&= - \frac{1}{2\alpha_h^2} \Big(1 - \frac{1}{N\!\alpha_h^2}  \sum_n \! \int \!\! \qn_{\Phi}\!(\zVec)  z_{h}^{2} \mathrm{d}\zVec  \Big) \enspace.
\end{align*}
As $\frac{1}{2\alpha_h^2}$ is greater zero, it follows that for each $h$:
\begin{equation*}
    1 - \frac{1}{N\!\alpha_h^2} \sum_n \int \qn_{\Phi}\!(\zVec)  (z_{h})^{2} \mathrm{d}\zVec = 0 \enspace.
\end{equation*}
The first sum over $h$ in \cref{ProofBB} is consequently zero, and we obtain at convergence:
\begin{equation}
\FF_1(\Phi,\Theta) = - \frac{1}{2} \log\!\big(  \det(2\pi e \Amat) \big) = -\HH[\pT(\zVec)].
\end{equation}
The term $\FF_3(\Phi)$ is directly given as the average entropy of the variational distribution. Taken together, we thus obtain the claim. %
\end{proof} 
\subsection{Convergence to Sums of Entropies}
We can now provide a result for the ELBO of the standard VAE given by \cref{def:VAE-1}
(VAE--1) by translating the result of \cref{theo:VAE-2}
for VAE--2 back to the original parameterization.

\begin{theorem}
\label{theo:VAE-1}
    Given a VAE--1 as in \cref{def:VAE-1} that
    satisfies \cref{assumption:linear_mapping}.
    Then at all stationary points the ELBO, $\FF(\Phi,\Theta)$, of VAE--1 is 
    equal to 
\begin{align}
&\frac{1}{N}\sum_{n=1}^N \HH[\qPhi(\zVec\given\xVecN)] - \HH[\pT(\zVec)] -\HH[\pT(\xVec\,|\,\zVec)] \label{PropTwoInEntropies}\\
&= \frac{1}{2N}\sum_{n=1}^N\sum_{h=1}^H \log\!\big( \tau^2_{h}(\xVecN;\Phi) \big) - \frac{D}{2} \log\!\big( 2\pi e \sigma^2 \big). \label{PropTwoA}
\end{align}
\end{theorem}

\begin{proof}\vspace{-2.5ex}
Let us start by showing that VAE--1 and VAE--2 indeed parametrize the same distributions.
Following \cref{assumption:linear_mapping} the initial mapping of $\mathrm{DNN}_\mu$ is linear with weight matrix $W^{0}$ for VAE--1 or $\tilde{W}^{0}$ for VAE--2.
With $\tilde{W}^{0} = \big(\WVecZOne, \ldots, \WVecZH\big)$ and $\Amat$ as in \cref{def:VAE-2}
we can now set
\begin{equation}
W^{0} =\ \tilde{W}^{0}\Amat^{\frac{1}{2}} = \big(\alpha_1 \WVecZOne, \ldots, \alpha_H\WVecZH\big) \enspace.\label{EqnWAsWTilde}
\end{equation}
Note that the column vectors of $\tilde{W}^{0}$ are constrained to unit length (see \cref{def:VAE-2}).
Thus, $\big(\tilde{W}^{0}\Amat^{\frac{1}{2}}\big)$ parameterize the same space of matrices as $W^{0}$. The first linear operation of %
$\mathrm{DNN}_\mu$ now becomes:
\begin{equation}
W^{0}\zVec + \bVec^{0} = \sum_h \WVecZh \alpha_h z_h + \bVec^{0} \enspace. \label{EqnProofzz}
\end{equation}
Considering the term $\alpha_h z_h$, we can now generate ${\tilde{z}_h \sim \NN(\tilde{z}_h;0,\alpha^2_h)}$ instead of $z_h \sim \NN(z_h;0,1)$.
We recognize that VAE--1 in this way takes on the form of \mbox{VAE--2}.
Hence, when the parameters of \mbox{VAE--1} represent a stationary point, the parameters with $W^{0}$ replaced by $\Amat$ and $\tilde{W}^{0}$ also
represent a stationary point. 

We thus conclude that \cref{theo:VAE-2} applies.
The decoder variance $\sigma^2$ remains unchanged and $\Amat$ could be obtained from the column vectors of $W^{0}$. 
However, it is left to express $\tilde{\TT}_{\Phi}(\xVec)$ in terms of $\TT_{\Phi}(\xVec)$. Let us drop subscript and argument of $\tilde{\TT}$ for readability.
Then $\tilde{\TT}$ is the covariance matrix of a Gaussian distribution defined in the space of $\zVect$.
In virtue of \cref{EqnProofzz} the random variable $\zVec$ is given by $\zVec=\Amat^{-\frac{1}{2}}\zVect$.
Consequently, if $\zVect$ is Gaussian distributed with covariance $\tilde{\TT}$, then $\zVec$ is Gaussian distributed with
covariance $\TT=\Amat^{-\frac{1}{2}} \tilde{\TT} \big(\Amat^{-\frac{1}{2}}\big)^\mathrm{T}$. As all matrices are diagonal, we get $\tilde{\TT}=\Amat\TT$.
Inserting into \cref{EqnPropOne} we observe the first term to cancel with part of the last term:
\begin{align}
\FF(\Phi,\Theta) \,&=\, - \frac{1}{2} \log\!\big(\det(2\pi e \Amat ) \big) - \frac{D}{2} \log\!\big( 2\pi e \sigma^2 \big) \nonumber \\
&\qquad+ \frac{1}{N}\sum_{n} \frac{1}{2} \log\!\big(\det( 2\pi e \Amat\TT_{\Phi}(\xVecN) \big) \nonumber \\
\begin{split}
    &\,=\, - \frac{H}{2} \log( 2\pi e ) - \frac{D}{2} \log\!\big( 2\pi e \sigma^2 \big) \\
    &\qquad+ \frac{1}{N}\sum_{n} \frac{1}{2} \log\!\big(\det( 2\pi e \TT_{\Phi}(\xVecN) \big) 
\end{split}
\label{EqnTermOrder}\\
\,=\,- &\frac{D}{2} \log\!\big( 2\pi e \sigma^2 \big) + \frac{1}{2N}\!\sum_{n} \log\!\big(\!\det( \TT_{\Phi}(\xVecN) \big). \hspace{-3.5pt} \nonumber
\end{align}

The middle equation we recognize as the sum of three entropies in \cref{PropTwoInEntropies}, which proofs the claim.
The last equation explicitly expresses the entropies (after further simplification) using the variance parameters of VAE--1. 
For \cref{PropTwoInEntropies} we moved the last term in \cref{EqnTermOrder} to the front to match the order of terms to the order of processing in VAEs.
\end{proof}

There are a number of implications and remarks if considering \cref{theo:VAE-1}:
As already pointed out in the introduction, no approximations of any integrals (nor any other approximations) are required.
The computation of the bound is consequently very straight-forward and efficient in practice.
More importantly, however, is the observation that learning of VAEs (as given by \mbox{VAE--1}) necessarily converges to sums of entropies. 
As a consequence, the value of the bound only depends on a subset of VAE parameters at convergence: the variances of encoder and decoder.
In particular, the entropies (and therefore the ELBO value at convergence) do not depend on the DNNs for Gaussian means.
\paragraph{Linear VAEs}
Standard VAEs as studied above exhibit complex learning behavior such that theoretical insights are notoriously difficult to obtain.
To better understand salient challenges of VAE training such as mode collapse, %
a natural approach is to first try to gain insights using as elementary as possible models.
For VAEs, the most elementary such model is presumably represented by a linear VAE
\citep[also compare][]{RumelhartEtAl1985,BaldiHornik1989,DaiEtAl2018,KuninEtAl2019,LucasEtAl2019},
i.e., a VAE with decoder and encoder given, respectively, by:
\begin{align}
\pT(\zVec)& = \NN(\zVec; \mathbf{0}, \One),\   
\pT(\xVec\,|\,\zVec) = \NN(\xVec; W\zVec+\muVec_0,\sigma^2\One), \  \label{EqnVAElinearDecoder}  \\[-1.5ex]
\qn_{\Phi}\!(\zVec) &=\ \NN(\zVec; V(\xVecN-\muVec_0), \TT),\phantom{i} 
\label{EqnVAElinearEncoder}
\end{align}
 with weight matrices $W$ and $V$, and covariance ${\TT = \diag\left(\tau_1^2, \dots, \tau_H^2\right)}$.  
The linear VAE is a special case of VAE--1, with linear DNNs instead of the usual deep non-linear versions.
Therefore, we can conclude the following:

\begin{corollary}
\label{corollary:LinearVAE}
Consider the linear VAE defined by \cref{EqnVAElinearDecoder,EqnVAElinearEncoder}.
At all stationary points of its 
ELBO (\cref{EqnELBO}) it applies that:
\begin{equation}
\FF(\Phi,\Theta) = \frac{1}{2}\sum_{h} \log\!\big( \tau^2_h \big) \,-\, \frac{D}{2} \log\!\big( 2\pi e \sigma^2 \big).\label{EqnVAEBoundLinear}
\end{equation}
\end{corollary}
\begin{proof}
For the proof of \cref{theo:VAE-1} we only required the reparametrization of the first linear mapping of the decoder DNN (cf. \cref{assumption:linear_mapping}).
\cref{theo:VAE-1} thus also applies for the linear VAE as a special case of VAE--1 (we elaborate in \cref{app:LinearVAEs}). 
Inserting the matrix $\TT$ of \cref{EqnVAElinearEncoder} into \cref{PropTwoA} 
proves the claim
as $\TT$ is independent of $n$.
\end{proof}

\cref{corollary:LinearVAE} further highlights that the variance parameters determine the bound at convergence. 
As it is known that linear VAEs can recover the exact maximum likelihood \citep[][]{DaiEtAl2018,LucasEtAl2019}, we can even conclude that
the bound given in \cref{EqnVAEBoundLinear} is tight at convergence. We use this result in \cref{SecNum} and elaborate in \cref{app:LinearVAEs}.

\subsection{More General Gaussian VAEs}
VAE--1 represents the presumably most common form of VAEs. However, generalizations which use a DNN to learn more complex decoder covariances %
alongside a DNN for decoder means represent a possible generalization \citep[][etc]{RezendeEtAl2014,DortaEtAl2018}.
To also include VAEs with DNNs for {\em decoder} variances, we have to extend our analysis as we, so far, considered covariance $\sigma^2\One$. As relatively straight-forward generalization,
we therefore consider the following VAE:

\begin{definition}[VAE--3; 
VAE with latent dependent diagonal decoder covariance]
\label{def:VAE-3} 
Consider a VAE with distributions
\begin{align*}
    p(\zVec) &= \NN(\zVec; \mathbf{0}, \One)  \enspace,  \\ 
    \qPhi(\zVec \given \xVec) &= \NN\big(\zVec; \nuVec_{\Phi}(\xVec), \TT_{\Phi}(\xVec)\big)  \enspace,\\
    \pT(\xVec \given \zVec) &= \NN(\xVec; \muVec_{\Theta}(\zVec), \Sigma_{\Theta}(\zVec)\big)\enspace,
\end{align*}
where $\nu_\Phi$ and $\TT_\Phi$ are defined and parametrized analogously to VAE--1 (\cref{def:VAE-1}). 
Also $\muVec_{\Theta}(\zVec)=\mathrm{DNN}_{\mu}(\zVec; \Wall)$ is defined as for VAE--1 (i.e., according to \cref{assumption:linear_mapping}).
However, in contrast to VAE--1,\linebreak the decoder covariance is now a diagonal matrix ${\Sigma_{\Theta}(\zVec) = \diag\left(\sigma^2_1(\zVec; \Theta), \dots, \sigma^2_D(\zVec; \Theta)\right)}$
with elements depending on the latent code $\zVec$, implemented as
\begin{align}
    \left({\sigma^2_1}(\zVec; \Theta), \dots, {\sigma^2_D}(\zVec;\Theta)\right)^\mathrm{T} = \mathrm{DNN}_{\sigma}(\zVec; \Mall) \enspace,
    \label{EqnVAEThreeDecoderSigmaDNN}
\end{align}
where $\mathrm{DNN}_{\sigma}(\zVec; \Mall)$ is a standard DNN of the form:
    \begin{align*}
    M^{L'} \Scal\big( M^{L'-1} \Scal( \cdots 
    \Scal(M^{0}\zVec + \cVec^{0}) 
    \cdots) + \cVec^{L'-1}\big) + \cVec^{L'},
    \end{align*}
    where $M^{l}$ denotes the weight matrix of layer $l$ and $\cVec^l$ denotes its bias terms. 
$\mathrm{DNN}_{\sigma}(\zVec; \Mall)$ is of the same form as $\mathrm{DNN}_{\mu}(\zVec; \Wall)$ but can have a different architecture and different non-linearities $\Scal(\cdot)$. Furthermore, we demand $\mathrm{DNN}_{\sigma}(\zVec; \Mall)$ to always output positive values to avoid singularities. Both decoder DNNs we require to have at least one hidden layer, and we require that they are parameterized by two different sets of parameters, $W$ and $M$, respectively. We refer to such a VAE as VAE--3. 
\end{definition}

Because of the $\zVec$-depending variances, it is obvious that \cref{theo:VAE-1} can not apply.
Furthermore, the proof of \cref{theo:VAE-1} explicitly used that $\sigma^2$ does not depend on $\zVec$, so the proof for $\zVec$-dependent variances can not be a straight-forward generalization.
It is still possible, however, to derive expressions for the bound in terms of entropies:

\begin{theorem}
\label{theo:VAE-3}
Consider a VAE--3 as in \cref{def:VAE-3}. %
At all stationary points the ELBO of VAE--3 is then given by
\begin{align}
    \begin{split}
        \FF(\Phi,\Theta) =\; &\frac{1}{N}\sum_{n=1}^N \HH[\qPhi(\zVec\given\xVecN)] - \HH[\pT(\zVec)]  \\
        &- \frac{1}{N} \sum_{n=1}^N \EEE{\qn_{\Phi}}{ \HH[\pT(\xVec\,|\,\zVec)] } 
    \end{split}
    \label{PropThree} \\
    =\; &\frac{1}{2N} \sum_{n=1}^N \sum_{h=1}^H \log\!\big(     \tau_h^2 (\xVecN; \Phi) \big) - \frac{D}{2} \log\!\big( 2 \pi e \big)\nonumber \\
    &- \frac{1}{2N} \sum_{n=1}^N \sum_{d=1}^D \EEE{\qn_{\Phi}}{\log\!\big( \sigma_d^2 (\zVec; \Theta) \big)}. \nonumber 
\end{align}
\end{theorem}

\begin{proof}[Proof Sketch]
The proof of \cref{theo:VAE-3} is considerably more intricate than those of \cref{theo:VAE-2} and \cref{theo:VAE-1}.
The main challenge is the integral over $\log(\pT(\xVecN\,|\,\zVec))$
(compare term $\FF_2(\Phi,\Theta)$ in the proof of \cref{theo:VAE-2}). A generalization
is, however, possible again by a reformulation of the integral in terms of the
entropy of $\pT(\xVecN\,|\,\zVec)$. We present the full proof in %
\cref{sec:appProofPropThree} which is itself based on parallel work \citep[][]{lucke2022convergence} that also 
considers non-Gaussian distributions for elementary (non-deep) generative models.
While the proof for VAE--3 shares with the proofs of \cref{theo:VAE-2,theo:VAE-1} the use of a reparameterized VAE and rewriting of ELBO terms using entropies, it
requires significantly more elaborate derivations (including details especially of the decoder DNN for the variances).
\end{proof}

Considering \cref{theo:VAE-3}, observe that the final result is concise
and its application to a given VAE is straight-forward (while the proof is long and technical).

Also observe that \cref{theo:VAE-3} is indeed a generalization of \cref{theo:VAE-1}: if we replace $\sigma^2_d(\zVec;\Theta)$ by
a scalar $\sigma^2$, then we drop back to \cref{PropTwoA}. 
As was the case for \cref{theo:VAE-1}, the result of \cref{theo:VAE-3} applies for commonly encountered conditions. For {\em idealized} conditions, convergence to sums of entropies as in \cref{theo:VAE-3} can be shown relatively easily \citep[see, e.g.,][]{LuckeHenniges2012}. However, idealized would in this context mean that four unrealistic conditions have to be fulfilled: (1)~the data have to be distributed according to the used generative model; (2)~the data set has to be infinitely large; (3)~the variational distributions
have to be equal to the posterior; and (4)~learning has to converge to a global optimum. In contrast, \cref{theo:VAE-3} states the convergence to sums of entropies for realistic conditions: for any (reasonable) finite or infinite data sets, for any stationary point, and for any variational distributions.

\begin{figure*}[ht]
\includegraphics[width=0.33\linewidth]{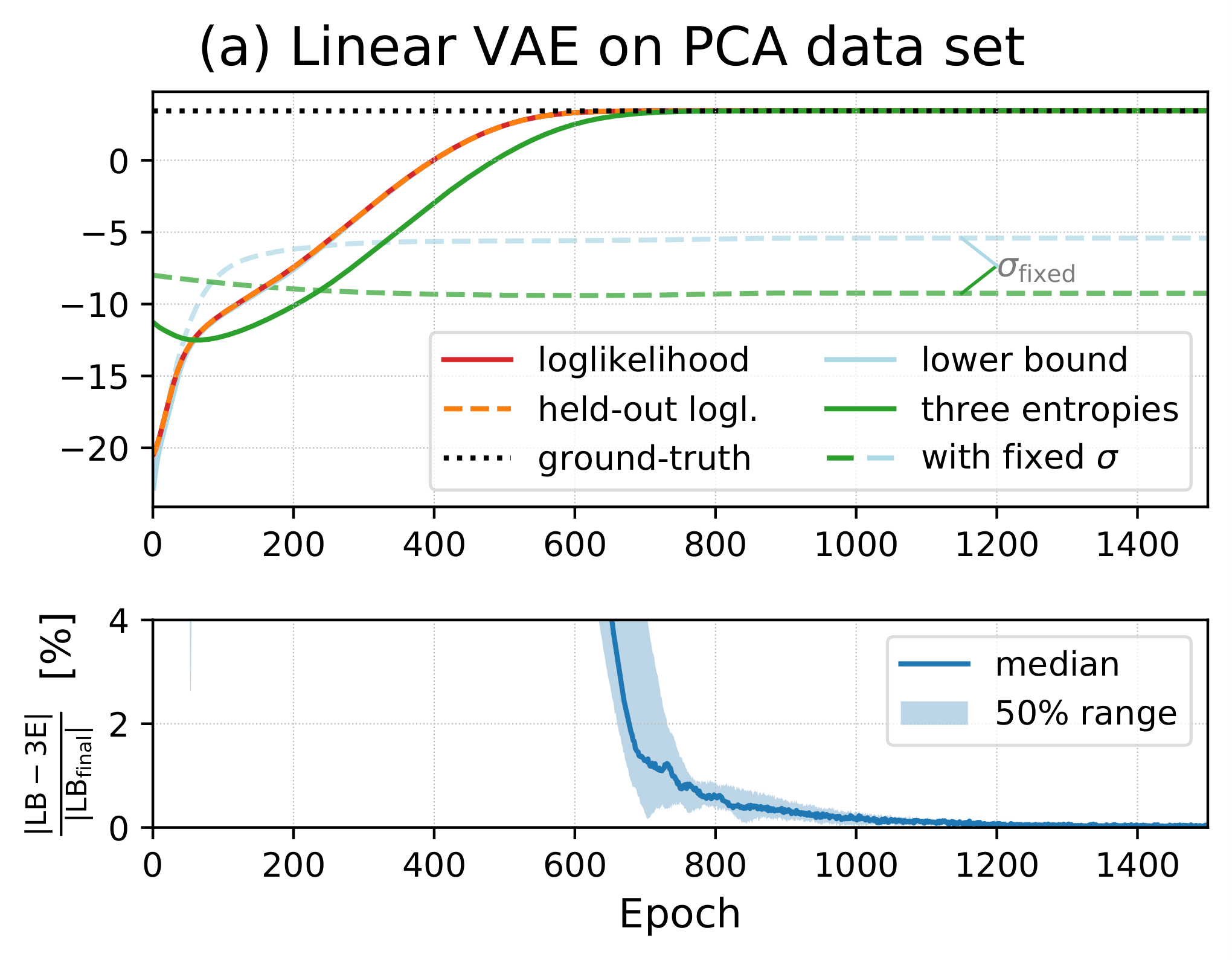}
\includegraphics[width=0.33\linewidth]{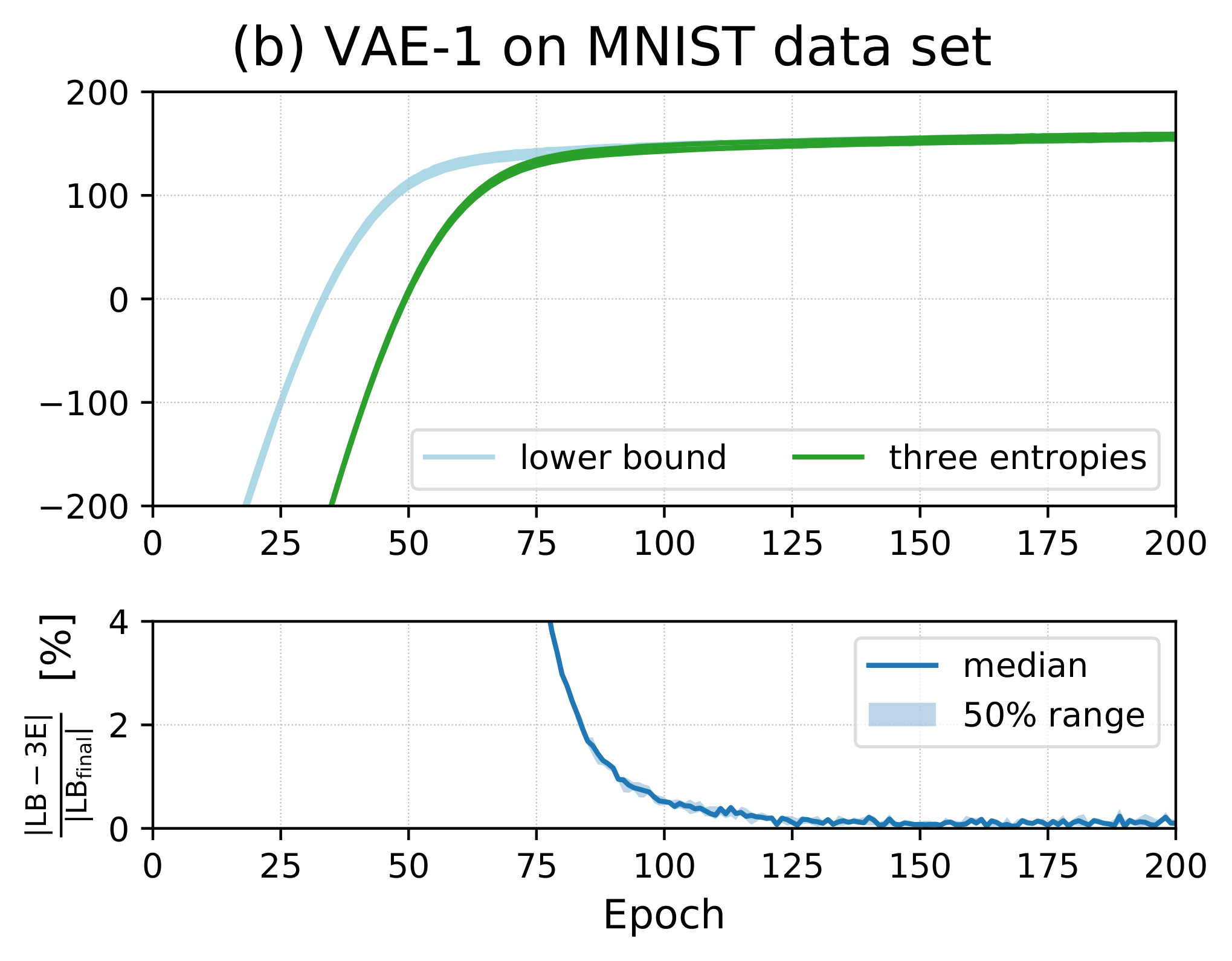}
\includegraphics[width=0.33\linewidth]{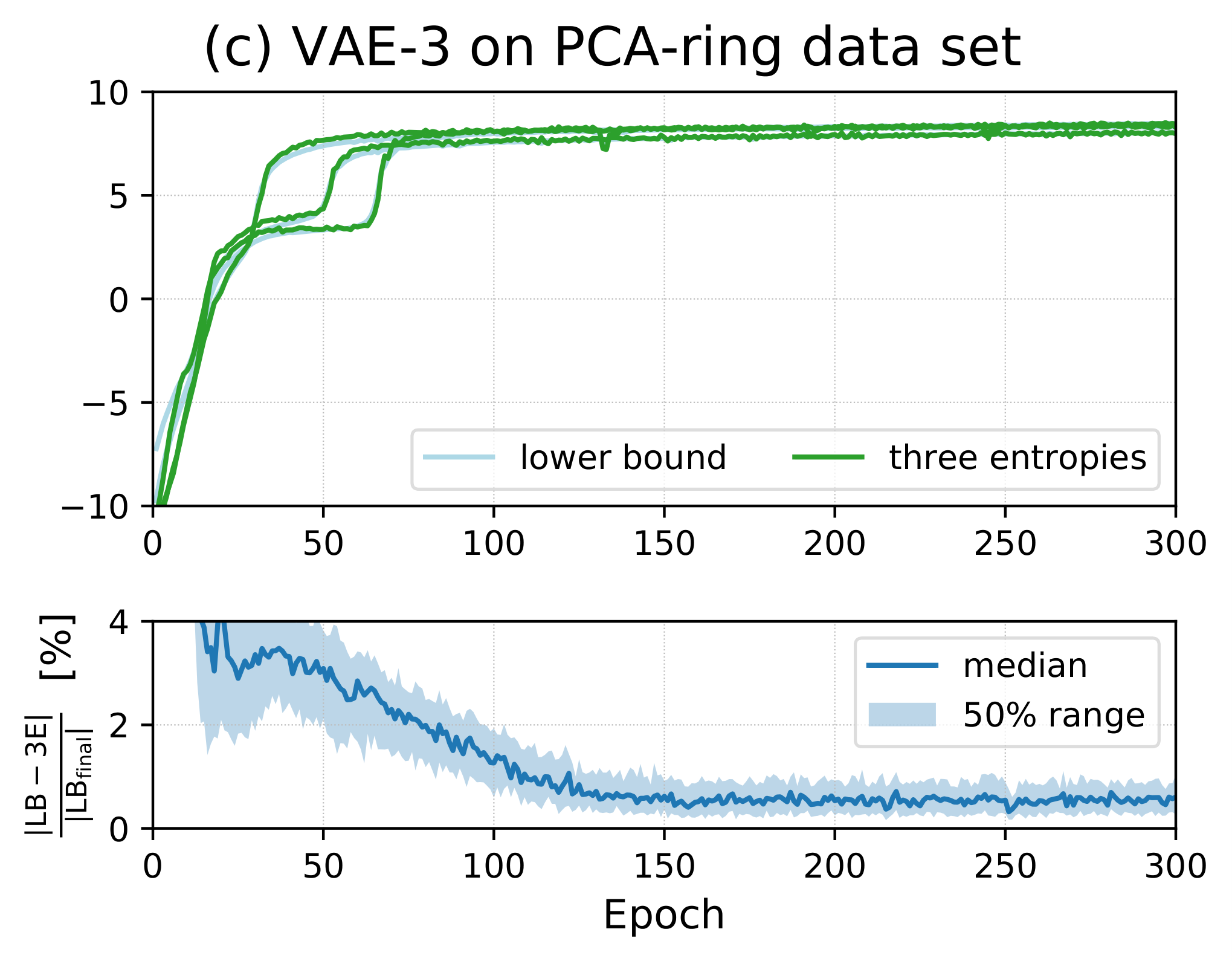}
\caption{\textbf{Verification of the entropy results} on VAE models of increasing complexity 
on different data sets.
{\textit{Top plots:} Absolute values of the given bounds per data point of single runs. 
In (a) the ELBO 
is essentially equal to the log-likelihood (zoom in to see).
In (c) both quantities are displayed for three different seeds.
\textit{Bottom plots:} Median and interquartile range of the relative difference between the ELBO and the sum of three entropies 
over multiple runs ($10$ for (a) and (b), $100$ for (c)). 
See \cref{fig:Verification_App} for further architectures and data sets, and \cref{app:Experiments} for details on the experiments.\vspace{-0ex}}}
\label{fig:Verification}
\vspace{-0pt}
\end{figure*}

\section{VERIFICATION AND ENTROPY-BASED ANALYSIS}
\label{SecNum}

The results of \cref{theo:VAE-1,theo:VAE-3} and \cref{corollary:LinearVAE} are the key theoretical contributions of this work. 
Still, we here study the results numerically\footnote{Code is available at \href{https://github.com/Learning-with-Entropies}{github.com/Learning-with-Entropies}.}, which will be instructive also about their potential practical relevance.
First, we investigate how well the results apply in those vicinities of stationary points that are reached when VAEs are optimized in practice.
Then, we discuss entropy-based perspectives on VAE learning: We investigate entropy-based forms of reconstruction and regularization terms and the analysis of posterior collapse with help of entropies. Moreover, we discuss improved ELBO estimation as well as fast model selection for linear VAEs based on the results presented in \cref{SecVarBounds}.

\paragraph{Verification}

\Cref{fig:Verification} (and \ref{fig:Verification_App} in \cref{app:Experiments}) show numerical experiments for linear VAEs, and for the non-linear
VAE--1 as well as VAE--3 applied to different data sets ranging from PCA data, over high-energy physics data \citep[SUSY,][]{baldi2014searching} to the image data sets MNIST \citep{LeCunEtAl1998} and CelebA \citep{liu2015faceattributes}.
We used VAEs with simple as well as relatively complex network architectures. 
Details about the experimental setup %
are given in \cref{app:Experiments}.
In all experiments, we observed a close correspondence of the original form of the
ELBO and the sum of entropies also in vicinities of stationary points that are reached in practice. 
Deviations between original ELBO values and three entropy expressions were essentially all due to stochasticity in ELBO computations (we elaborate in \cref{app:ExpNoise}).

\paragraph{Entropy-Based Analysis of VAE Learning}
The results presented in \cref{theo:VAE-1,theo:VAE-3} show that central figures in VAE optimization can be expressed solely based on entropies. For instance, reconstruction and regularization score (cf.\,\cref{EqnELBO}) are at stationary points given by
\begin{align}
\begin{split}
    S_{\mathrm{reg}}(\Phi,\Theta) &= \frac{1}{N}\sum_{n} \HH[\qPhi(\zVec\given\xVecN)] \,-\, \HH[\pT(\zVec)] \enspace, \\ %
    S_{\mathrm{rec}}(\Theta) &= -\,\HH[\pT(\xVecN\,|\,\zVec)] 
\end{split}
\label{eq:Scores_Reg_Rec}
\end{align}
where for VAE--3 the entropy $\HH[\pT(\xVecN\,|\,\zVec)]$ is replaced by the expected entropy (compare \cref{theo:VAE-3}). 
VAE optimization with the ELBO can therefore be re-interpreted by recalling that differential entropies characterize the volume of the typical set, the effective volume of a distribution \citep{thomas2006elements}: Maximizing the ELBO ultimately corresponds to minimizing the volume of the decoder's typical set, resembled by $\HH[\pT(\xVecN\,|\,\zVec)]$, \textit{and} the difference between the volume of the typical sets of prior and encoder distributions, captured in \mbox{$\HH[\pT(\zVec)] - \frac{1}{N}\sum_{n} \HH[\qPhi(\zVec\given\xVecN)]$} (see \cref{app:TypicalSet} for the full discussion).
In \cref{app:Entropy-BasedAnalysis} we also present an additional discussion of the optimization landscape based on the entropy results.
In the following we make use of the entropy expressions to show how the decoder entropy $\HH[\pT(\xVecN\,|\,\zVec)]$ enables improved ELBO estimations and model selection, and how the encoder entropy $\frac{1}{N}\sum_{n} \HH[\qPhi(\zVec\given\xVecN)]$ naturally provides an analysis tool for posterior collapse.

\paragraph{ELBO Estimation}
By using \cref{theo:VAE-1} it is possible to significantly simplify 
ELBO estimation for the wide-spread \mbox{VAE--1} (\cref{def:VAE-1}). 
To estimate the ELBO after convergence, we can by knowing \cref{PropTwoA} simply
use the values of the model parameters after training. Furthermore, merely the variance parameters are required.

No integrals have to be solved for ELBO estimation, while conventional estimation would require a numerical approximation of the integrals for the reconstruction term $S_{\mathrm{rec}}(\Theta)$ in \cref{EqnELBO}. 
Estimations based on the original ELBO, e.g., by using Monte-Carlo approximation of integrals, is of course possible and can be sufficiently precise. However, such estimations are stochastic,
require additional hyper-parameters (e.g., number of samples/data points or some smoothing parameter) and can be computationally costly. 
Hence, one way of interpreting the result is that the problem of solving the integral of \cref{EqnELBO} has already been solved by training the VAE, so it does not have to be solved again for the estimation of the ELBO value.\footnote{We remark that all our results apply for convergence based on the training data. The resulting ELBO values that can be estimated are instructive about the optimization process. The results do not directly transfer to validation or test sets (other data) unless the relevant parameters are retrained.}
Also regarding the second term, $S_{\mathrm{reg}}(\Phi,\Theta)$, which is already given in closed-form for the original ELBO, \cref{PropTwoA} provides a simplified analytical expression (and no analytical solutions of the integral are required).

\begin{figure}[h!]
\begin{center}
    \includegraphics[width=0.5\linewidth]{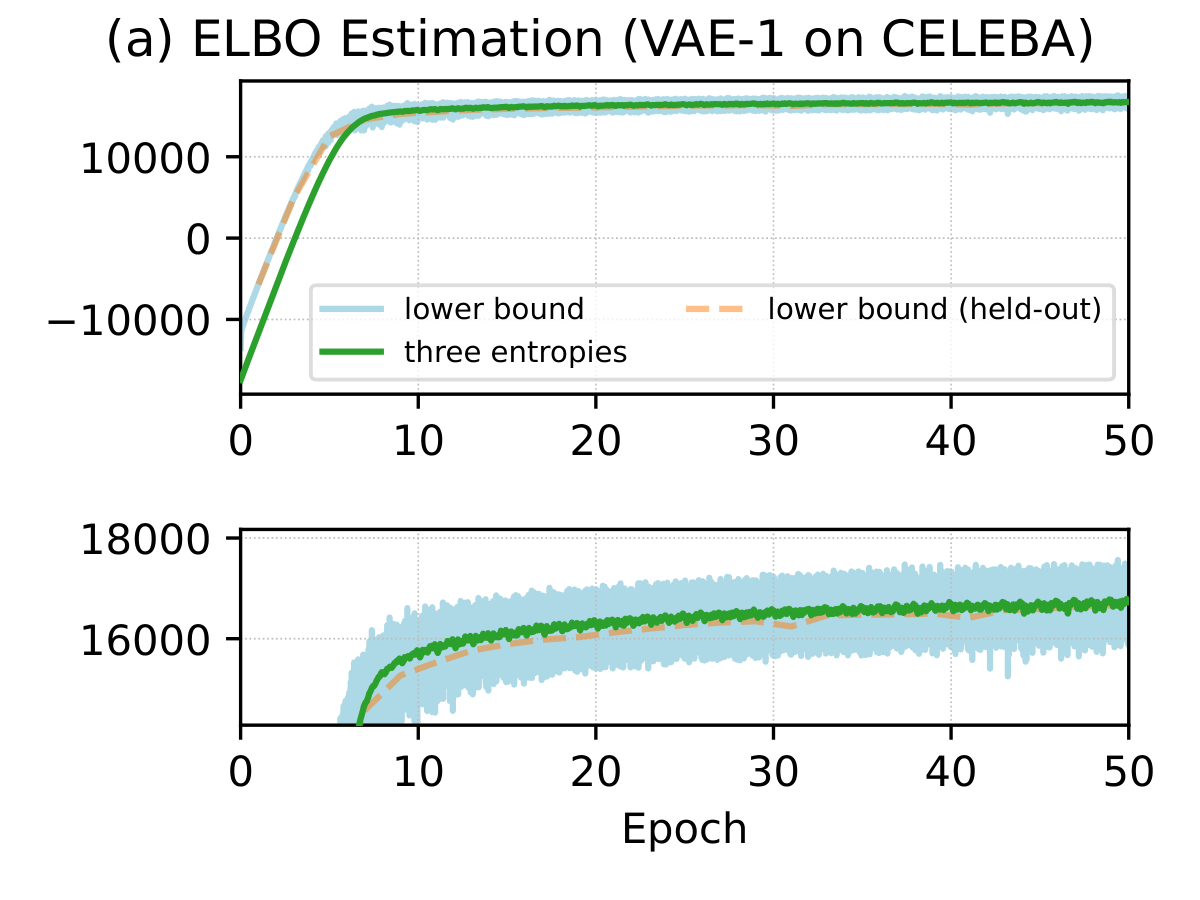}
    \hspace{-3mm} 
    \includegraphics[width=0.5\linewidth]{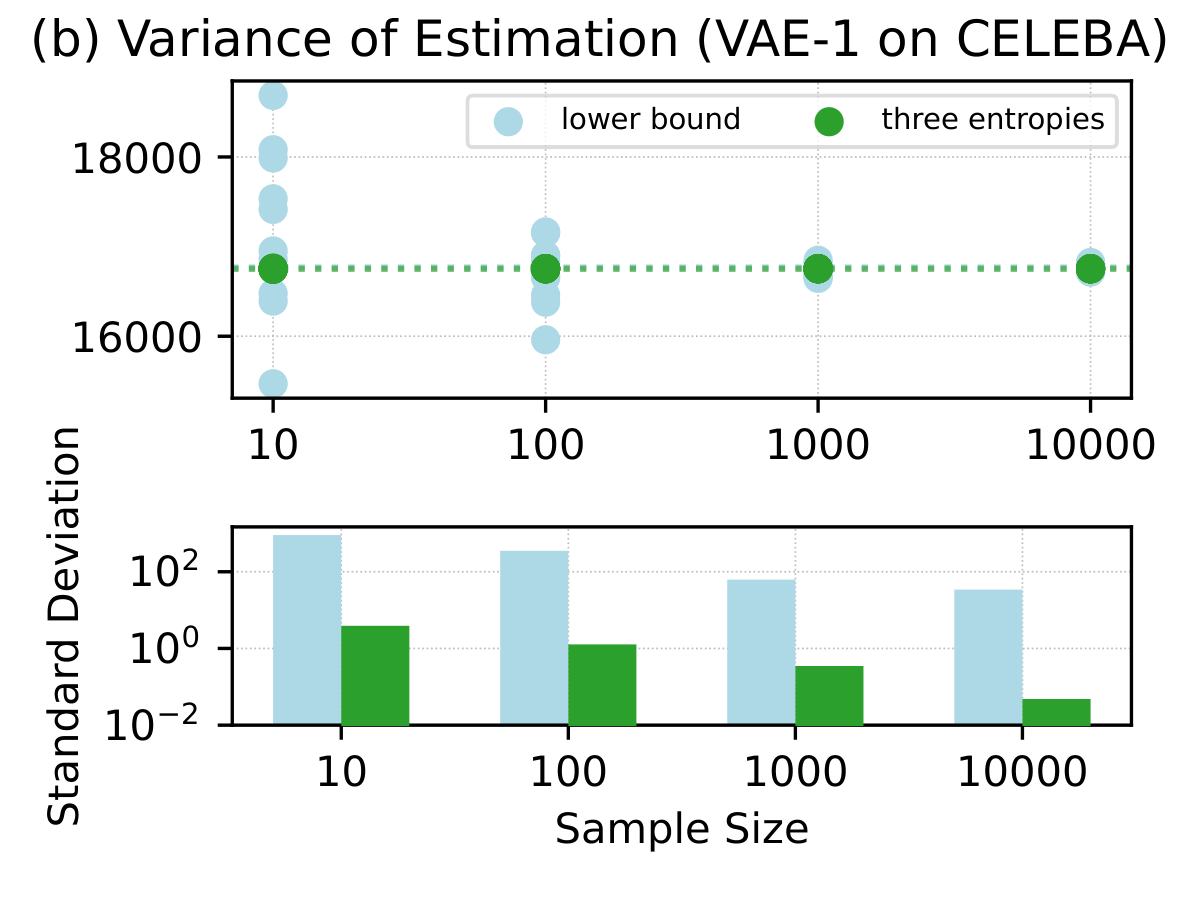}
\end{center}
\caption{\textbf{
ELBO Estimation} for the VAE--1 model on CelebA.
(a) ELBO and the sum of the three entropies during training. The close-up reveals that the ELBO of mini-batches fluctuates around the sum of entropies. 
(b) 
Direct approximation of ELBO and three entropies for the trained model with different sample sizes, repeated 10 times. Note, that the standard deviation is depicted on logarithmic scale.}
\label{fig:ELBOEstimation}
\end{figure}

The accuracy of ELBO estimation using \cref{theo:VAE-1} can be quantified. In \cref{fig:ELBOEstimation} (and \cref{fig:ELBOEstimation_App} in \cref{app:ELBOEstimation}) we compare ELBO estimation using the model parameters for the entropies to the conventional and direct solution of the ELBO integrals using mini-batches.

\paragraph{Model Selection for Linear VAEs}
In practice and for large data sets, PCA is often used as the first processing to reduce data dimensionality. Streaming PCA algorithms have found applications in `big data' domains where storing the full dataset is prohibitively expensive, e.g.,~high energy physics \citep{guglielmo_reconfigurable_2021} and online neural data analysis \citep{wu_streaming_2017, wu_deep_2018, migenda_adaptive_2021}. Here we demonstrate a straight-forward application of the three entropies result to online model selection in the case of linear VAEs applied to streaming data. Linear VAEs effectively perform probabilistic PCA \citep[e.g.,][]{LucasEtAl2019}.

For our experiments, we trained three linear VAEs of different complexity on non-stationary streaming data (see \cref{app:LinearVAEs} for details).
To perform model selection for such data, a very common approach is to compute the Bayesian Information Criterion \citep[BIC, e.g.,][]{schwarz_estimating_1978}, which requires an estimate of the likelihood at current model parameters and for different models. For linear VAEs the ELBO can itself be used as likelihood estimation (also see \cref{app:LinearVAEs}). Different ways to estimate the ELBO are conceivable. For instance, (1)~the presumably most common one would use the standard stochastic estimation of the ELBO itself (using the reparameterization trick and data batches), (2)~one could use the closed-form expression derived by \citet[][their App.~C]{LucasEtAl2019} and data batches, or (3)~the ELBO could be estimated based on the closed-form analytical solution by \citet[][]{TippingBishop1999}. Notably all these three alternatives require the data and are computationally demanding.\footnote{Alternative (1) requires potentially many samples, alternative (2) involves a series of matrix multiplications, and alternative (3) the computation of the data covariance matrix and eigenvalue computations.}
Using the result derived in this contribution, in particular \cref{corollary:LinearVAE}, we are provided with a novel alternative to estimate the likelihood. Importantly, and in contrast to all previous %
approaches, 
this alternative does not require the data nor does it involve any costly computations (see \cref{corollary:LinearVAE}).
Instead, we are merely using the variance parameters $\{\tau^2_h\}$ and $\sigma^2$ for \cref{EqnVAEBoundLinear}.
Of course, the linear VAE still has to be trained using a method of choice, which in a streaming setting can simply be standard stochastic VAE training. But any additional cost to estimate the ELBO itself is negligible knowing \cref{corollary:LinearVAE}.

\begin{figure}[ht!]
  \vspace{-0ex}
  \begin{center}
    \includegraphics[width=0.5\linewidth]{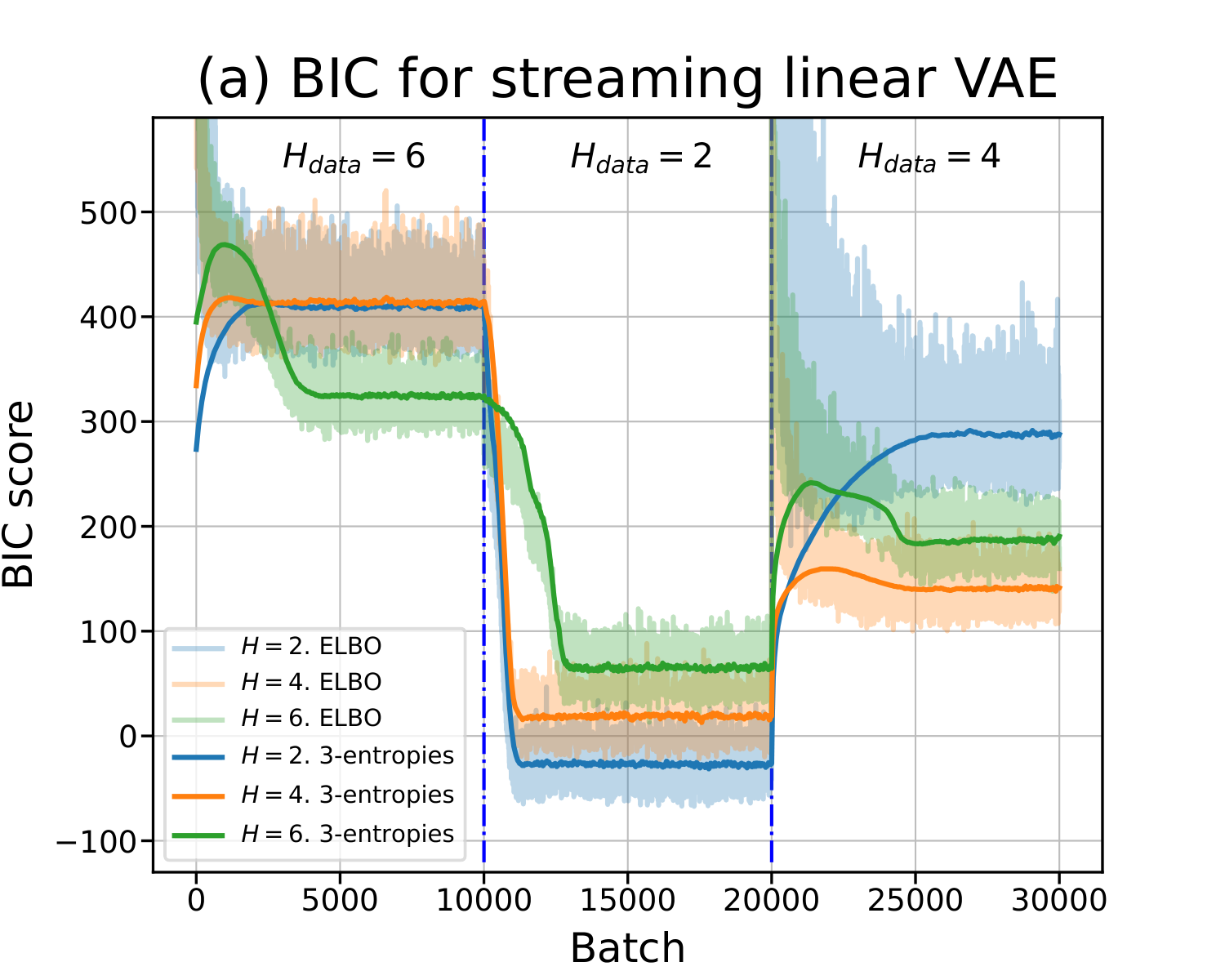}
    \hspace{-1em}
    \includegraphics[width=0.5\linewidth]{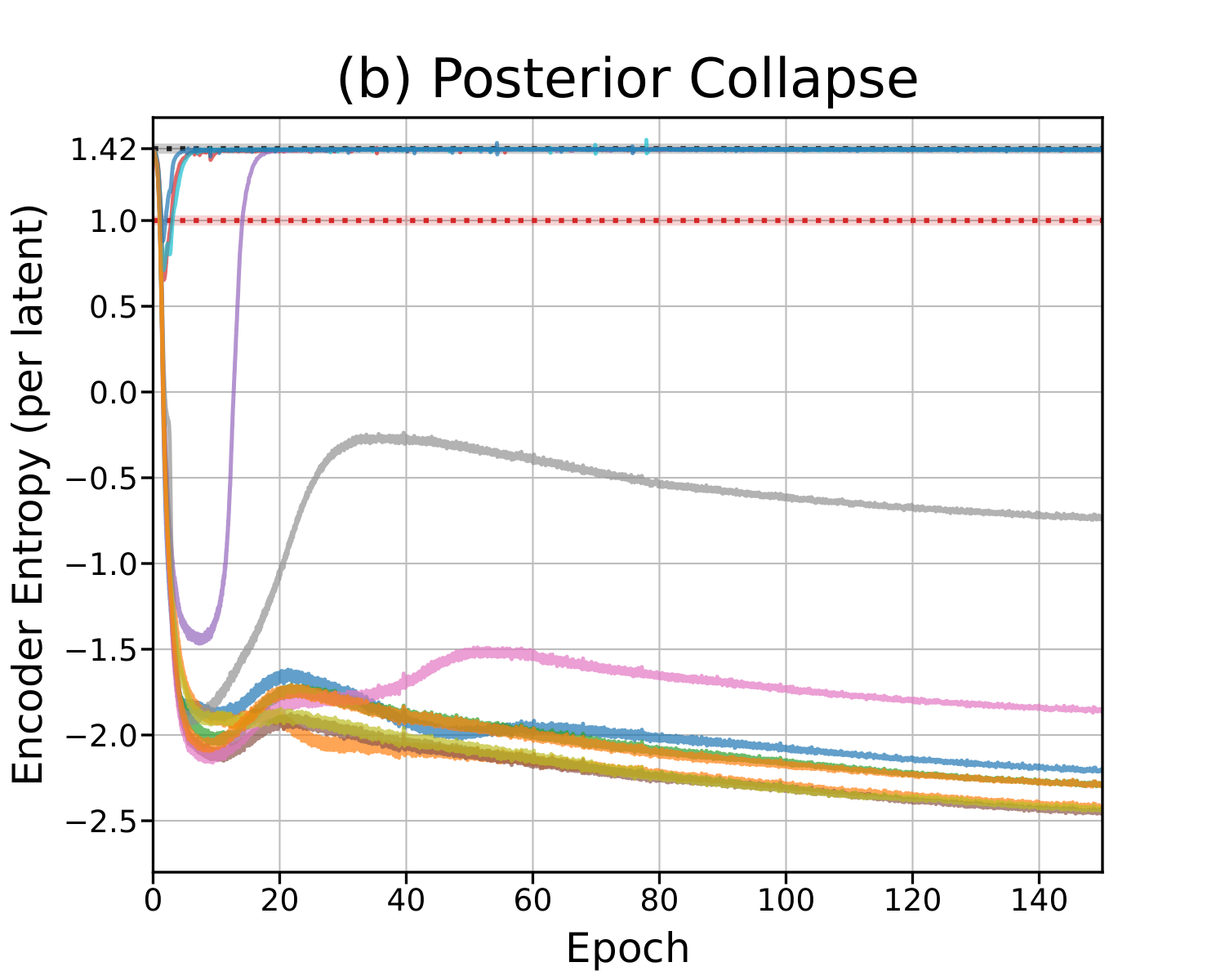}
  \end{center}
    \vspace{-2ex}
    \caption{\textbf{(a) Streaming VAE application.}
    Three linear VAEs trained on streaming data with changing dimensionality. BIC score based on online ELBO estimate (transparent plots) is very noisy. The three entropies BIC score (smooth solid plots) is a estimator depending only on the model parameters. Notice, that it clearly allows for easier and more stable model selection (the lower the BIC score, the better the model).
    \textbf{(b) Posterior collapse} monitoring for VAE--1 on SUSY. Latent variables are collapsed if $\frac{1}{N}\sum_{n=1}^N \HH[\qPhi(z_h\given\xVecN)] > 1$ (\cref{EqnPostCollCriterion}).
    See \cref{app:PosteriorCollapse,app:Experiments} for details on posterior collapse and experimental set-up.\vspace{-0ex}}
\label{fig:StreamingVAE_PosteriorCollapse}
\vspace{-1ex}
\end{figure}

In \cref{fig:StreamingVAE_PosteriorCollapse}(a) we compare three conventionally trained linear VAEs %
based on BIC scores computed using stochastic ELBO estimation and using \cref{corollary:LinearVAE} for ELBO estimation, respectively. 
The BIC score based on stochastic ELBO estimation has a high variance due to the noise in the used data batches. 
While both estimations allow for model selection, the entropy-based BIC score is inherently less noisy while reliably tracking the dimensionality information in the data stream. The experiments provide evidence for the entropy-based approach to allow for reliable, low variance model selection of VAEs in streaming settings, with almost no additional computation cost. More details are given in \cref{app:LinearVAEs}.


\paragraph{Entropy-based Analysis of Posterior Collapse.}
Posterior collapse is an empirically observed phenomenon:
a subset of VAE latent variables ceases to participate in encoding and decoding and instead assume a high variance \citep{HeEtAl2018,DiengEtAl2019}.
High encoder variance in turn means that the corresponding latents favor the optimization of the regularization score over the reconstruction score. We remark that posterior collapse is not an issue \emph{per se}, but needs to be carefully monitored such that -- for the data set with (unknown) intrinsic dimensionality at hand -- a reasonable amount of latent variables are pruned out.
\citet[][]{LucasEtAl2019} point out that ``despite a large volume of work studying posterior collapse, it has not been measured [or even defined] in a consistent way'' and proposed a measurement for posterior collapse based on the KL-divergence.
However, this requires two threshold values to be hand-set and is quite {\em ad hoc} (see \cref{eq:PostCollapse_Lucas} and the accompanying discussion).
Considering the above discussed entropy-based measures in \cref{eq:Scores_Reg_Rec},
posterior collapse can be defined much more naturally.
Starting with the assumption of uncorrelated Gaussians for VAE encoder and prior, we can use the entropy-based regularization score $S_{\mathrm{reg}}(\Phi,\Theta)$ and elementary properties of entropies to
derive a criterion for posterior collapse (we elaborate in \cref{app:PosteriorCollapse}).
Concretely, we will consider a latent distribution $\qPhi(z_h\given\xVecN)$ as collapsed if for $\delta > 0$
\begin{align}
\frac{1}{N}\sum_{n=1}^N \HH[\qPhi(z_h\given\xVecN)] >  \HH[p(z_h)] - \delta \enspace.  \label{EqnPostCollCriterion}
\end{align}
That is, a latent variable is collapsed if it is nearly as widely dispersed as its corresponding prior variable (and in turn provides no reliable information about the input). Note that we do not need to monitor the encoder means \citep[as opposed to][]{LucasEtAl2019}, the variance parameters suffice to detect posterior collapse.
\cref{fig:StreamingVAE_PosteriorCollapse}(b)
visualize this definition in practice for VAE--1 throughout the optimization process
with the threshold $\HH[p(\zVec)] - \delta$ set to one. The full experiments are presented in \cref{app:PosteriorCollapse_Exp} (\cref{fig:PosteriorCollapse_App}).

\section{DISCUSSION}
Stationary points of learning objectives are of central importance for our understanding of learning algorithms. Their properties are consequently frequently studied in the literature.
Many such theoretical results are obtained for elementary models \citep[][]{YedidiaEtAl2001,OpperSaad2001}, under idealized conditions \citep[][]{LuckeHenniges2012}, or for linear boundary cases \citep[][]{DaiEtAl2018,LucasEtAl2019}. Idealizations and/or boundary cases are especially relevant for deep generative models, for which general and exact theoretical results are notoriously difficult to obtain. %
Here we studied VAEs, which are one of the most popular deep generative models. Even though VAEs range among the currently most intricate models because of the DNNs they are based on, we can here report an exact and novel theoretical result: at all stationary points the VAE learning objective (the ELBO) becomes equal to a sum of three entropies.
\cref{theo:VAE-1,theo:VAE-3} report these concise results, which imply that the ELBO value at stationary points is determined exclusively by the entropies of those distributions defining a VAE. For the most common standard VAEs, the result notably provides closed-form expressions for the ELBO at stationary points. Importantly, only very mild assumptions have to apply for the involved DNNs:
We here essentially excluded weight-sharing between DNNs, and decoder DNNs required linear output units if also decoder covariances are set by DNNs. 
Furthermore, the derived results apply under realistic conditions, and we have numerically verified that the derived entropy-form of the ELBO has very high accuracy also in those vicinities of stationary points reached in practice. The reported results consequently go beyond being of purely theoretical relevance, i.e., the derived expressions can directly be used in practice for any task that requires the estimation of the ELBO itself. To give an intuition, we discussed (theoretically and numerically) tasks such as ELBO estimation, model selection for streaming data, or the analysis of posterior collapse. 
We do stress, however, that the main contribution remains the theoretical result itself: the ELBOs of the most common VAEs converge to %
entropy sums.

Future work will be based on the observation that only subsets of parameters have to be at stationary points, which can allow for purely entropy-based learning objectives to optimize, e.g., encoder DNNs. Also the investigation of VAEs with less standard prior distributions as well as the investigation of stacked VAEs or other deep generative models \citep[e.g.,][]{BondTaylerEtAl2022} represent natural future research directions.

\subsubsection*{Acknowledgments}
This work is funded by the German Research Foundation (DFG) within the priority program SPP~2298 “Theoretical Foundations of Deep Learning” - project 464104047 (FI~2583/1-1 and LU~1196/9-1).

\bibliography{References}

\appendix
\onecolumn

\onecolumn

\appendix

\section{LINEAR VAES AND STREAMING APPLICATIONS}
\label{app:LinearVAEs}
For the proof of \cref{corollary:LinearVAE} note that for the linear DNN of the decoder~(i.e., for the matrix multiplication) of \cref{EqnVAElinearDecoder} the same reparametrization is possible as was used for VAE--1 (i.e., using $\tilde{W}$ with constraint columns instead of $W$ in \cref{EqnVAElinearDecoder}). Furthermore, no conditions were imposed on the encoder DNN for \cref{theo:VAE-1}, and the used properties of the
stationary points (\cref{EqnStatPoints}) are the same for the linear VAE as for the VAEs above.
\cref{theo:VAE-1} consequently applies for the linear VAE as a special case of \mbox{VAE--1}.

The result of \cref{corollary:LinearVAE} then highlights 
some properties of the variational bound at convergence. First, note that the variational bound at the
stationary points can be computed efficiently and solely based on the variance parameters $\sigma^2$ and $\tau^2_h$. 
For linear VAEs, the bound is even independent of the data points, i.e., just the $H+1$ variance parameters determine its value.
In addition to this simplification, the linear VAE has a further property that makes it interesting from a theoretical perspective.
Linear VAEs can be used to recover the maximum likelihood solution \citep[compare][]{DaiEtAl2018,LucasEtAl2019} of probabilistic PCA (p-PCA). 
\citet[][]{LucasEtAl2019}, for instance, showed that no stable stationary points of the variational lower bound exist other than the global maximum. Training of the
linear VAE will thus always converge to the global maximum of the lower bound. Second, they showed that the  
linear encoder is flexible enough to finally recover full posteriors exactly, which makes the variational lower bound tight at convergence. 
As the decoder is identical to the generative model of p-PCA \citep[][]{TippingBishop1999}, the linear VAE thus converges to recover the optimal
p-PCA likelihood. According to \cref{corollary:LinearVAE} it thus applies that after convergence of the linear VAE, \cref{EqnVAEBoundLinear} is equal
to the p-PCA log-likelihood.

We can combine those earlier results \citep[e.g.,][]{TippingBishop1999,LucasEtAl2019} with \cref{corollary:LinearVAE} and obtain the following.

\noindent{}{\bf Corollary 2.}
\textit{Consider the linear VAE defined by \cref{EqnVAElinearDecoder,EqnVAElinearEncoder} with decoder parameters $\Theta=(\sigma^2,W,\muVec_0)$.
Then after convergence, the parameters $(\sigma^2,W,\muVec_0)$ represent the maximum likelihood
solution for p-PCA, and the value of the log-likelihood $\LL(\Theta)$ is given by:}%
\begin{equation}
\LL(\Theta) = \frac{1}{2}\sum_h \log\!\big( \tau^2_h \big) - \frac{D}{2} \log\!\big( 2\pi e \sigma^2 \big) ,\label{EqnVAEBoundLinearC}
\end{equation}
\textit{where $\tau^2_h$ are the learned variances of the VAE encoder.}
\vspace{1ex}

\noindent{}{\textit{Proof.}}
As Eqn.\,(\ref{EqnVAEBoundLinear}) applies for all stationary points, it also applies for the global maximum of the variational lower bound. At the global maximum, the variational
bound is equal to the p-PCA log-likelihood \citep[e.g.,][]{LucasEtAl2019}, which proves the claim. \hfill $\square$\\[-2ex]

Other than for standard non-linear VAEs, the existence of a closed-form result is itself not surprising for the linear VAE.
The linearities make analytic solutions of the integrals of the variational bound possible. Indeed,
we can instead of Eqn.\,(\ref{EqnVAEBoundLinearC}) simply use the well-known closed-form solution of the p-PCA likelihood \citep[][]{TippingBishop1999}:
\begin{equation}
 \LL(\Theta) = -\frac{D}{2} \log(2\,\pi) -\frac{1}{2} \log\!\big(\det(C)\big)  - \frac{1}{2} \Tr{}(C^{-1}S) ,\label{EqnLLPCA}
\end{equation}

\begin{equation}
\mbox{where}\ C=WW^T+\sigma^2\One\ \mbox{\ and where \ }\ S=\frac{1}{N}\sum_{n}(\xVecN-\muVec)(\xVecN-\muVec)^T \nonumber
\end{equation}
is the data covariance matrix. 
At convergence, we thus have two alternatives to compute the log-likelihood: Eqn.\,(\ref{EqnLLPCA}) and Eqn.\,(\ref{EqnVAEBoundLinearC}).
The two expressions differ, however. While both are closed-form, the well-known p-PCA likelihood (\cref{EqnLLPCA}) requires
the data in the form of the data covariance matrix $S$. And even neglecting the computational effort to compute $S$, the computation of (\cref{EqnLLPCA}) is much
more expensive than computing (\ref{EqnVAEBoundLinearC}): For $\LL(\Theta)$ the inverse and the determinant of the $D\times{}D$ matrix $C$ have to be computed.
In contrast, for the computation of $\FF(\Phi,\Theta)$ in \cref{EqnVAEBoundLinearC}
no such computations nor data is required (although we require a linear VAE that has sufficiently converged). The derived result of Eqn.\,(\ref{EqnVAEBoundLinearC}) can therefore be of theoretical and
practical relevance especially considering the exceptionally widespread use of PCA in Machine Learning, Statistics and beyond.

\paragraph{Streaming Applications}
One concrete practical example is PCA applied to streaming data, which is an increasingly common setting especially in recent years, see e.g., \citet[][]{AllenZhuEtAl2017} for an overview and \citet[][]{ChouEtAl2020} for a recent example..
In such a setting the application of a linear VAE is straight-forward, and it can be operated as alternative or in parallel to other PCA algorithms.
For streaming data (as for other data) the value of the likelihood itself can be of high interest, e.g., to monitor the fit to the data or for the selection of PCA dimensions. In this context, Eqn.\,(\ref{EqnVAEBoundLinearC}) provides an
exceedingly easy way to compute the log-likelihood value. Also if the streamed data changes, the parameters of the linear VAE will change and 
Eqn.\,(\ref{EqnVAEBoundLinearC}) can be used to track the changes of the likelihood without any effort. No data has to be kept in memory and
no costly computations are required.

For the purposes of this paper, our first numerical experiments use the linear VAE (\cref{app:Experiments}). The result can then directly be compared to the well known closed-form log-likelihood (\cref{EqnLLPCA}) as well as to sampling based approximations of the variational lower bound. \cref{app:Experiments} also shows empirically that the derived three entropy expressions match the lower bound with high accuracy also in vicinity of stationary points that are
reached in practice.

For our experiments with streaming VAEs, we construct an artificial dataset that comprises three chunks of data of dimensionality $D=10$. Each chunk is limited to have different intrinsic dimensionality $H_{data} \in \{2, 4, 6\}$ by setting the remaining Gaussian covariance matrix eigenvalues to zero. We feed this dataset into three separate linear VAEs with latent space dimensionality $H$ of 2, 4, and 6 by providing small batches of 10 data points each in order to simulate the streaming training regime.

To compute the BIC for model $\mathcal{M}$ we used the following equation \citep{wit_all_2012}:
\begin{align}
  \textrm{BIC}(\mathcal{M}) = -2 \, \hat{l}(X) + P \log(N),
\end{align}
where $\hat{l}(X)$ is the maximum log-likelihood of the data $X$,  $N$ is the number of data points, and $P$ is the number of parameters which have only point estimation (\textit{maximum likelihood} or \textit{maximum a posteriori}). In our experiment with linear VAEs $P$ is equal to the number of elements in the generative matrix plus one for the observation noise variance: $P = H D + 1$.
As learning the posterior distribution for a subset of the parameters (in our case it is posterior over $\zVec$) for VAEs yields a lower bound of the log-likelihood, effectively integrating them out, we exclude the latent points $\zVec$ and the amortized encoder from the set of point-estimated parameters.

In the numerical experiment for training we used 100 (reparameterized) random samples for each data point to estimate the ELBO and the gradient. Each batch of 10 data points was used to run only one step of ADAM optimization over the parameters. We set the learning rate to 0.002, which we found to be a good trade-off between the speed of convergence and the update noise for such a simple linear model.

\section{PROOF OF THEOREM \ref{theo:VAE-3}}
\label{sec:appProofPropThree}
The proof of \cref{theo:VAE-3} which applies for general VAE decoders of VAE--3 of \cref{def:VAE-3} follows a similar intuition as the proofs of \cref{theo:VAE-2,theo:VAE-1}. In many aspects it is, however, significantly more intricate.
An important element shared by the more general proof and the proofs for VAE--1 is a reparameterization of the original VAE model by using part of the DNN weights of the decoder to parameterize the prior distribution. 

Before we reparameterize, let us reconsider the DNNs of the VAE-3 decoder which are given by:
\begin{align}
\muVec_{\Theta}(\zVec) &= \muVec(\zVec, W)\ \ \mbox{and}\ \ \\
\Sigma_{\Theta}(\zVec) &= \diag\big(\sigmaT_1(\zVec, M),\ldots,\sigmaT_D(\zVec, M)\big)\ \ 
\mbox{and}\ \ \sigTVec(\zVec, M) = \big(\sigmaT_1(\zVec, M), \ldots, \sigmaT_D(\zVec, M) \big)^{\mathrm{T}}.
\end{align}

The expressions make explicit that the two DNNs depend on two different sets of parameters ($W$ and $M$, respectively). Furthermore, let us reiterate the DNNs themselves which are, in more detail, given as follows:
\begin{align}
\muVec(\zVec, W) &= \mathrm{DNN}_{\mu}(\zVec; \Wall) =   W^L\Scal( W^{L-1} \Scal( \cdots \Scal(W^0\zVec + \bVec^{0}) \cdots + \bVec^{L-1}) + \bVec^L, \label{EqnMuDNNApp}\\
\sigTVec(\zVec, M) &= \mathrm{DNN}_{\sigma}(\zVec; \Mall)  =  M^{L} \Scal( M^{L-1} \Scal( \cdots \Scal(M^0\zVec + \,\cVec^{0}) \cdots + \cVec^{L-1}) + \cVec^{L}, \label{EqnSigDNNApp}
\end{align}
where $W^l$ and $\bVec^l$ are the weight matrices and biases of $\mathrm{DNN}_{\mu}(\zVec; \Wall)$, and where $M^l$ and $\cVec^l$ are the weight matrices and biases of $\mathrm{DNN}_{\sigma}(\zVec; \Mall)$. 
$\mathrm{DNN}_{\mu}(\zVec; \Wall)$ and $\mathrm{DNN}_{\sigma}(\zVec; \Mall)$ {\em can} have different numbers of layers (in the main text we used $L$ and $L'$ for maximal layer indices). To simplify notation for the proof, we dropped the distinction (we just use $L$ for both DNNs), and mainly $\mathrm{DNN}_{\sigma}(\zVec; \Mall)$ will be of interest.
$W$ and $M$ we take to contain all weight matrices and biases for their corresponding DNN. 
Both DNNs have a linear output layer and point-wise non-linearities $\Scal(\cdot)$, and each of these two properties is a standard assumption for DNNs \citep[compare, e.g.,][]{Yarotsky2017,AroraEtAl2018}. The non-linearities also {\em can} differ between the DNNs, and they can be different from layer to layer. 
For the variance DNN we assume that learning does not result in degeneracies at stationary points, i.e., we assume that $\sigT_d(\zVec;\Theta)$ is never equal to zero in order to avoid singularities.
In practice, variances equal zero mean perfect data reconstruction, so our assumption at stationary points only excludes rather artificial conditions.

After reiterating the DNNs in more detail, the first step of the proof is now  a reparameterization of VAE--3 along the same lines as done for VAE--1 in \cref{SecReparameterization}, i.e., we consider the following auxiliary VAE model:
\begin{align}
 p_\Amat(\zVecT) &= \NN(\zVecT; \mathbf{0}, \Amat)  \enspace,  \label{Eqn:VAE4Prior}\\ 
\qPhi(\zVecT \given \xVec) &= \NN\big(\zVecT; \tilde{\nuVec}_{\Phi}(\xVec), \tilde{\TT}_{\Phi}(\xVec)\big)  \enspace, \\
\pT(\xVec \given \zVecT) &= \NN(\xVec; \tilde{\muVec}_{\Theta}(\zVecT), \tilde{\Sigma}_{\Theta}(\zVecT)\big) \enspace \label{Eqn:VAE4Noise},
\end{align}
where decoder mean $\muT_\Theta(\zVecT) = \mathrm{DNN}_{\muT}(\zVecT; W)$
and decoder covariance $\tilde{\Sigma}_{\Theta}(\zVec) = \mathrm{DNN}_{\sigT}(\zVecT; \Mall)$ are the same as the DNNs for VAE--3 in \cref{EqnMuDNNApp,EqnSigDNNApp} with the exception that
their first weights ($W^0$ and $M^0$) now have their column vectors constrained to unit norm. The prior distribution is now parameterized by the diagonal matrix $\Amat = \diag\left(\alphaT_1, \dots, \alphaT_H\right)$. For $\Amat$ we used with $\alphaT_h$ instead of $(\alpha_h)^2$ an abbreviation analogous to the one for the observable variances (and we assume $\alphaT_h>0$ for all $h$).

With the same argumentation as given in \cref{SecVarBounds}, the reparameterized VAE does parameterize the same model as VAE--3 of \cref{def:VAE-3}.  
As preparation for the proof steps below, we abbreviate the decoder DNN for the variances (\cref{EqnSigDNNApp}) as follows:
\begin{equation}
\sigT_d(\zVecT; \Theta) =  \big(\mathrm{DNN}_{\sigT}(\zVecT; \Mall)\big)_d\ =\ \sum_j M^{L}_{dj}\ \varphi_j(\zVecT;\Mall)\,+\,c^{L}_d,\label{ProofDNNsigmaApp}
\end{equation}
where $\varphi_j(\zVecT;\Theta)$ is simply the remainder of the DNN after the lowest weights are removed (i.e., $\varphi_j(\zVecT;\Theta)$ is the output of unit $j$ in layer $L-1$). We demanded in \cref{def:VAE-3} that the network has at least one hidden layer, i.e., $L\geq{}1$, which ensures
that the weights $M^{L}$ and $M^{0}$ represent two different weight matrices (and $\cVec^{L}$ and $\cVec^0$ two different bias vectors). The constraint on  $M^{0}$ (i.e., that it has unit length columns) does therefore not effect $M^{L}$.

Using the parameterization in \cref{Eqn:VAE4Prior,Eqn:VAE4Noise} we could now in principle prove \cref{theo:VAE-3} explicitly. However, for our purposes it is more convenient to make use of a recent generalization of the proof of \cref{theo:VAE-1} which includes also non-Gaussian distributions \citep[see][]{lucke2022convergence}. 
We note that the work by \cite{lucke2022convergence} is work in parallel to this contribution, and that it does not treat VAEs nor other DNN-based models.
We will, however, show in the following how the VAE--3 model can be shown to satisfy the conditions for which the general result applies.

The general result \citep[see][Theorem 1]{lucke2022convergence} applies for any generative model with one set of latents and one set of observables, where latent and observable distributions are both exponential family distributions (with constant base measure). If the link between latents and observables then fulfills a specific parameterization condition then we can conclude a convergence to entropy sums. The crucial property of VAE--3 is that the link is given by the decoder DNNs of \cref{EqnMuDNNApp,EqnSigDNNApp}. The task in the following will be to use the reparameterized version of VAE--3 in Equations (\ref{Eqn:VAE4Prior}) to (\ref{Eqn:VAE4Noise}), and to then show that the conditions for the general theorem \citep[][Theorem 1]{lucke2022convergence} are fulfilled. 

The conditions for the theorem are formulated for the exponential family parameterization of a generative model's distributions. An exponential form of the VAE in Equations (\ref{Eqn:VAE4Prior}) to (\ref{Eqn:VAE4Noise}) can be given as follows. First, the Gaussian prior can be rewritten as:
\begin{align}
& p_\Amat(\zVecT) = \NN(\zVecT; \mathbf{0}, \Amat) = h(\zVecT) \exp\big( \xiVec(\Amat)^{\mathrm{T}}\,\TVec(\zVecT)\,-\,\Acal(\xiVec(\Amat))\big) =  p_{\xiVec(\Amat)}(\zVecT)\\
\mbox{where\ \ } & \TVec(\zVecT)=\zVecT \odot \zVecT,\ \ \xi_h(\Amat)=-\frac{1}{2\alphaT_h}\ \ \mbox{and\ \ }\ h(\zVecT) = (2\pi)^{-\frac{H}{2}},\ \ %
\Acal\big(\xiVec(\Amat)\big) = \frac{1}{2}\sum_h \log(\alphaT_h),\label{EqnPriorExpFa}
\end{align}
with $\odot$ denoting point-wise multiplication, and with $\xiVec(\Amat)$ denoting the natural parameters of the prior. 
The exponential family form of the observable distribution is more intricate because it contains the DNNs and it has a  non-zero mean. An exponential family parameterization can be given as follows:
\begin{align}
& \pT(\xVec \given \zVecT) = \NN(\xVec; \tilde{\muVec}_{\Theta}(\zVecT), \tilde{\Sigma}_{\Theta}(\zVecT)\big) = h(\xVec) \exp\Big( \etaVec(\zVecT; \Theta)^{\mathrm{T}}\,\TVec(\xVec)\,-\,\Acal\big(\etaVec(\zVecT; \Theta)\big)\Big) =  p_{\etaVec(\zVecT; \Theta)}(\xVec)\\
\mbox{where\ \ } & 
\TVec(\xVec) = \left(\begin{array}{c} \xVec \\ \xVec \odot \xVec \end{array}\right),\ \ 
\etaVec(\zVecT; \Theta) = \left(\begin{array}{c} \muVec(\zVecT; W) \odot \sVec(\zVecT;M) \\[1mm] -\frac{1}{2}\,\sVec(\zVecT;M) \end{array}\right),
\ \mbox{with\ }\ \sVec(\zVecT;M) = \left(\begin{array}{c} \frac{1}{\sigT_1(\zVecT;M)} \\ \vdots \\  \frac{1}{\sigT_D(\zVecT;M)} \end{array}\right),\label{EqnDNNLink}\\
\mbox{and \ \ } & h(\xVec) = (2\pi)^{-\frac{D}{2}},\ \  \Acal\big(\etaVec(\zVecT; \Theta)\big) = \frac{1}{2}\sum_d \Big( \frac{\big(\mu_d(\zVecT; W)\big)^2}{\sigT_d(\zVecT;M)}\,+\,\log\big( \sigT_d(\zVecT;M) \big) \Big).
\end{align}
The reformulation of the VAE in terms of an exponential family parameterization (and with $h(\zVecT)$ and $h(\xVec)$ being constant base measures) is a first prerequisite for the result in \citep[][]{lucke2022convergence} to be applicable. The crucial property that is needed for the theorem to apply now concerns the natural parameter vectors $\xiVec(\Amat)$ and $\etaVec(\zVecT; \Theta)$, where the second is defined by potentially intricate and large DNNs. 
The condition for the first function, $\xiVec(\Amat)$, is relatively easy to show. We do require that for any vector $\vVec\in\RRR^H$ applies that:
\begin{align}
\IIT_{(\Amat)}\ \vVec &\ =\ \mathbf{0} \ \ \ \Rightarrow\ \ \ \
\xiVec^{\mathrm{T}}_{(\Amat)}\ \vVec\ =\ 0\,, \ \ \mbox{where}\ \ 
\big( \IIT_{(\Amat)} \big)_{hh'} \ =\ \disT  \del{\alphaT_h}\, \xi_{h'}(\Amat) \,
\label{EqnParaCondPrior} 
\end{align}
is the (transposed) Jacobian of $\xiVec(\Amat)$. Using \cref{EqnPriorExpFa} for $\xi_{h'}(\Amat)$, we get $\IIT_{(\Amat)}=\frac{1}{2}\, \diag(\alphaT_1^{-2},\ldots,\alphaT_H^{-2})$, i.e., $\IIT_{(\Amat)}$ is a squared and diagonal matrix with positive entries on the diagonal. Condition (\ref{EqnParaCondPrior}) is consequently almost trivially fulfilled as the first equation implies $\vVec=\mathbf{0}$ which implies $\xiVec^{\mathrm{T}}_{(\Amat)}\ \vVec\ =\ 0$.

The corresponding condition for the function $\etaVec(\zVecT; \Theta)$ is the for our purposes crucial part because this function (defined by \cref{EqnDNNLink}) links the latents to the observables using the decoder DNNs. 
For the link $\etaVec(\zVecT; \Theta)$, it will turn out to be easier if we show a more general result for VAEs than stated by \cref{theo:VAE-3}. Concretely, it will be easier to proof that equality to entropy sums 
holds also if only a subset of the parameters $\Theta$ have converged to a stationary point. We denote this subset by $\thetaVec$ which we take to contain all weights and biases of the output layer of the variance network, i.e., $M^L$ and $\cVec^L$ (see \cref{ProofDNNsigmaApp}). We take $\thetaVec$ to contain all these weights and biases arranged in one long column vector with scalar entries. All other parameters we take as being fixed (but they can have arbitrary values). 

For the link function $\etaVec_{(\zVecT;\Theta)}$, we now have to show \citep[see][Def.\,C]{lucke2022convergence} that the following specific condition is true:\\[-2ex]

For any function $\gVec:\RRR^H\rightarrow\RRR^{2D}$ from the latents to the space of natural parameters $\etaVec$ of the observable distribution it has to hold that
\begin{align}
\int \JJT_{(\zVecT;\thetaVec)}\ \gVec(\zVecT) \,\mathrm{d}\zVecT\ =\ \boldsymbol{0}
\ \ \Rightarrow\ \ 
\int \etaVec^{\mathrm{T}}_{(\zVecT;\Theta)}\ \gVec(\zVecT) \,\mathrm{d}\zVecT\ =\ 0\,,
\label{EqnParaCondNoise}
\end{align}
where $\JJT_{(\zVecT;\thetaVec)}$ denotes the (transposed) Jacobian of $\etaVec_{(\zVecT;\Theta)}$ constructed only using the subset $\thetaVec$, i.e.,
\begin{align}
\JJT_{(\zVecT;\,\thetaVec)} \ =\ \disT \Big( \del{\thetaVec} \eta_1(\zVecT;\Theta), \ldots, \del{\thetaVec}\eta_{2D}(\zVecT;\Theta) \Big)\,.
\end{align}
Some intuition on how the general proof then works is the following. Similar to the proof of \cref{theo:VAE-2}, we rewrite the different terms of the ELBO, \cref{EqnELBO}, in terms of entropies. For $\FF_2(\Phi,\thetaVec)$ this results (in exponential family notation) in the following expression:\vspace{-1ex}
\begin{align}
\FF_2(\Phi,\Theta)\ &= -\frac{1}{N} \sum_{n} \EEE{\qn_{\Phi}}{ \HH[\,p_{\etaVec(\zVecT; \Theta)}(\xVec)] } \,-\, \int \etaVecT_{(\zVecT;\,\Theta)}\ \gVec(\zVecT)\ \mathrm{d}\zVecT \label{EqnFFThreeShort}
\end{align}
where $\gVec(\zVecT)$ is a function whose concrete properties are not relevant for the proof.
At stationary points of the parameters $\thetaVec$ we can show that $\int \JJT_{(\zVecT;\thetaVec)}\ \gVec(\zVecT) \,\mathrm{d}\zVecT\ =\ \boldsymbol{0}$ applies for the same function $\gVec(\zVecT)$. So if condition (\ref{EqnParaCondNoise}) holds, then the integral in expression (\ref{EqnFFThreeShort}) vanishes, and $\FF_2(\Phi,\Theta)$ becomes equal to an (expected) entropy. For more details see \citep[][]{lucke2022convergence}.

To show that condition (\ref{EqnParaCondNoise}) holds in our case, we now write $\gVec(\zVecT)=\left(\begin{array}{c} \gVecA(\zVecT) \\[1mm] \gVecB(\zVecT) \end{array}\right)$, i.e., we take the function to consist of two functions $\gVecA(\zVecT)$ and $\gVecB(\zVecT)$ mapping to $\RRR^D$. Now using the definition of $\etaVec_{(\zVecT;\Theta)}$ in Equation (\ref{EqnDNNLink}), we obtain: 
\begin{align}
\boldsymbol{0} &= \int \JJT_{(\zVecT;\thetaVec)}\ \gVec(\zVecT) \,\mathrm{d} \zVecT \label{DeriCondInit}\\
\Rightarrow\ \ \ \boldsymbol{0} &= \int \Big(\ \del{\thetaVec}\big( \muVec(\zVecT; W) \odot \sVec(\zVecT;M) \big)^{\mathrm{T}},\  -\frac{1}{2}\,\ \del{\thetaVec}\big( \sVec(\zVecT;M) \big)^{\mathrm{T}} \,\Big)\ \left(\begin{array}{c} \gVecA(\zVecT) \\[1mm] \gVecB(\zVecT) \end{array}\right)\ d\zVecT \\
\Rightarrow\ \ \ \boldsymbol{0} &= \int \Big( \sum_{d=1}^D \ \mu_d(\zVecT; W)\  \big(\ \del{\thetaVec} s_d(\zVecT;M) \big) \ \gA_d(\zVecT) \ -\ \frac{1}{2}\ \sum_{d=1}^D  \big(\ \del{\thetaVec} s_d(\zVecT;M) \big) \ \gB_d(\zVecT) \Big)\ d\zVecT\,. \label{EqnCondDeri}
\end{align}
As we took $\thetaVec$ to contain all weights $M^L_{dj}$ of the output layer of $\mathrm{DNN}_{\sigmaT}(\zVecT; \Mall)$ and the corresponding biases $c^L_d$, we conclude that \cref{EqnCondDeri} implies that
\begin{align}
\mbox{for all $d$ and $j$:}&&
\int \sum_{d'=1}^D \Big( \mu_{d'}(\zVecT; W)\  \big(\ \Del{s_{d'}(\zVecT;M)}{M^L_{dj}} \big) \ \gA_{d'}(\zVecT) \ -\ \frac{1}{2}\,\big(\ \Del{s_{d'}(\zVecT;M)}{M^L_{dj}}  \big) \ \gB_{d'}(\zVecT) \Big)\ d\zVecT &\ =\ 0,\label{CondDeriW}\\
\mbox{and for all $d$:}&& 
\int \sum_{d'=1}^D \Big( \mu_{d'}(\zVecT; W)\  \big(\ \Del{s_{d'}(\zVecT;M)}{c^L_{d}} \big) \ \gA_{d'}(\zVecT) \ -\ \frac{1}{2}\,\big(\ \Del{s_{d'}(\zVecT;M)}{c^L_{d}}  \big) \ \gB_{d'}(\zVecT) \Big)\ d\zVecT &\ =\ 0, \label{CondDeriB}
\end{align}
We now evaluate the derivatives using the definition of $\sVec(\zVecT;M)$ in \cref{EqnDNNLink} and by inserting expression (\ref{ProofDNNsigmaApp}) for $\sigT_{d'}(\zVecT; \thetaVec)$. We obtain:
\begin{align}
\Del{s_{d'}(\zVecT;M)}{M^L_{dj}} &= \del{M^L_{dj}}\,\big( \frac{1}{\sigT_d(\zVecT;M)} \Big)
\ =\ -\, \delta_{dd'}\ \frac{\varphi_j(\zVec;M)}{ \big( \sigT_{d'}(\zVecT;M) \big)^2},\\
\mbox{and \ \ \ }\ \,\Del{s_{d'}(\zVecT;M)}{c^L_{d}}\, &= \del{c^L_{d}}\,\big( \frac{1}{\sigT_{d'}(\zVecT;M)} \Big)
\ =\ -\, \delta_{dd'}\ \frac{1}{ \big( \sigT_d(\zVecT;M) \big)^2}.
\end{align}
By inserting the derivatives into expressions (\ref{CondDeriW}) and (\ref{CondDeriB}), we obtain:
\begin{align}
\mbox{for all $d$ and $j$:}&&
\int \Big( \mu_{d}(\zVecT; W)\  \frac{\varphi_j(\zVec;M)}{ \big( \sigT_d(\zVecT;M) \big)^2} \ \gA_{d}(\zVecT) \ -\ \frac{1}{2}\,\frac{\varphi_j(\zVec;M)}{ \big( \sigT_d(\zVecT;M) \big)^2} \ \gB_{d}(\zVecT) \Big)\ d\zVecT &\ =\ 0,\label{CondDeriWW}\\
\mbox{and for all $d$:}&& 
\int \Big( \mu_{d}(\zVecT; W)\  \frac{1}{ \big( \sigT_d(\zVecT;M) \big)^2} \ \gA_{d}(\zVecT) \ -\ \frac{1}{2}\,\frac{1}{ \big( \sigT_d(\zVecT;M) \big)^2} \ \gB_{d}(\zVecT) \Big)\ d\zVecT &\ =\ 0. \label{CondDeriBB}
\end{align}
We now multiply expression (\ref{CondDeriWW}) by $W^L_{dj}$ and then sum over $j$ to obtain:
\begin{align}
\mbox{For all $d$:}&& \sum_j\,W^L_{dj}
\int \Big( \mu_{d}(\zVecT; W)\  \frac{\varphi_j(\zVec;M)}{ \big( \sigT_d(\zVecT;M) \big)^2} \ \gA_{d}(\zVecT) \ -\ \frac{1}{2}\,\frac{\varphi_j(\zVec;M)}{ \big( \sigT_d(\zVecT;M) \big)^2} \ \gB_{d}(\zVecT) \Big)\ d\zVecT &\ =\ 0,\label{CondDeriWWWhfjad}\\
\Rightarrow\ \ \ \mbox{For all $d$:}&& 
\int \Big( \mu_{d}(\zVecT; W)\  \frac{\sum_j\,W^L_{dj}\,\varphi_j(\zVec;M)}{ \big( \sigT_d(\zVecT;M) \big)^2} \ \gA_{d}(\zVecT) \ -\ \frac{1}{2}\,\frac{\sum_j\,W^L_{dj}\,\varphi_j(\zVec;M)}{ \big( \sigT_d(\zVecT;M) \big)^2} \ \gB_{d}(\zVecT) \Big)\ d\zVecT &\ =\ 0.\label{CondDeriWWW}
\end{align}
Expression (\ref{CondDeriBB}) we multiply by $c^L_d$ to obtain:
\begin{align}
\mbox{For all $d$:}&& 
\int \Big( \mu_{d}(\zVecT; W)\  \frac{c^L_d}{ \big( \sigT_d(\zVecT;M) \big)^2} \ \gA_{d}(\zVecT) \ -\ \frac{1}{2}\,\frac{c^L_d}{ \big( \sigT_d(\zVecT;M) \big)^2} \ \gB_{d}(\zVecT) \Big)\ d\zVecT &\ =\ 0. \label{CondDeriBBB}
\end{align}
By adding equations (\ref{CondDeriWWW}) and (\ref{CondDeriBBB}) we can conclude:
\begin{align}
\mbox{For all $d$:}&&
\int \Big( \mu_{d}(\zVecT; W)\  \frac{\sum_j\,W^L_{dj}\,\varphi_j(\zVec;M)\,+\,c^L_d}{ \big( \sigT_d(\zVecT;M) \big)^2} \ \gA_{d}(\zVecT) \ -\ \frac{1}{2}\,\frac{\sum_j\,W^L_{dj}\,\varphi_j(\zVec;M)\,+\,c^L_d}{ \big( \sigT_d(\zVecT;M) \big)^2} \ \gB_{d}(\zVecT) \Big)\ d\zVecT &\ =\ 0\\
\Rightarrow\ \ \ \mbox{For all $d$:}&& 
\int \Big( \mu_{d}(\zVecT; W)\  \frac{\sigT_d(\zVecT;M)}{ \big( \sigT_d(\zVecT;M) \big)^2} \ \gA_{d}(\zVecT) \ -\ \frac{1}{2}\,\frac{\sigT_d(\zVecT;M)}{ \big( \sigT_d(\zVecT;M) \big)^2} \ \gB_{d}(\zVecT) \Big)\ d\zVecT &\ =\ 0\\
\Rightarrow\ \ \ \mbox{For all $d$:}&& 
\int \Big( \mu_{d}(\zVecT; W)\  \frac{1}{ \sigT_d(\zVecT;M) } \ \gA_{d}(\zVecT) \ -\ \frac{1}{2}\,\frac{1}{ \sigT_d(\zVecT;M) } \ \gB_{d}(\zVecT) \Big)\ d\zVecT &\ =\ 0. \label{CondDeriPreFinal}
\end{align}
By summing expression (\ref{CondDeriPreFinal}) over $d$ we finally obtain:
\begin{align}
\sum_d
\int \Big( \mu_{d}(\zVecT; W)\  \frac{1}{ \sigT_d(\zVecT;M) } \ \gA_{d}(\zVecT) \ -\ \frac{1}{2}\,\frac{1}{ \sigT_d(\zVecT;M) } \ \gB_{d}(\zVecT) \Big)\ d\zVecT &\ =\ 0\\
\Rightarrow\ \ \  
\int \Big( \sum_d \mu_{d}(\zVecT; W)\  s_d(\zVecT;M) \ \gA_{d}(\zVecT) \ -\ \frac{1}{2}\,\sum_d\,
s_d(\zVecT;M) \ \gB_{d'}(\zVecT) \Big)\ d\zVecT &\ =\ 0\\
\Rightarrow\ \ \  
\int \Big( \big( \muVec(\zVecT; W)\,\odot\,\sVec(\zVecT;M) \big)^{\mathrm{T}} \ \gVecA(\zVecT) \ -\ \frac{1}{2}\, \big( \sVec(\zVecT;M) \big)^{\mathrm{T}} \ \gVecB_{d}(\zVecT) \Big)\ d\zVecT &\ =\ 0\\
\Rightarrow\ \ \  
\int \left(\begin{array}{c} \muVec(\zVecT; W) \odot \sVec(\zVecT;M) \\[1mm] -\frac{1}{2}\,\sVec(\zVecT;M) \end{array}\right)^{\mathrm{T}}\ \left(\begin{array}{c} \gVecA(\zVecT) \\[1mm] \gVecB(\zVecT) \end{array}\right)\ d\zVecT &\ =\ 0\\
\Rightarrow\ \ \  
\int \etaVec^{\mathrm{T}}_{(\zVecT;\Theta)}\ \gVec(\zVecT) \,\mathrm{d}\zVecT &\ =\ 0\,,
\label{CondDeriFinal}
\end{align}
where for the last step we have used the definition of $\etaVec_{(\zVecT;\Theta)}$ in \cref{EqnDNNLink}.

To summarize, we have for the VAE derived from expression (\ref{DeriCondInit}) the expression (\ref{CondDeriFinal}), i.e., we have shown that for the VAE generative model condition (\ref{EqnParaCondNoise}) holds. 

For a generative model that satisfies the parameterization conditions (\ref{EqnParaCondPrior}) and (\ref{EqnParaCondNoise}) it holds at all stationary points \citep[Theorem~1,][]{lucke2022convergence} that
\begin{align}
    \begin{split}
        \FF(\Phi,\Theta) =\; &\frac{1}{N}\sum_{n=1}^N \HH[\qPhi(\zVecT\given\xVecN)] - \HH[p_{\Amat}(\zVecT)] \,-\, \frac{1}{N} \sum_{n=1}^N \EEE{\qn_{\Phi}}{ \HH[p_{\Theta}(\xVec\,|\,\zVecT)] }\,. \label{DeriEntropiesA}
    \end{split}
\end{align}
The result notably applies even though we only used that derivatives w.r.t.\ $\Amat$ and w.r.t.\ $\thetaVec$ vanish where $\thetaVec$ only contains the output layer parameters of the decoder network $\mathrm{DNN}_{\sigT}(\zVecT; \Mall)$,  see Eqn. (\ref{ProofDNNsigmaApp}). In other words, we only required  
\begin{align}
\forall h:\ \ \del{\alphaT_h}\,\FF(\Phi,\Theta) \,=\, 0, 
\ \ \forall d,j:\ \ \del{M^L_{dj}}\,\FF(\Phi,\Theta) \,=\, 0,\ \ \mbox{and}\ \  
\ \ \forall d:\ \  \del{c^L_{d}}\,\FF(\Phi,\Theta) \,=\, 0  
\end{align}
to show equality of the ELBO to expression (\ref{DeriEntropiesA}).
Additional information about vanishing derivatives for the remaining parameters is not required. Remaining parameters can therefore take on any value (including values that correspond to stationary points, of course).
Hence, we can conclude that at all stationary points applies:
\begin{align}
    \begin{split}
        \FF(\Phi,\Theta) =\; &\frac{1}{N}\sum_{n=1}^N \HH[\qPhi(\zVecT\given\xVecN)] - \HH[p_{\Amat}(\zVecT)] \,-\, \frac{1}{N} \sum_{n=1}^N \EEE{\qn_{\Phi}}{ \HH[p_{\Theta}(\xVec\,|\,\zVecT)] } \\
 =\; &  \frac{1}{N}\sum_{n=1}^N \frac{1}{2} \log\!\big(\det( 2\pi e \tilde{\TT}_{\Phi}(\xVecN) \big) - \frac{1}{2} \log\!\big(\det(2\pi e \Amat ) \big) 
 - \frac{1}{N} \sum_{n=1}^N \EEE{\qn_{\Phi}}{ \sum_{d=1}^D \frac{1}{2} \log\!\big( 2\pi\, e\, \sigT_d(\zVecT; M) \big) } \,.        \label{DeriEntropiesApp}
    \end{split}
\end{align}
As the last step, we transform the result back to the original VAE--3 parameterization, which we do along the same lines as was done in the proof of \cref{theo:VAE-1}. Concretely, we
again express $\tilde{\TT}_{\Phi}(\xVec)$ in terms of $\TT_{\Phi}(\xVec)$. We drop subscript and argument of $\tilde{\TT}$ for readability, and note again that $\tilde{\TT}$ is the covariance matrix of a Gaussian distribution defined in the space of $\zVecT$. The random variable $\zVec$ depends according to the reparameterization \cref{Eqn:VAE4Prior} on $\zVecT$ via $\zVec=\Amat^{-\frac{1}{2}}\zVecT$. Consequently, if $\zVecT$ is Gaussian distributed with covariance $\tilde{\TT}$, then $\zVec$ is Gaussian distributed with
covariance $\TT=\Amat^{-\frac{1}{2}} \tilde{\TT} \big(\Amat^{-\frac{1}{2}}\big)^\mathrm{T}$. As all matrices are diagonal, we get $\tilde{\TT}=\Amat\TT$.
Inserting into \cref{DeriEntropiesA} we observe the first term to cancel with part of the last term:
\begin{align}
    \FF(\Phi,\Theta) =\; &  \frac{1}{N}\sum_{n=1}^N \frac{1}{2} \log\!\big(\det( 2\pi e \Amat \TT_{\Phi}(\xVecN) \big) - \frac{1}{2} \log\!\big(\det(2\pi e \Amat ) \big) 
     - \frac{1}{N} \sum_{n=1}^N \EEE{\qn_{\Phi}}{ \sum_{d=1}^D \frac{1}{2} \log\!\big( 2\pi\, e\, \sigT_d(\zVecT; M) \big) } \nonumber \\
     =\; &  \frac{1}{N}\sum_{n=1}^N \frac{1}{2} \log\!\big(\det( 2\pi e \TT_{\Phi}(\xVecN) \big) - \frac{1}{2} \log\!\big(\det(2\pi e ) \big) 
     - \frac{1}{N} \sum_{n=1}^N \EEE{\qn_{\Phi}}{ \frac{1}{2} \log\!\Big( \det\big( 2\pi\, e\, \Sigma_{\Theta}(\zVec) \big) \Big) } \nonumber \\
    =\; &\frac{1}{N}\sum_{n=1}^N \HH[\qPhi(\zVec\given\xVecN)] - \HH[p(\zVec)] \,-\, \frac{1}{N} \sum_{n=1}^N \EEE{\qn_{\Phi}}{ \HH[p_{\Theta}(\xVec\,|\,\zVec)] },
\label{DeriEntropiesAA}
\end{align}
which proofs the claim of \cref{theo:VAE-3}. \hfill $\square$\\
\ \\
\noindent{}As a comment on the proof: at first it may not sound intuitive that we can only consider stationary points of the subset $\thetaVec$ instead of all parameters. Do note, though, that the stationary point condition serves to show that the integral of expression (\ref{EqnFFThreeShort}) vanishes. The integrand contains all model parameters $\Theta$ within the natural parameters $\etaVec^{\mathrm{T}}_{(\zVecT;\Theta)}$. If we now just use a subset of parameters to construct the Jacobian of the left-hand-side expression of \cref{EqnParaCondNoise}, then such a subset is sufficient as long as we can conclude the right-hand-side of \cref{EqnParaCondNoise}. This is again in analogy to the explicit proof of \cref{theo:VAE-2} where the subset of parameters to show that $\FF_2(\Phi,\Theta)$ becomes equal to an entropy is just consisting of the single parameter $\sigma^2$.
In our case, we required all weights and biases of the output layer of $\mathrm{DNN}_{\sigma}(\zVec; \Mall)$ as subset (but none of the DNN's other parameters). Choosing a smaller subset would not have been sufficient. Already using fixed biases $\cVec^L$ would break the proof (unless all biases are fixed to zero). Larger subsets $\thetaVec$ would just have made the proof more intricate without changing the final results.

{\em Author Contributions:} We remark that \cref{theo:VAE-2} to \cref{theo:VAE-3} were hypothesized by JL who provided the first versions of the proofs with significant contributions by SD.

\section{ENTROPY-BASED ANALYSIS OF VAE OPTIMIZATION}
\label{app:Entropy-BasedAnalysis}

\subsection{Differential Entropies and Typical Sets}
\label{app:TypicalSet}

At all stationary points the ELBO decomposes into three entropies which implies that all model parameters determine the ELBO value through the same mathematical object: entropy. 
This novel observation give rise to an entropy-based interpretation of VAE optimization.

Let us briefly recall a potential interpretation of (differential) entropies in terms of the typical set \citep[e.g.,][]{mackay2003information,thomas2006elements}.
Consider sequences of i.i.d. samples from a continuous probability distribution $p(\xVec)$ over $\mathcal{X}$.
The typical set of this distribution $p(\xVec)$ in dependence of $N \in \mathbb{N}$ and $\epsilon > 0$ is then defined as
\begin{align}
    A^{(N)}_\epsilon = \Big\{(\xVec_1, \dots, \xVec_N) \in \mathcal{X}^N: 
    \Big\lvert - \frac{1}{N} \log(p(\xVec_1, \dots, \xVec_N) - \HH[p(\xVec)]\Big\rvert \le \epsilon \Big\} \enspace .
\end{align}
Recall that we have 
$\mathbb{P}\big(A^{(N)}_\epsilon\big) > 1 - \epsilon$
for large enough $N$, i.e., most sequences that occur in practice will belong to the typical set with high probability.
Moreover, the entropy of a distribution can be related to the volume of the typical set: %

\begin{equation}
    (1-\epsilon)2^{N(\HH[p(\xVec)]- \epsilon)} \le \mathrm{Vol}(A^{(N)}_\epsilon) \le 2^{N(\HH[p(\xVec)] + \epsilon)} \enspace.
\end{equation}

The second inequality holds for all $N$, and the first for sufficiently large $N$ \citep[Theorem 8.2.2]{thomas2006elements}.
Thus, the volume of the typical set $A^{(N)}_\epsilon$ is characterized by the differential entropy (besides the growth in sequence length $N$). Considering the entropies of discrete random variables the volume of the typical set translates to its cardinality.
To conclude, continuous random variables with small entropy are ``confined to a small effective volume'', while random variables with high entropy are ``widely dispersed''\citep{thomas2006elements}.

With this interpretation in mind, \cref{theo:VAE-1,theo:VAE-3} allow to re-interpret VAE learning. 
First, recall that ELBO maximization can be expressed in terms of reconstruction score $S_\mathrm{rec}$ and regularization score $ S_\mathrm{reg}$, i.e., $\FF = S_\mathrm{rec} + S_\mathrm{reg}$ (see \cref{EqnELBO}). 
We are confronted with two conflicting goals here as with $S_\mathrm{reg}$ too large, i.e., $\DKL{\qPhi(\zVec \given \xVec)}{p(\zVec)}$ close to zero, no successful reconstruction (high $S_\mathrm{rec}$) can be expected. %
Given \cref{theo:VAE-1,theo:VAE-3} we can now expressed both scores based on entropies as in \cref{eq:Scores_Reg_Rec}%
\begin{align*}
\begin{split}
    S_{\mathrm{reg}}(\Phi,\Theta) &= \frac{1}{N}\sum_{n} \HH[\qPhi(\zVec\given\xVecN)] \,-\, \HH[\pT(\zVec)] \enspace, \\ %
    S_{\mathrm{rec}}(\Theta) &= -\,\HH[\pT(\xVecN\,|\,\zVec)] 
\end{split}
\end{align*}
in which the entropy $\HH[\pT(\xVecN\,|\,\zVec)]$ is replaced by the expected entropy in the case of VAE--3. %

Optimization of the decoder therefore seeks to \textit{decrease} its entropy $\HH[\pT(\xVec\given\zVec)]$ (or its expectation in case of VAE--3), which corresponds to reducing the volume of the typical set, the effective volume of the decoding distribution in data space $\mathcal{X}$.
As discussed in \cref{SecNum}
the decoder entropy naturally resembles the reconstruction performance of the model (see also \cref{EqnAlphaSigmaSolutions} that makes this relationship explicit).
Thus, a small volume of the decoder's typical set corresponds to a low reconstruction error/uncertainty (or equivalently a high reconstruction score $S_{\mathrm{rec}}(\Theta)$).

On the other hand, the average encoder entropy $\frac{1}{N}\sum_n\HH[\qPhi(\zVec\given\xVecN)]$ is maximized. 
Hence, the (logarithm of the) volume of the typical set in latent space $\mathcal{Z}$ should be increased.
Intuitively, this corresponds to a larger variety in the encoding distribution (per data sample $\xVecN$).
Note that the prior entropy $\HH[p(\zVec)]$ represents an upper bound on the average encoder entropy. 
Accordingly, the regularization score $S_{\mathrm{reg}}(\Phi,\Theta)$ captures (and minimizes) the difference between the (log-)volumes of typical sets of prior and encoder distribution. 
With too widely dispersed latent representations $\zVec$, i.e., when $\frac{1}{N}\sum_{n} \HH[\qPhi(\zVec\given\xVecN)]$ is too close to $\HH[\pT(\zVec)]$, no successful reconstruction can be demanded.
Overall, VAE optimization translates to increase of encoder and decrease of decoder entropy, which in turn reflect the increase or decrease of the volumes of the corresponding typical sets. And notably, central figures for monitoring VAEs are easily accessible solely based on entropies (and even more reliably as their conventional counter-parts, as demonstrated in \cref{SecNum}).

\subsection{Posterior Collapse}
\label{app:PosteriorCollapse}

\cref{theo:VAE-1,theo:VAE-3} represent by themselves theoretical results. 
\Cref{SecNum} discussed some
practical applicability in terms of entropy-based definitions for regularization and reconstruction scores as well
as for posterior collapse. 
We here first derive and elaborate on the criterion (\cref{EqnPostCollCriterion}) used in \cref{SecNum} to measure the percentage of collapsed latents in practice. The accompanying experimental results are depicted in \cref{fig:PosteriorCollapse_App} in \cref{app:PosteriorCollapse_Exp}.
Our entropy-based results can, however, also be used to further study VAE optimization theoretically. Below we, therefore, secondly discuss optimization and turning points for learning that can be described using entropies, and we will finally come back to posterior collapse from this theoretical perspective.

Quantifying posterior collapse has been of interest previously (see, e.g., \cite{BowmanEtAl2016,HeEtAl2018,DiengEtAl2019}).
The contribution by \citet[][]{LucasEtAl2019} concretely and systematically investigates the effect and its dependence on fixed decoder variance $\sigma^2$. 
The paper first argues for a quantification of posterior collapse in terms of percentage of collapsed latents, and then introduces an $(\epsilon,\delta)$-measure based on the KL-divergence \citep[][p.\,7]{LucasEtAl2019} between encoder and prior distribution (for individual latents $h$). 
Values of the two thresholds $\epsilon$ and $\delta$ are hand-set, and a latent dimensions $h$ is defined to be collapsed whenever
\begin{equation}
    \label{eq:PostCollapse_Lucas}
    \mathbb{P}_{\xVec \sim p_\mathrm{data}}\left[ \DKL{\qPhiEnc(z_h \vert \xVec)}{\pTPrior(z_h)} < \epsilon \right] \le 1 - \delta \enspace.
\end{equation}
The measure is then used as a core VAE analysis tool monitoring individual latents or the fraction of collapsed latent dimensions.

Based on the convergence to entropies results, an alternative measure for posterior collapse offers itself if we consider the entropy-based regularization measure $S_{\mathrm{reg}}(\Phi,\Theta)$ of Eqn.\,(\ref{eq:Scores_Reg_Rec}).
Posterior collapsed latents increase the regularization score while they usually have negligible effects on the reconstruction score. 
Because of the entropy-based formulation of the regularization score, it is now very straight-forward to break down the score into a sum over latents.
Using the conventional assumption of Gaussian encoder and prior distributions both with diagonal covariance matrices, we obtain:
\begin{align}
 S_{\mathrm{reg}}(\Phi,\Theta)
&=   \Big(  \frac{1}{N}\sum_{n=1}^N \HH[\qPhi(\zVec\given\xVecN)] \,-\, \HH[\pT(\zVec)] \Big)\nonumber\\ 
&=   \Big(  \frac{1}{N}\sum_{n=1}^N \HH[\prod_{h=1}^H\qPhi(z_h\given\xVecN)] \,-\, \HH[\prod_{h=1}^H\pT(z_h)] \Big)\nonumber\\
&=   \sum_{h=1}^H \Big(  \frac{1}{N}\sum_{n=1}^N \HH[\qPhi(z_h\given\xVecN)] \,-\, \HH[\pT(z_h)] \Big)\nonumber\\ 
&= \sum_{h=1}^H \Big( \frac{1}{N}\sum_{n=1}^N \HH[\qPhi(z_h\given\xVecN)] \,-\, \frac{1}{2}\log(2\pi{}e) \Big) \enspace .
\label{eq:PostCollapse_Entropy_App}
\end{align}
As $S_{\mathrm{reg}}(\Phi,\Theta)$ becomes equal to the conventional regularization measure 
\begin{align}
S^{\mathrm{conv}}_{\mathrm{reg}}(\Phi,\Theta)=- \frac{1}{N}\sum_{n} \DKL{\qn_{\Phi}\!(\zVec)}{\pT(\zVec)}
\end{align}
at stationary points, $S_{\mathrm{reg}}(\Phi,\Theta)$ can at stationary points never be positive. It can, furthermore, be shown for VAE--1 that also each summand of Eqn.\,(\ref{eq:PostCollapse_Entropy_App}) (i.e., the regularization score per latent $h$) can never be positive
at stationary points (we discuss further below). In turn, this means that there is a clearly defined highest possible value the encoder entropy (per latent variable $h$) can converge to, concretely
\begin{align}
\frac{1}{N}\sum_{n=1}^N \HH[\qPhi(z_h\given\xVecN)]
&\leq{}\frac{1}{2}\log(2\pi{}e) \approx 1.42 \enspace.
\end{align}
Values of the encoder entropy close to the prior entropy (for standard normal prior dimensions a value of $1.42$) can consequently be used to identify posterior collapsed latents, which motivates the definition presented in \cref{EqnPostCollCriterion}.

Notably, we do not claim that no other sensible measures for posterior collapse can be defined. For instance, the measure (\cref{eq:PostCollapse_Lucas}) represents a perfectly valid definition. Also the decomposition into sums over latents can be done using the original KL-divergence\footnote{This is how one can show that encoder entropy of an individual latent is upper-bounded by $\frac{1}{2}\log(2\pi{}e)$.}.
We would argue, however, that the previous measure (\cref{eq:PostCollapse_Lucas}) is more ad hoc and requires two hand-set parameters $\epsilon$ and $\delta$ for the threshold. The measure (\cref{EqnPostCollCriterion})
may consequently be perceived as more natural, and may be a better starting point to define similar measures also for non-Gaussian VAEs or VAEs with learnable priors.
Furthermore, the entropy-based definitions may be perceived as being easier to use, and the derivation (\cref{eq:PostCollapse_Entropy_App}) may serve as an example. 

We like to remark that we \emph{do not} require the conditions of \cref{theo:VAE-1,theo:VAE-3} to be fulfilled in order to apply the entropy-based criterion given in \cref{EqnPostCollCriterion}.
In fact, the definition is largely independent of the decoder architecture.  
In particular, the entropy-based measurement of posterior collapse is also valid and meaningful when the decoder (co-)variance is fixed, e.g., to $\sigma = 1$ in case of VAE--1, as commonly done in practice. See also the accompanying experiments in \cref{app:PosteriorCollapse_Exp} where we include this scenario (\cref{fig:PosteriorCollapse_App}(c)).

Importantly, the entropy-based measures (e.g., for ELBO, regularization or reconstruction) are more than just reformulations of the conventional definitions. There are qualitative differences. The regularization score does, for instance, only depend on the encoder variances: it contains neither dependencies on decoder parameters (the priors only formally depend on $\Theta$) nor does it depend on the encoder means. The latter is in contrast to the KL-divergence, which does contain the encoder means (for example the measure in \cite{LucasEtAl2019} given in \cref{eq:PostCollapse_Lucas}). Similarly (and more drastically), the entropy-based reconstruction score for VAE--1 exclusively depends on one single decoder parameter, the decoder variance. In contrast, the conventionally defined decoder variance does depend on decoder and encoder parameters. In turn, this means that at stationary points dependencies on all parameters except of the decoder variance necessarily have to vanish for the conventional reconstruction score. 

For completeness, the treatment of VAEs of type VAE--3 would be different but similar. For instance, the reconstruction score would (based on \cref{theo:VAE-3}) now be defined as:
\begin{align}
  S_{\mathrm{rec}}(\Theta) =  - \frac{1}{N} \sum_{n=1}^N \EEE{\qn_{\Phi}}{ \HH[\pT(\xVec\,|\,\zVec)] }\,,
\end{align}
and consequently depends on encoder and decoder parameters. The definition of the regularization score remains unchanged, however.

\subsection{Optimization landscape} 
\label{app:OptimizationLandscape}
\begin{figure*}[ht!]
\begin{center}
\includegraphics[width=0.8\linewidth]{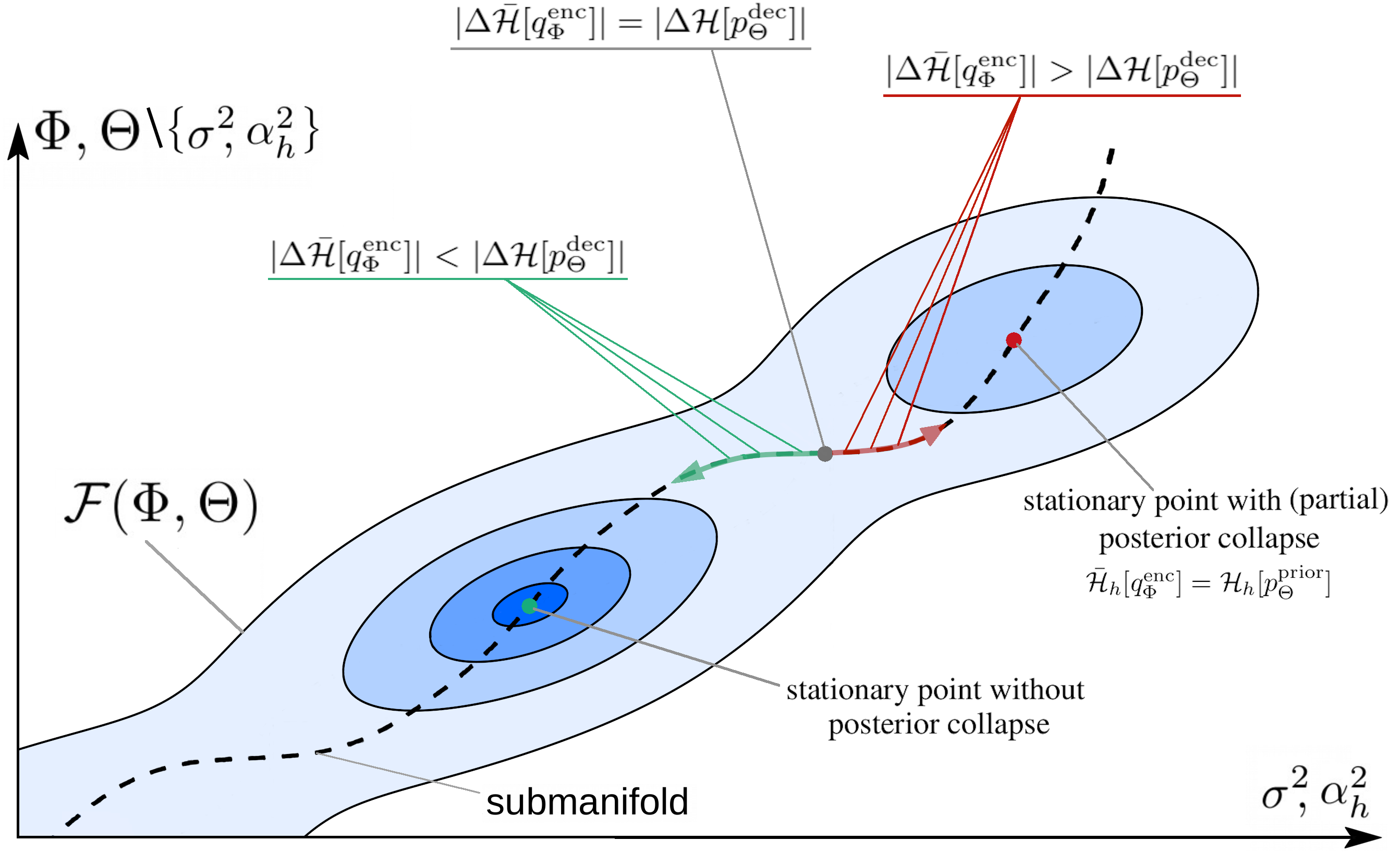}
\end{center}
\vspace{-10pt}
\caption{Visualization of the variational lower bound and its relation to the three entropies expression.
The figure shows a two dimensional visualization with the following axes: the x-axis represents the hyperplane
of variance parameters $\alpha^2_1$ to $\alpha^2_H$ together with decoder variance $\sigma^2$. We assume a VAE
of type VAE--1 for the figure. The $\alpha^2_1$ to $\alpha^2_H$ we take to be implicitly defined by the decoder
weights (compare VAE--2). The dotted black line represents a submanifold in which the parameters of the x-axis have converged.
Within the submanifold, the variational lower bound is equal to a sum of three entropies (\cref{theo:VAE-1}). In the illustrated
example, the submanifold connects a stationary point without posterior collapse to a stationary point with (partial)
posterior collapse.
Depending on the location on the manifold, learning is dominated by the change in the reconstruction score $S_{\mathrm{rec}}(\Theta)$ (green arrow) or dominated by the change in the regularization score $S_{\mathrm{reg}}(\Phi,\Theta)$ (red arrow), with qualitatively different outcomes. As both scores can be defined based on entropies (see \cref{eq:Scores_Reg_Rec}), changes of the scores directly translate to changes in entropies such that the optimization landscapes can be characterized using changes in entropies. Let us denote by $|\Delta{}\HH[\pTDec]|$ the absolute change of the decoder entropy, while we denote by $|\Delta{}\bar{\HH}[\qPhiEnc]|$ the absolute average change of the encoder entropy, i.e., $\bar{\HH}[\qPhiEnc]=\frac{1}{N}\sum_n{}\HH[\qPhiEnc(\zVec\given\xVecN)]$.      
If we start at the high local optimum and traverse the submanifold from left to right then the reconstruction score
$S_{\mathrm{rec}}(\Theta)$ decreases while the regularization score $S_{\mathrm{reg}}(\Phi,\Theta)$ will tend to increase (to a lesser extent). 
Translated to entropies, we have within the submanifold between high maximum and saddle point $|\Delta{}\bar{\HH}[\qPhiEnc]|<|\Delta{}\HH[\pTDec]|$, and ELBO optimization would favor reconstruction improvements. At the saddle point (the turning point), the changes of the entropies become equal. To the right of the turning point, learning is dominated by the regularization term, which means that the change
in encoder entropy is dominant $|\Delta{}\bar{\HH}[\qPhiEnc]|>|\Delta{}\HH[\pTDec]|$. ELBO optimization in this part of the manifold would result in (partial) posterior collapse.
$|\Delta{}\bar{\HH}[\qPhiEnc]|=|\Delta{}\HH[\pTDec]|$
At the local optimum with (partly) collapsed
posterior, the encoder entropy, $\bar{\HH}_h[\qPhiEnc]$, of a collapsed latent $h$ will be equal to the prior entropy.
}
\label{fig:App}
\vspace{0pt}
\end{figure*}

To elaborate on an application of the results for the theoretical analysis
of VAE optimization, consider first VAEs of type VAE--1 (\cref{def:VAE-1}).
For such VAEs the closed-form bound of \cref{theo:VAE-1} applies at all stationary points. The result was notably derived using
properties of two types of variance parameters: decoder variance and prior variance. Importantly, the prior variance is in the usual parametrization of VAEs not part of the prior but part of the decoder DNN, i.e., the $\alpha_h$
are part of the first DNN layer (see VAE--2 for an explicit encoding with prior variance). While Eqn.\,(\ref{PropTwoA}) of \cref{theo:VAE-1} does by itself not describe an optimization landscape,
we can recover a description of an optimization landscape if we insert solutions for $\alpha^2_h$ for $\sigma^2$
into Eqn.\,(\ref{PropTwoA}). Solutions for $\alpha_h$ and $\sigma^2$ are given by the following
explicit functions:
\begin{align}
 \alpha^2_h &= \frac{1}{N}\sum_{n=1}^N \int \qPhi(\zVec\given\xVecN) \, z_h^2\, d\zVec\,,\phantom{xx}   
 \sigma^2 \,=\, \frac{1}{DN}\sum_{n=1}^N \int \qPhi(\zVec\given\xVecN)\, \| \xVecN - \muVec_{\Theta}(\zVec) \|^2\, d\zVec,  \label{EqnAlphaSigmaSolutions}
\end{align}
and the $\alpha^2_h$ expression further simplifies for diagonal covariances.
After the solutions for $\alpha_h$ and $\sigma^2$ are inserted into Eqn.\,(\ref{PropTwoA}), the resulting expression
describes the optimization landscape within submanifolds of the parameter space (\cref{fig:App} shows an illustration).
The description of the optimization landscape within submanifolds can be used to study the optimization landscape of the whole parameter space. Using this view, we further below relate back to the above discussed posterior collapse.

Similar approaches can be used for more general VAEs but an analysis necessarily becomes more intricate, e.g., because of the dependence of decoder variances on the latents for VAEs of type VAE--3 (but note the similar properties for $\alpha_h$).

\subsection{Theoretical Analysis of Posterior Collapse}

To which extend posterior collapse is problematic (e.g., neglecting large parts of the models capacity) or not harmful (or even beneficial) is still discussed in the community \citep[see, e.g.,][]{asperti2020variance}.
It is important for our study to observe that the ``mathematical mechanism of this phenomenon is not well understood'' \citep[as, e.g., stated by][]{LucasEtAl2019}. 
Hence, the phenomenon requires additional analysis and this is where the contribution of the presented entropy perspective (provided by \cref{theo:VAE-1,theo:VAE-3}) can also go beyond the practical aspects of the introduced entropy-based measures. Concretely, the theorems allow a contribution to the question {\em why} some latent variables sometimes cease to participate in encoding/decoding and instead do assume a high variance.

Considering \cref{theo:VAE-1} and the entropy-based measures for regularization score and
reconstruction score, we can investigate the optimization of the bound $\FF(\Phi,\Theta)$ a bit more systematically. The bound is finally given
by the sum of the two scores:
\begin{align}
  \FF(\Phi,\Theta) &= S_{\mathrm{reg}}(\Phi,\Theta) + S_{\mathrm{rec}}(\Theta)
    \label{EqnAppPostColl}
\end{align}
By changing the parameters, there are essentially two ways to obtain a higher bound:
(1)~One can increase the reconstruction score $S_{\mathrm{rec}}(\Theta)$ without too much decreasing the regularization score $S_{\mathrm{reg}}(\Phi,\Theta)$. %
(2)~One can (visa versa) obtain a large bound with an increased regularization score $S_{\mathrm{reg}}(\Phi,\Theta)$ as long as the reconstruction score is not decrease too much.

The considerations may (at first sight) only serve for comparisons of ELBO values
at different stationary points. However, if we combine the expression (\ref{EqnAppPostColl}) with our discussion of
the optimization landscape above, we can furthermore obtain information about
a continuous change of the bound within the above discussed submanifold of parameter space. By inserting\footnote{A VAE--1 can be regarded as being reparameterized as VAE--2 (see \cref{theo:VAE-1}). In the conventional VAE--1 form, the converged $\alpha_h$ can be taken as implicit in the decoder weights.} the expressions (\ref{EqnAlphaSigmaSolutions}) into (\cref{EqnAppPostColl}), we obtain a description of the optimization
landscape within a submanifold in parameter space (see dotted line of \cref{fig:App}), i.e., within the submanifold the three entropy expression
becomes identical to the original bound.  
The submanifold in the figure connects an optimum without posterior collapsed latents
to an optimum with (partial) posterior collapse. When traversing from one optimum to the other along the submanifold, a saddle point will be reached. On the one side of the saddle point, gradient learning does increase the reconstruction score $S_{\mathrm{rec}}(\Theta)$ (as is usually desired for representation learning; green arrow in \cref{fig:App}); on the other side of the saddle point, the regularization score $S_{\mathrm{reg}}(\Phi,\Theta)$ is increased (the posterior of the latent starts to collapse; red arrow in \cref{fig:App}). 
Regularization increase of a single latent directly corresponds to that latent's increase to high entropy, removing the latent's influence on decoding is then a plausible consequence. Otherwise, the increasingly high entropy would increase the decoder entropy and, therefore, would negatively effect the bound.

The discussion of optimization landscapes adds a further example for the utility of entropy-based descriptions. In this case, changes of entropies $\Delta{}\HH$ naturally characterize saddle points in the optimization landscape (see \cref{fig:App}).

\section{EXPERIMENTAL SET-UP AND RESULTS}
\label{app:Experiments}
We here provide further details on the numerical experiments of \Cref{SecNum}.
An example implementation can be found at the end of this section (\cref{app:Code}).
The code to reproduce the experiments is available at \href{https://github.com/Learning-with-Entropies/ELBO-Entropies-AISTATS23}{github.com/Learning-with-Entropies}.

This section is organized as follows. In \cref{app:DiscVerification}  we elaborate on the verification of the main result itself, invoking Linear VAE, VAE--1 and VAE--3 (i.e., \cref{fig:Verification,fig:Verification_App}).
We continue with two proposed applications of the entropy results for non-linear VAEs: The entropy-based estimation of the ELBO in \cref{app:ELBOEstimation} and the results of the entropy-based posterior collapse analysis in \cref{app:PosteriorCollapse_Exp}.
We then provide details on the experimental setup including the data sets considered as well as the used network architectures in \cref{app:ExpSpecifications,app:ExpArtificailPCAdata}.
Lastly, we provide further insight on the stochasticity and small remaining gaps between ELBO and the sum of entropies in \cref{app:ExpNoise}.

\begin{figure*}[t]
\begin{center}
    \includegraphics[width=0.33\linewidth]{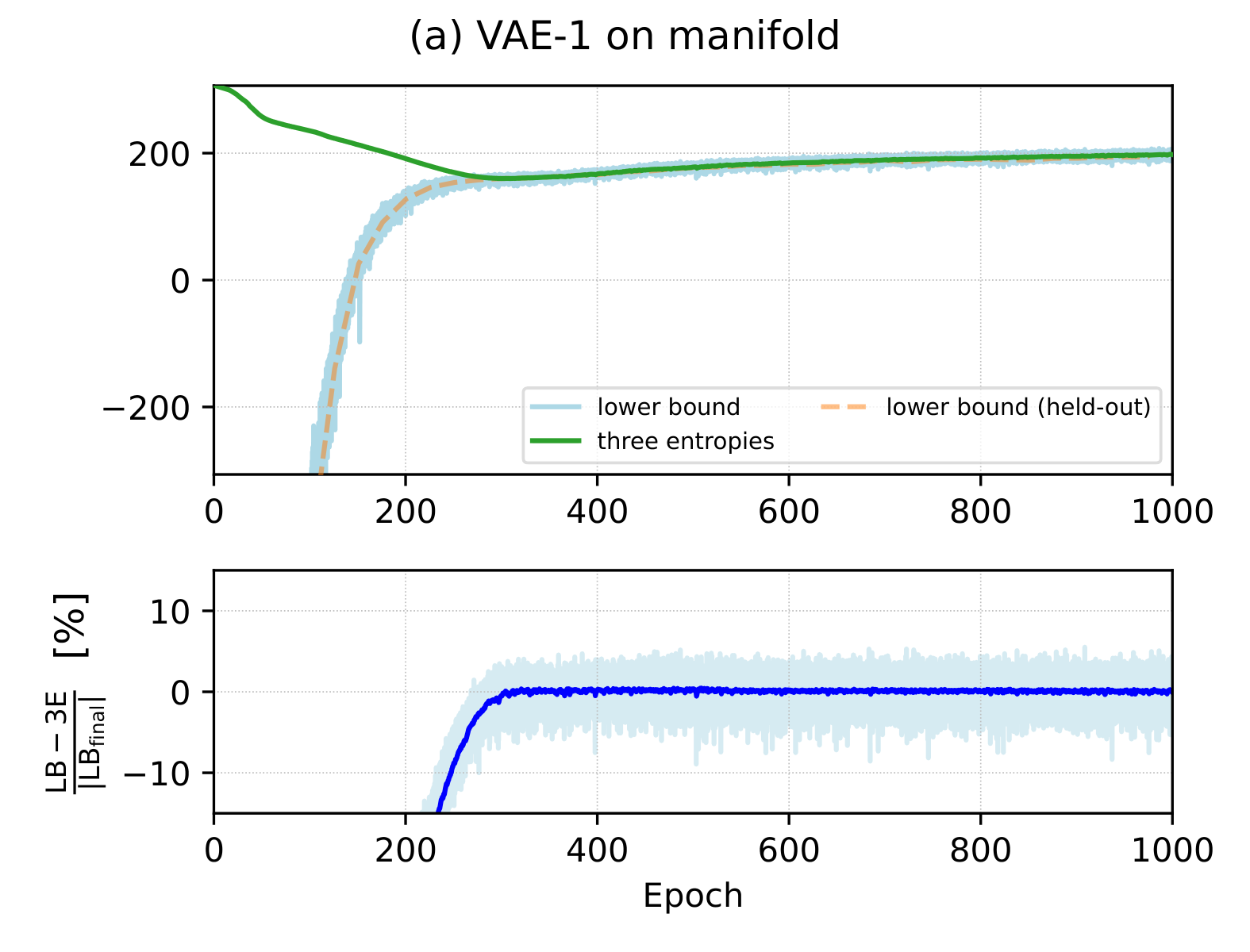}
    \includegraphics[width=0.33\linewidth]{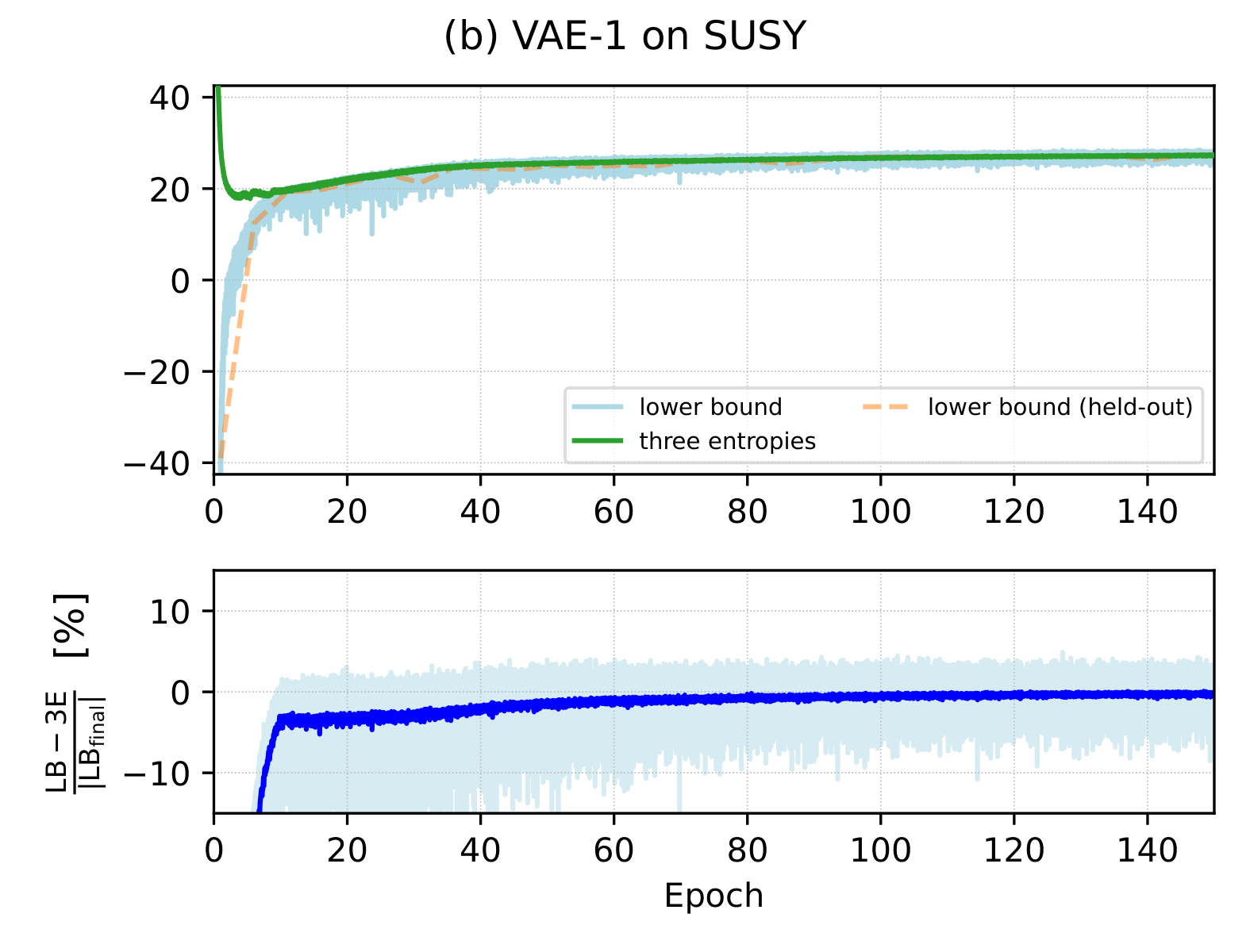}
    \includegraphics[width=0.33\linewidth]{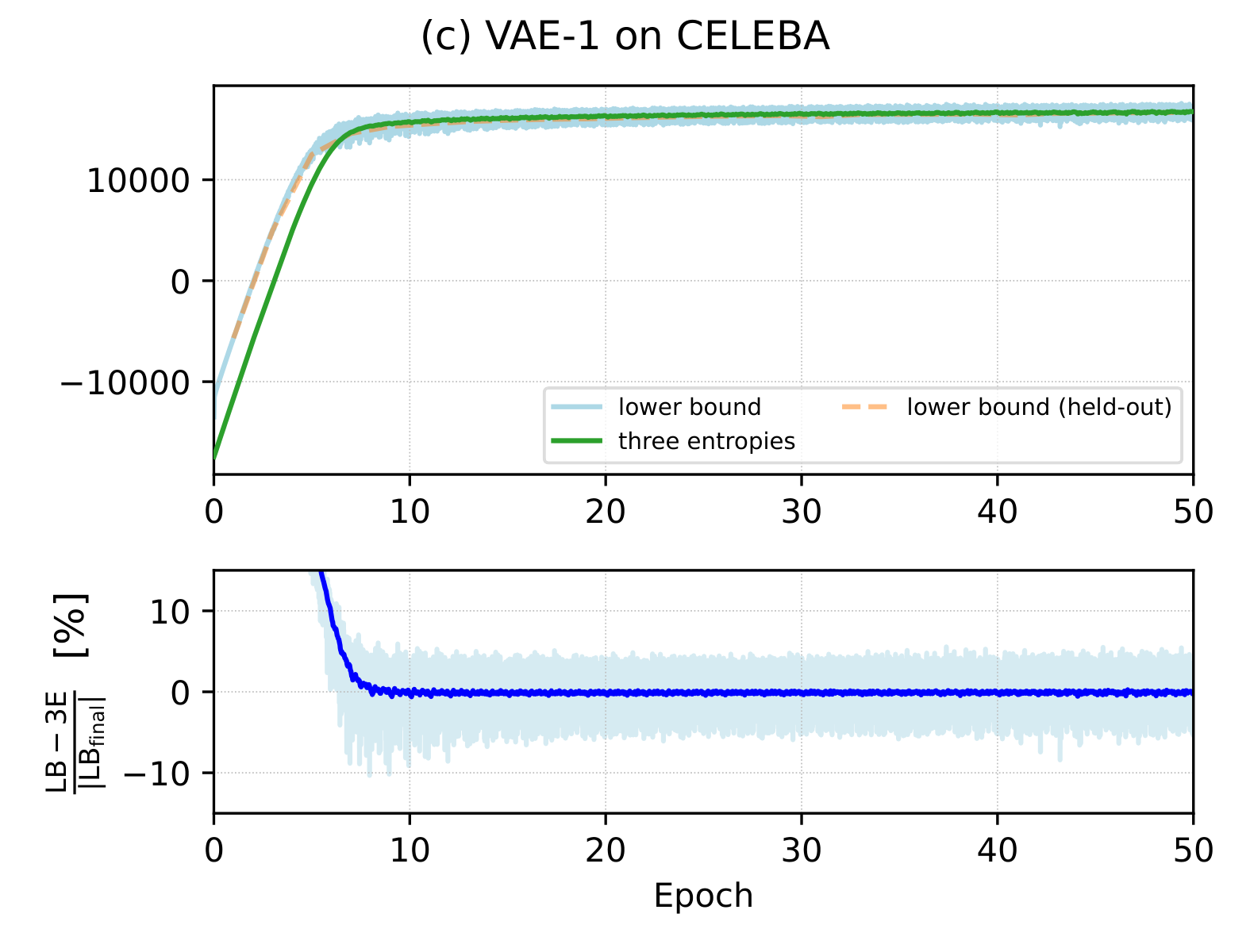}
\end{center}
\caption{\textbf{Additional (verification) experiments} on artificial manifold data, SUSY and CelebA with VAE--1 models. Depicted are the lower bound (ELBO), the lower bound on held-out-test data and the sum of entropies. The lower plot depicts the relative difference with an exponential moving average in dark blue. We remark the the ELBO fluctuates around the three entropies expression.
See \cref{app:ExpSpecifications} for further details.
}
\label{fig:Verification_App}
\end{figure*}

\subsection{Verification (\texorpdfstring{\cref{fig:Verification,fig:Verification_App}}{Figs. 1 and 5})}
\label{app:DiscVerification}

\paragraph{Linear VAEs} The linear VAE given by \cref{EqnVAElinearDecoder,EqnVAElinearEncoder} allows for the most direct investigation of the result of \cref{theo:VAE-1}.
It has the advantage that we know the optimal solution in this case: it is given by the well-known maximum likelihood solution of p-PCA \citep[][]{TippingBishop1999,Roweis1998}.
For the experiments we therefore first generate data according to the p-PCA generative model (details in \cref{app:ExpArtificailPCAdata}).
In \cref{fig:Verification}(a), top plot, we then compare the three entropies of \cref{EqnVAEBoundLinear} with the standard lower bound~(\cref{EqnELBO}), estimated by sampling, and the known exact log-likelihood solution (\cref{app:LinearVAEs}, \cref{EqnVAEBoundLinearC}) for the same set of data points.
For verification purposes, we additionally show the ground-truth log-likelihood for the generative parameters, as well as the model log-likelihood on held-out data.

Following the proof of \cref{theo:VAE-1}, the lower bound~(\cref{EqnELBO}) and the three entropies~(\cref{EqnVAEBoundLinear}) only have to be identical at stationary points of the variance parameters.
This suggests that e.g., for fixed $\sigma^2$ we can not expect the lower bound to converge to the three entropies.
This is shown by the dashed lines in \cref{fig:Verification}(a), top plot, which shows two examples with $\sigma^2$ fixed to two sub-optimal values.

\paragraph{Standard VAEs}
We demonstrate the validity of the theoretical result on MNIST, CelebA and SUSY using VAE--1 (see \cref{fig:Verification} in the paper itself and \cref{fig:Verification_App} for the latter two).
For verification of \cref{theo:VAE-3}, we also show experiments with \mbox{VAE--3} in \cref{fig:Verification}(c), where we used the same PCA data as for the linear VAE but introduced an additional non-linear transformation on the data, projecting it onto a ring-like structure in the $D$-dimensional space (details in \cref{app:ExpArtificailPCAdata}).
Since no ground-truth log-likelihood is available for these models, we only show the standard lower bound~(\ref{EqnELBO}), estimated by sampling, and compare it to the three entropies~(\ref{PropTwoA}) and~(\ref{PropThree}), respectively.
More details and visualizations are provided at the end of this section.
Additionally, we considered more practically relevant and large data sets, namely SUSY and CelebA.

\paragraph{Experimental Results: Verification} As can be observed, the sampled lower bound converges towards the sum of three entropies when the variance parameters are converging towards stationary points (which usually happens within the first epochs).
In \cref{fig:Verification}(a), top, we also see that both the lower bound and the three entropies converge to the log-likelihood (with the tight lower bound mostly invisible under the log-likelihood) and recover the ground-truth well. Regarding \cref{fig:Verification_App}, we used the held out computation of the lower-bound to ensure the absence of over-fitting.

Notably, the gap between lower bound and three entropies might become small even before all parameters converged, as seen, e.g., in \cref{fig:Verification}(c), since only the gradients of the variances have to vanish for the results to hold. Consequently, we observe high approximation qualities also distant from convergence points including, e.g., at saddle points or saddle surfaces, which commonly occur for DNN optimization.

The bottom plots in \cref{fig:Verification} show the absolute difference between the sampled lower bound and the three entropies relative to the final sampled lower bound (in percent).
Shown is the median over 10~runs for \cref{fig:Verification}(a) and~(b) and over 100~runs for \cref{fig:Verification}(c) with the interquartile range as shaded area, i.e., 50\% of the deviations fall within this area.
At convergence, the difference between lower bound and three entropies consistently approaches zero.
Remaining gaps we can account primarily to fluctuations caused by finite learning rates, batch sizes and numbers of samples.
E.g., for VAE--3, $\sigma_d(\mathbf{z};\Theta)$ has to be approximated via sampling, and we confirmed that the small gap of $<\!\!0.5\%$ in \cref{fig:Verification}(c), bottom, is in the order of magnitude of the sampling noise of $\sigma_d(\mathbf{z};\Theta)$, which makes the three entropies fluctuate around the lower bound (see also \cref{fig:visualization}).

There are some subtleties to pay attention to in the case of practical experiments for VAE--3.
We have assumed non-degenerated values for the decoder variance
$\mathrm{DNN}_{\sigma}(\zVec; \Mall)$ (\cref{EqnVAEThreeDecoderSigmaDNN})
at stationary points, for instance. However, during learning and due to stochasticity and finite learning rates, close to zero or negative values for $\sigT_d(\zVec; \Theta)$
can and do sometimes occur
during learning. To explicitly exclude such degeneracies in transient behavior, non-negativity can be ensured using constraints. In practice, we, for instance, used ReLU units with a small offset for the output layer or softplus activation. For our numerical results near stationary points, we then made sure that the activation functions of observables do operate in a linear regime after learning has converged, which is then again consistent with the conditions assumed for the variance DNNs (\cref{EqnVAEThreeDecoderSigmaDNN}). Concretely, we measured in experiments the percentage of variance DNN output units where enforcing non-negativity remained necessary. When the variance DNNs converged to stationary points, the percentage of non-linear units converged to zero.

\paragraph{Experimental Details of \texorpdfstring{\cref{fig:Verification}}{}}
Regarding these experiments we used a batch size of 2000, learning rates of $10^{-3}$ and 100~samples (with the exception of \cref{fig:noise}(b) and (c), as stated).
These values were roughly chosen to give results with little fluctuations in the sampled lower bound and three entropies within reasonable time.
The encoder and decoder DNNs for VAE--1 and VAE--3 were built with two hidden layers of 50~hidden units each and ReLU activations.

\subsection{Entropy-based ELBO estimation}
\label{app:ELBOEstimation}
While the verification experiments discussed in the former section (regarding \cref{fig:Verification}) focus on demonstrating the theoretical result itself, thereby trying to reduce the noise-level (see also \cref{app:ExpNoise}), \cref{fig:Verification_App} demonstrates a more practical perspective as outlined in \cref{SecNum}: For the prominent VAE--1 the lower bound fluctuates around the sum of entropies after convergence. Thus, a first handy utility %
is straight-forward entropy-based computation of ELBO values, which are routinely used to monitor and compare models on the training set.

\begin{figure}[ht!]
\begin{center}
    \includegraphics[width=0.45\linewidth]{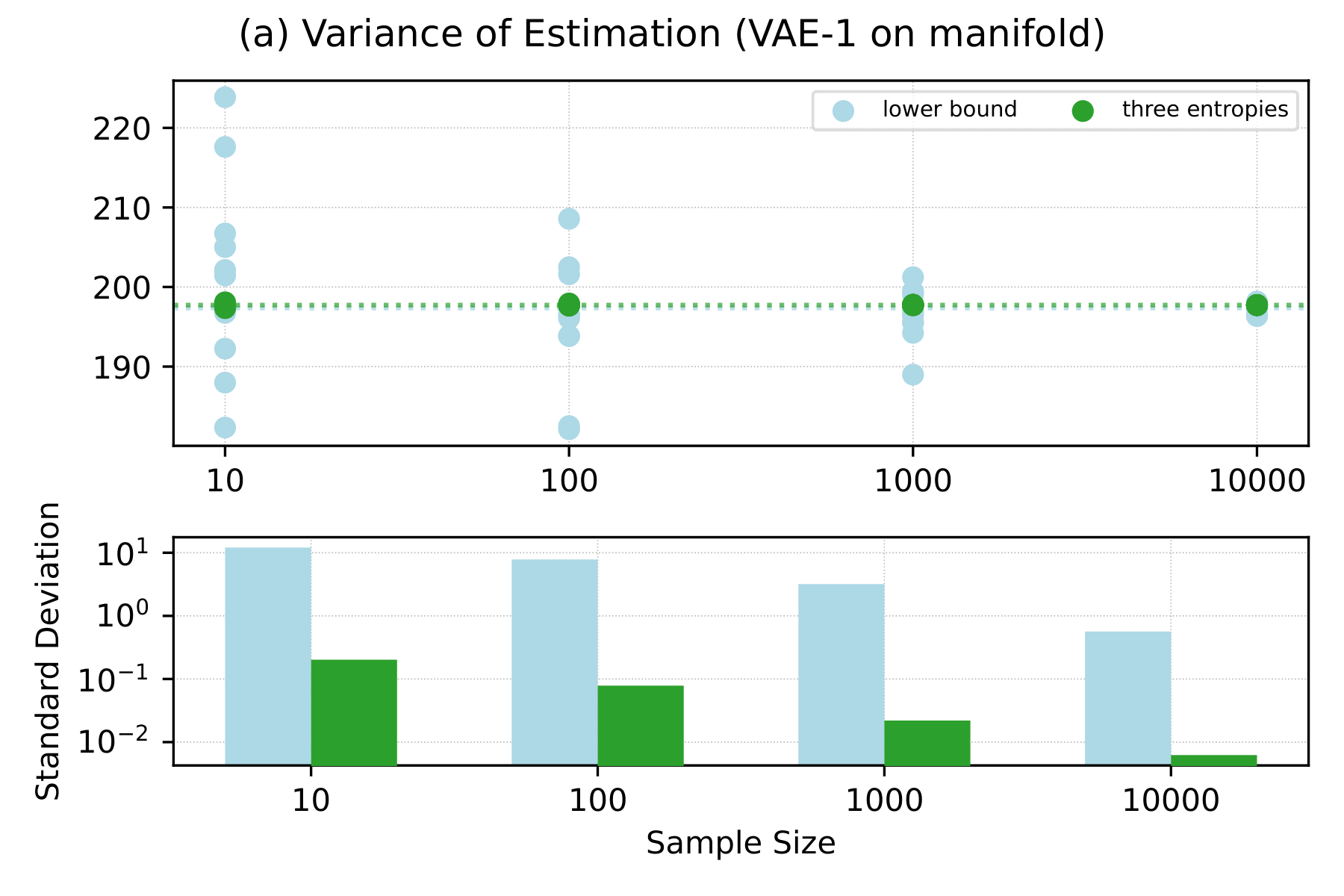}
    \includegraphics[width=0.45\linewidth]{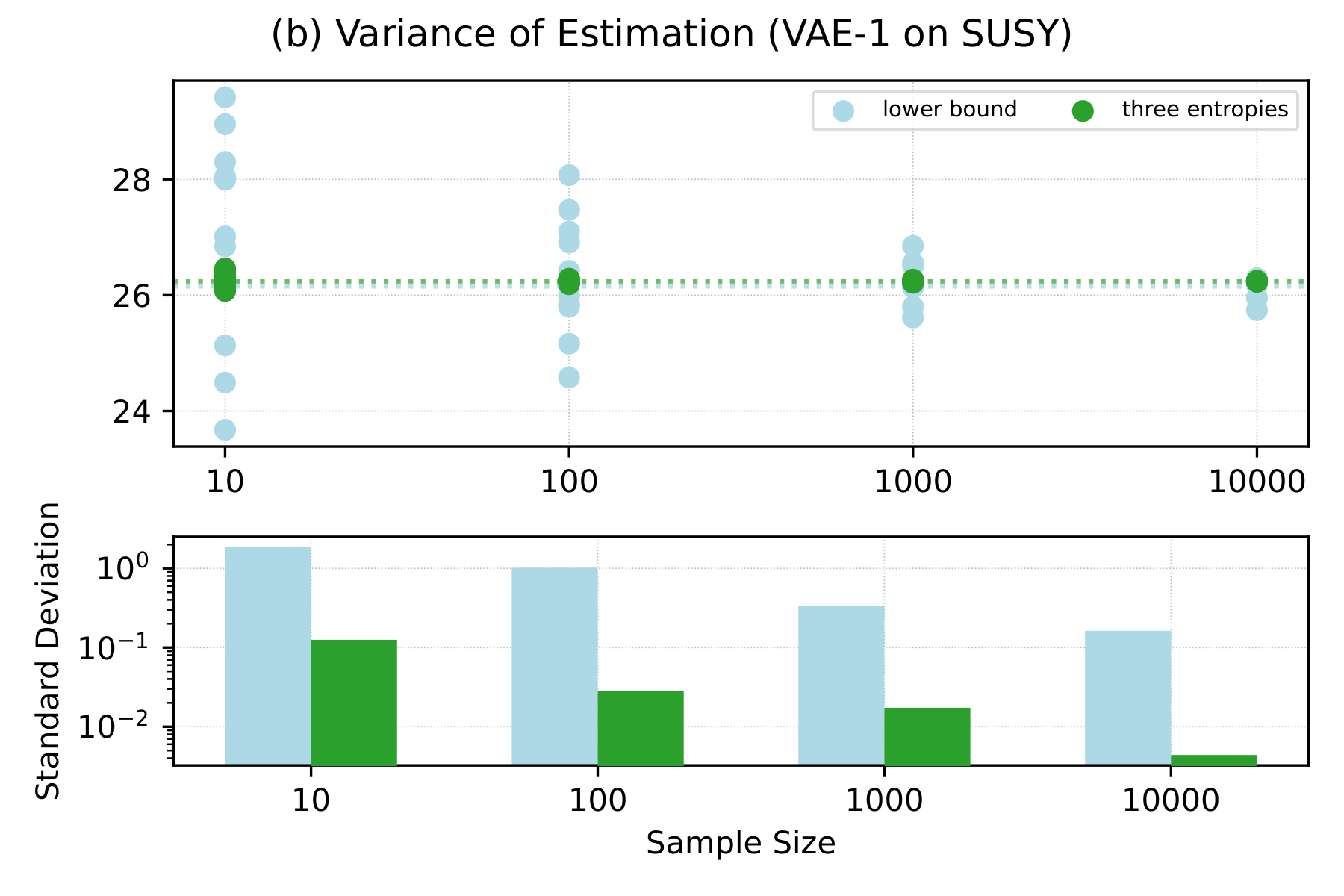}
\end{center}
\vspace{-12pt}
\caption{\textbf{ELBO Estimation} for VAE--1 models on (a) artificial manifold data set and (b) SUSY data set.
Approximation of ELBO and sum of entropies for the fully trained model with different sample sizes, repeated 10 times. Mean of the estimates with 10k samples is visualized as a dotted horizontal line. 
Considering the three entropies expression, the variance in the estimates is barely visible in the scatter plot (upper plot). 
Note, that the standard deviation (lower plot) is displayed on logarithmic scale.
See \cref{SecNum} for the accompanying discussion and
\cref{app:ExpSpecifications} for details on the experimental setup.}
\label{fig:ELBOEstimation_App}
\vspace{-4pt}
\end{figure}

That is, our theoretical results allow for using \textit{analytical} expressions (\cref{theo:VAE-1}) instead of numerical approximations of the analytical integrals of the original objective (\cref{EqnELBO}). Please also see the discussion in \cref{SecNum} in the main paper.
In our experiments, we  observed convergence of the variance parameters after several epochs (depending on architecture, data set and learning rate), %
and thus to the point
at which the conditions for obtaining our closed-form expression are (approximately)
fulfilled.
To demonstrate the advantage of the sum of entropies, we compared it to the na\"ive approximation of the ELBO 
for different number of samples (\cref{fig:ELBOEstimation,fig:ELBOEstimation_App}) using Monte-Carlo sampling.
The three entropies expression can provide very precise estimates of the training-ELBO with very few samples already.

We remark that this effect is slightly stronger for high-dimensional data.
Also note that the decoder entropy of a trained VAE--1 model is a property of the model and not data-dependent. It follows that the decoder entropy clearly should not be confused with a reconstruction measure on unseen data.

\subsection{Entropy-based Posterior Collapse Detection.}
\label{app:PosteriorCollapse_Exp}
\begin{figure}[ht!]
\begin{center}
    \includegraphics[width=0.245\linewidth]{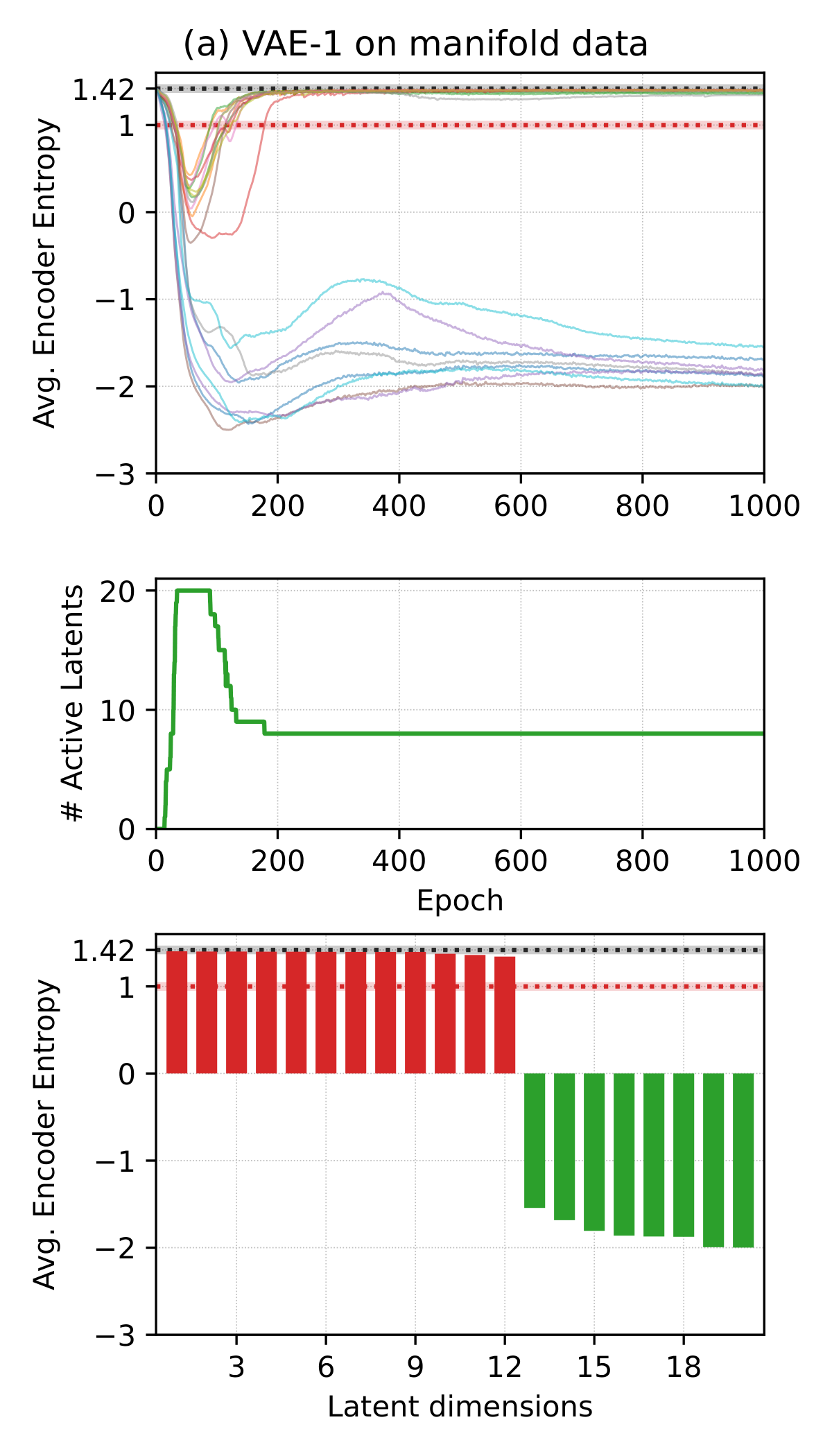}
    \includegraphics[width=0.245\linewidth]{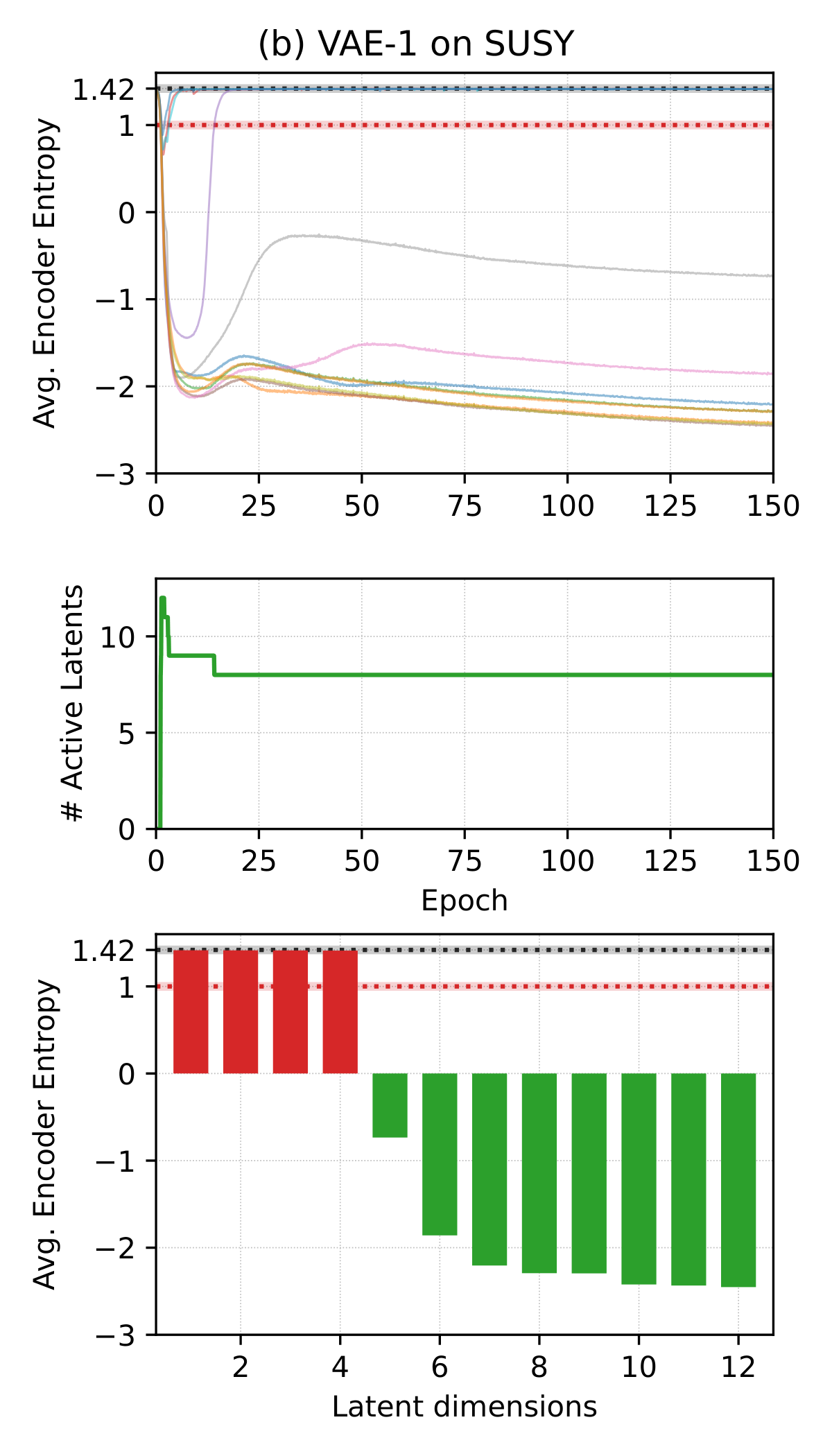}
    \includegraphics[width=0.245\linewidth]{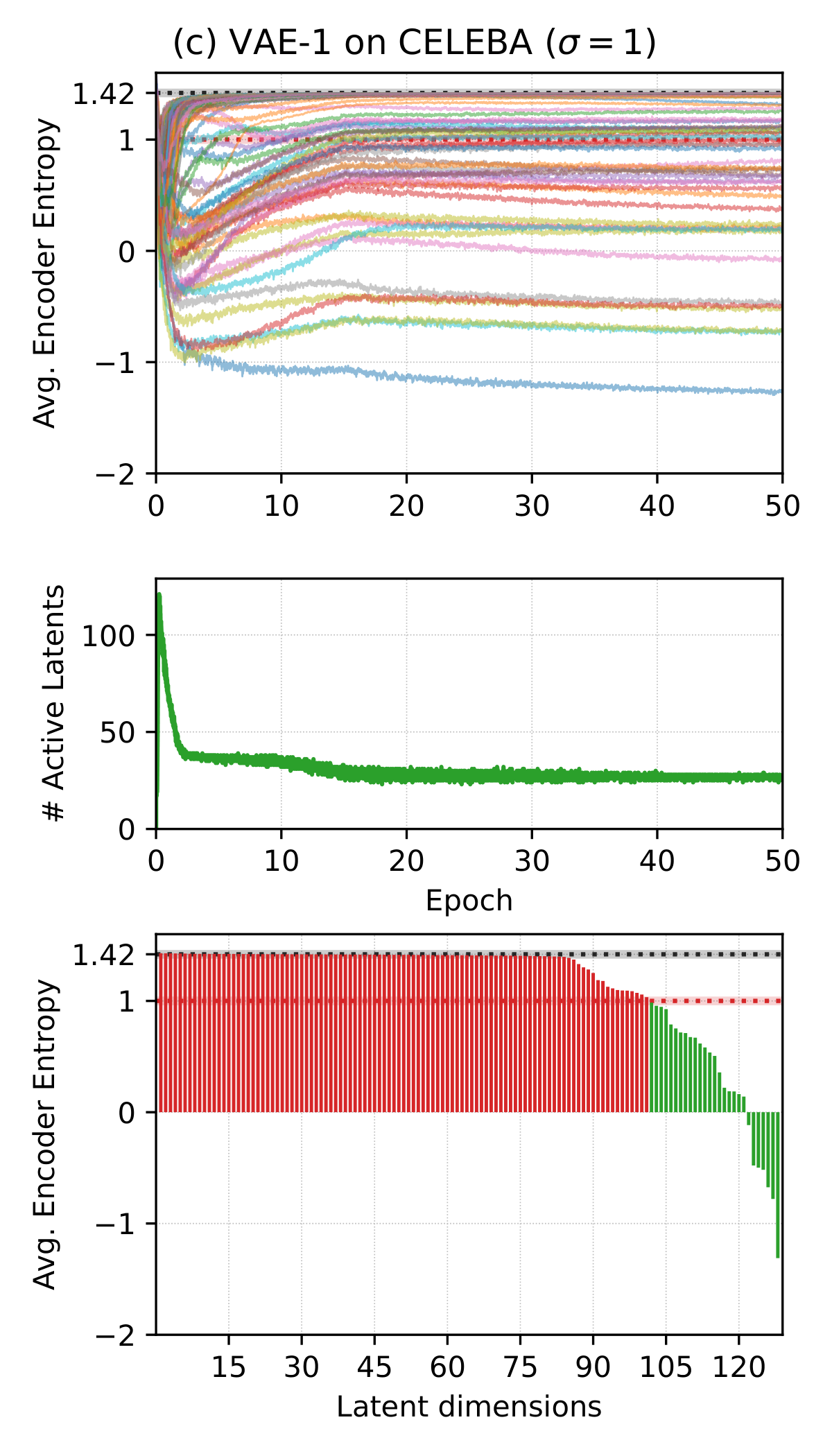}
    \includegraphics[width=0.245\linewidth]{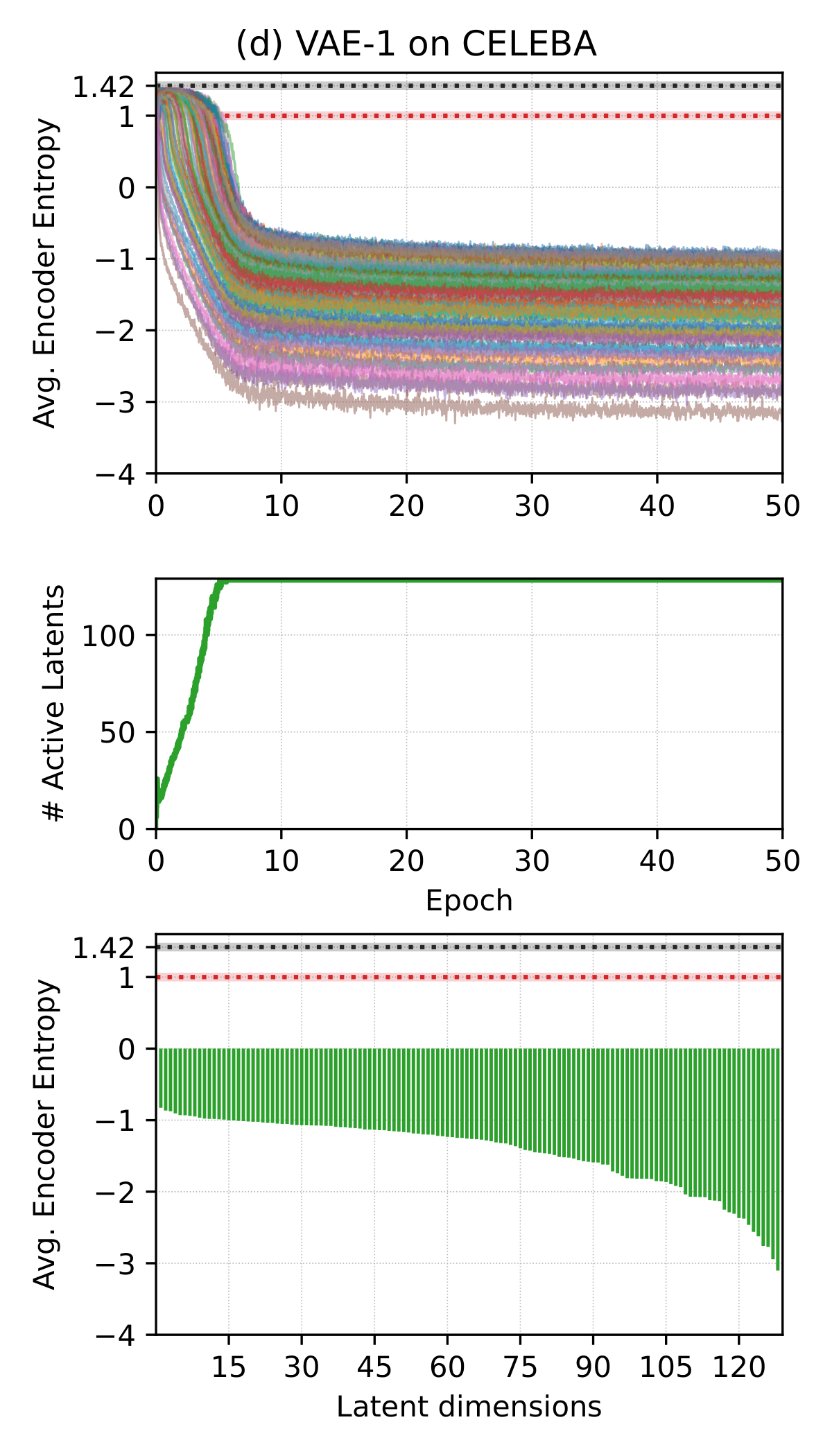}
\end{center}
\vspace{-12pt}
\caption{\textbf{Posterior collapse} analysis for VAE--1 models on different data sets based on the entropy criterion (\cref{EqnPostCollCriterion}). For the threshold we used $\frac{1}{N}\sum_n \HH[\qPhi(z_h \given \xVecN)] > 1$.
\textit{Top plots:} Average encoder entropies $\frac{1}{N}\sum_n \HH[\qPhi(z_h \given \xVecN)]$ over the course of training with upper bound $\HH[p(z_h)] \approx 1.42$ and threshold of one.
\textit{Middle plots:} Number of non-collapsed latent dimensions over the course of training based on the entropy criterion.
\textit{Bottom plots:} Average encoder entropies of the fully trained model per latent in descending order. Collapsed latent dimensions are colored in red.
Note that in (c) we consider a VAE--1 with \textit{fixed} decoder variance which in turn leads to posterior collapse. Encoder entropies are calculated on ten batches for the trained models.
See  \cref{SecNum,app:PosteriorCollapse} for details on posterior collapse and further discussion and \cref{app:Experiments} for the experimental set-up.}
\label{fig:PosteriorCollapse_App}
\vspace{-4pt}
\end{figure}

As outlined in the main part of this paper, the entropy perspective paves the way to elegantly define and analyze posterior collapse based on entropies.
The criterion, given in \cref{EqnPostCollCriterion}, states that a latent variable $z_h$ is $\delta$-collapsed, if the encoder entropy matches the corresponding prior entropy too closely, i.e.,
$
\frac{1}{N}\sum_{n=1}^N \HH[\qPhi(z_h\given\xVecN)] >  \HH[p(z_h)] - \delta $. 
For the derivation and additional discussion see \cref{app:OptimizationLandscape}.
The results are presented in \cref{fig:PosteriorCollapse_App} to which we here add the experiment-specific details.
\textbf{Artificial Data.} We used $H=20$ latents while the true dimensionality of the artificial manifold data set is $9$. Thus, $8$ non-collapsed latent dimensions are reasonable. 
\textbf{SUSY.} Again, we have knowledge of the manifold dimensionality on which the data is concentrated. In this case it is $8$ as all remaining features are functions of the first $8$ features. 
In the considered experiment the intrinsic dimensionality of the data can be successfully recovered. 
\textbf{CelebA.} The manifold hypothesis seems to be valid for this data set as well, however we cannot determine the dimensionality of the data manifold. 
Notably, we do not observe posterior collapse for the full VAE--1 which we relate to the dimension-depending scaling of encoder/prior entropy (or equivalently the KL divergence, which scale with latent dimension $H$) and decoder entropy (scaling with data dimension $D$). Clearly, this is an instance of under-regularization.
In contrast, when restricting $\sigma^2 = 1$ we observe severe posterior collapse (\cref{fig:PosteriorCollapse_App}(c)). 
This relates to the observation that fixing $\sigma^2$ to some sub-optimal value is related to the re-balancing issue of KL divergence and reconstruction term (which is also addressed by \cite{HigginsEtAl2016}). Here, enforcing $\sigma^2 = 1$ increases the weight of the KL-divergence $S_\mathrm{reg}$ (relative to reconstruction score $S_\mathrm{rec}$) such that many latent variables are pruned out. As noted in the main text, posterior collapse is not an issue \emph{per se} but needs to be investigated carefully with respect to the (assumed) intrinsic dimensionality of the data in order not to over- or under-regularize. For further discussion see \cref{SecNum} and for details of the experimental set-up the \cref{app:ExpSpecifications}.

An interesting finding from these experiments is %
that the relative dimensionality ($D$ vs. $H$) contributes in preventing posterior collapse in VAE optimization (provided that, in particular, the decoder variance is learnable). With $D \gg H$ as on CelebA we did not find a single collapsed latent variable ($D = 64^2 \times 3 = 12,288$ vs. $H = 128$). We plan to investigate these findings in future.

\subsection{Experimental Set-up and Implementation Details}
\label{app:ExpSpecifications}

\begin{wrapfigure}{r}{0.44\textwidth}
\vspace{-5ex}
  \begin{center}
    \includegraphics[width=0.44\textwidth]{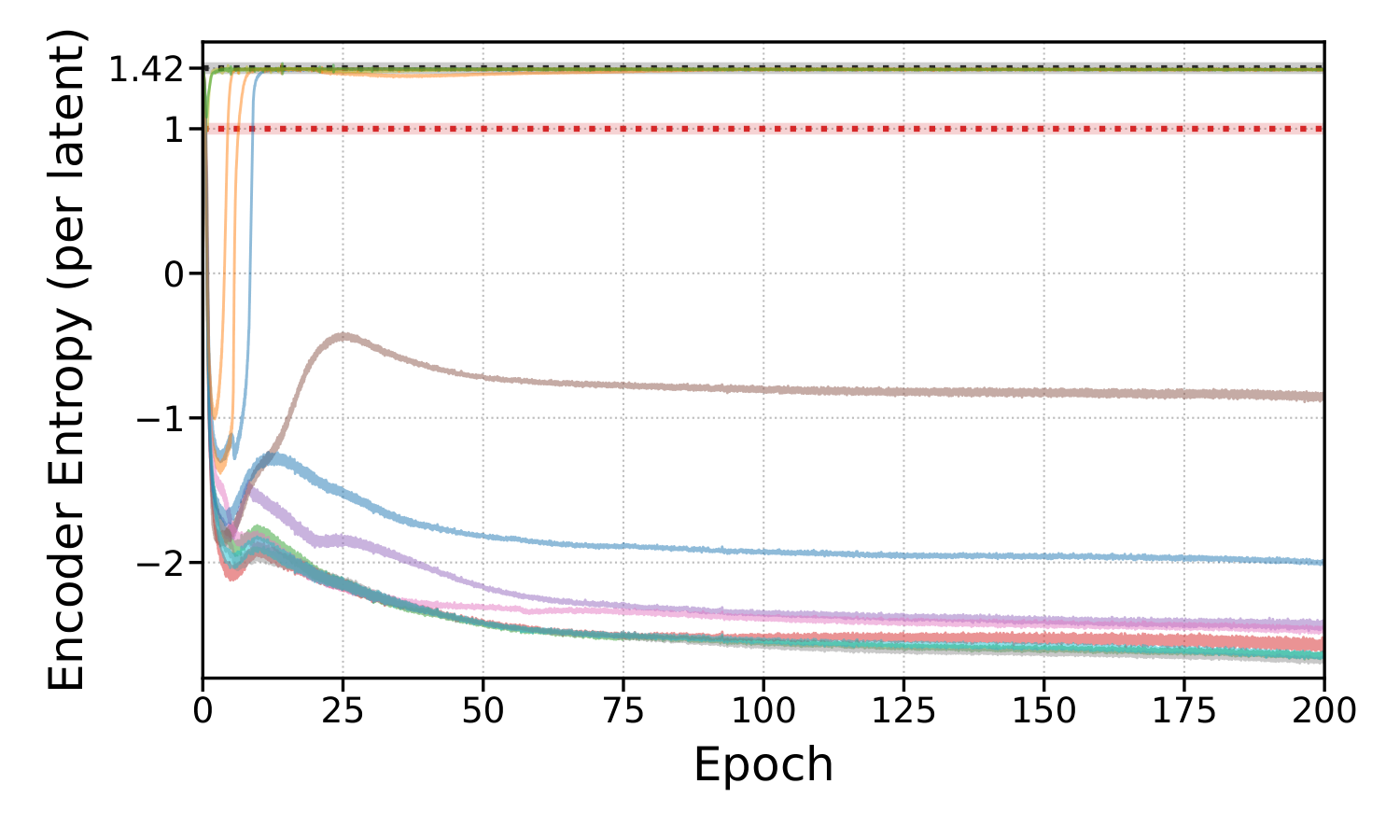}
  \end{center}
  \caption{\textbf{Additional Posterior Collapse} experiment with VAE--1 on the SUSY data set. 
  We altered the latent dimension, here to $H=13$, and observed reliable recovery of the intrinsic data dimensionality across different random seeds.}
\vspace{-5mm}
\end{wrapfigure}
We here provide the specifications of the considered experiments on SUSY, CelebA and artificial manifold data.

\paragraph{Experiments on CelebA (VAE--1)}

In order to demonstrate the theoretical result on a more complex architecture we used the publicly available VAE--1 implementation of \cite{Subramanian2020} that involves convolutional layers and batch normalization. As this implementation (like many other) use a fixed decoder variance $\sigma^2$ (and then rely on rescaling the KL-Term in the ELBO), the only architectural change we made is to turn $\sigma^2$ into a learnable parameter which we then train end-to-end with the rest of the model.
For the experiments on posterior collapse (\cref{fig:PosteriorCollapse_App}) we trained another model with fixed decoder variance $\sigma^2 = 1$ and simple linear KL-annealing schedule from $\beta=0.1$ to $\beta=1.0$ over the first 15 epochs.
We resized the data set such that the data dimension amounts to $D = 64 \times 64 \times 3 = 12,288$ and set the latent dimension to $H=128$.
We used a batch size of $256$, a train-test-split of $60\%/40\%$, and optimized for $50$ epochs with ADAM (using default learning rate of $10^{-3}$). %
We employed data augmentation using random horizontal flips.

\paragraph{Experiments on SUSY (VAE--1)}

SUSY is a machine learning data set from the field of high-energy physics and consists of 5 million measurements of processes that potentially produce super-symmetric particles \citep{baldi2014searching}. We decided to include experiments on this real-world dataset to illustrate the phenomenon of posterior collapse: Despite having data dimensionality $D=18$, only the first $8$ features are measurements of kinetic properties, while the remaining $10$ are functions of the first $8$ features.\footnote{For further description see the corresponding website \url{https://archive.ics.uci.edu/ml/datasets/SUSY} and \cite{baldi2014searching}.} Thus, we have knowledge about the dimensionality of the manifold on which the data is located and can therefore assess posterior collapse on a solid basis.\footnote{On usual high-dimensional data sets (thinking of image data sets like CIFAR or CelebA) we cannot discern the dimensionality of the manifold on which the data is concentrated.}

We used a VAE of type VAE--1 implemented as a simple two-layer perceptron with $128$ and $24$ hidden neurons and Leaky ReLU as activation function for the encoder and in reversed order for the decoder.
We use $H=12$ latent dimensions and optimized with ADAM with the standard initial learning rate of $10^{-3}$ for (at least) $150$ epochs with a batch size of $1024$. We reduced the data set size to 2.5 million samples, used a train-test-split of $50\%/50\%$ and scaled the features to range $[0,1]$ before training. %

In order to learn a proper embedding in the latent space first (and avoid over-regularization in early training stages) we initialized $\ln(\sigma^2_\text{init}) = -9$. This small value puts high weight on reducing the reconstruction term first. During training $\sigma^2$ becomes optimal which restores the balance between regularization and reconstruction. We found this simple trick helpful to avoid widespread posterior collapse in the first few epochs for this data set.

\paragraph{Experiments on artificial manifold data set (VAE--1)}
Besides the real world data sets described above we want to test our ideas regarding posterior collapse on a data set where we have explicit control over the intrinsic dimensionality of the data.
For this reason we created a $D=100$ dimensional data set for which we can control the dimensionality of the lower-dimensional manifold on which the data is concentrated (just as in the case of SUSY). In the experiments presented here we use $k = 9$ dimensions which we initialized with Gaussian noise. The remaining dimensions are then created using non-linear functions applied to (a subset) of the first $k$ dimensions. For the experiments presented here, we simply applied the cosine function to a random linear combination of a pair of the first $k$ dimensions (squared). E.g., dimension $p \in [10,100]$ is simply (element-wise) $\xVec
_p = \cos(\xVec^2_r + \xVec_s^2)$ for $r,s$ randomly drawn among the first $k=9$ dimensions (with replacement).
We added small Gaussian noise to this manifold and resized all features back to range $[0,1]$.

The experiments are conducted on $50.000$ samples from this data set with a split of $50\%$ training set and $50\%$ test set.
We set the latent dimension to $H=20$. Again we used simple two-layer perceptrons with $128$ and $40$ hidden neurons with Leaky ReLU for the encoder, and in reversed order for the decoder. For the same reason as in the SUSY experiments, we initialized $\ln(\sigma^2_\text{init}) = -9$. Again, we used ADAM with the standard initial learning rate of $10^{-3}$ and a batch size of $512$.

\subsection{Generation of Artificial PCA Data Sets and Learning Visualizations}
\label{app:ExpArtificailPCAdata}
We generated the PCA data set according to the following generative model of probabilistic PCA:
\begin{align}
p(\mathbf{z}) &= \NN(\zVec; \mathbf{0}, \One) \\
p(\mathbf{x}\,|\,\mathbf{z}) &= \NN(\mathbf{x}; W_\mathrm{gen.}\mathbf{z} + \mathbf{\mu}_\mathrm{gen.}; \sigma_\mathrm{gen.}^2 \One).
\end{align}
The generative parameters $W_\mathrm{gen.}$ and $\mathbf{\mu}_\mathrm{gen.}$ were drawn randomly from uniform distributions between 0 and 1 in each dimension for each new run.
$\sigma_\mathrm{gen.}$ was always set to $0.1$.
We used $H=2$ as latent dimension and $D=10$ as output dimension and generated 10.000~new training and testing data points for each experiment.

For the PCA-ring data sets, we introduced an additional non-linear transformation $\mathbf{x}' = g(\mathbf{x})$ with
\begin{align}
p(\mathbf{z}) &= \NN(\zVec; \mathbf{0}, \One) \\
p(\mathbf{x}\,|\,\mathbf{z}) &= \NN(\mathbf{x}; W_\mathrm{gen.}\mathbf{z}; \sigma_\mathrm{gen.}^2 \One) \\
\mathbf{x}' &= \mathbf{\mu}_\mathrm{gen.} + \frac{\mathbf{x}}{10} + \frac{\mathbf{x}}{|\mathbf{x}|}
\end{align}
projecting the data onto a ring-like structure in $D$-dimensional space \citep[see, e.g.,][]{Doersch2016}.

\cref{fig:visualization}(c), on the last page, shows a PCA projection of the training data set, visualizations of the $\zVec-$ and $\xVec$-space during training of VAE--3 as well as plots of the lower bound and the three entropies.
Figs.\,\ref{fig:visualization}(a) and (b) show the same plots for the linear VAE and VAE--1 on the PCA and MNIST data set, respectively.
As expected, we see a slight overfitting of the linear VAE to the PCA training data, with the training log-likelihood converging to a value slightly above of the ground-truth.
However, even with early stopping at around 700~iterations (see third segment of \cref{fig:visualization}(a), we see that the three entropies already compute the lower bound very well.

\subsection{Noise in Three Entropies and ELBO}
\label{app:ExpNoise}
\begin{figure*}[hb!]
\includegraphics[width=0.33\linewidth]{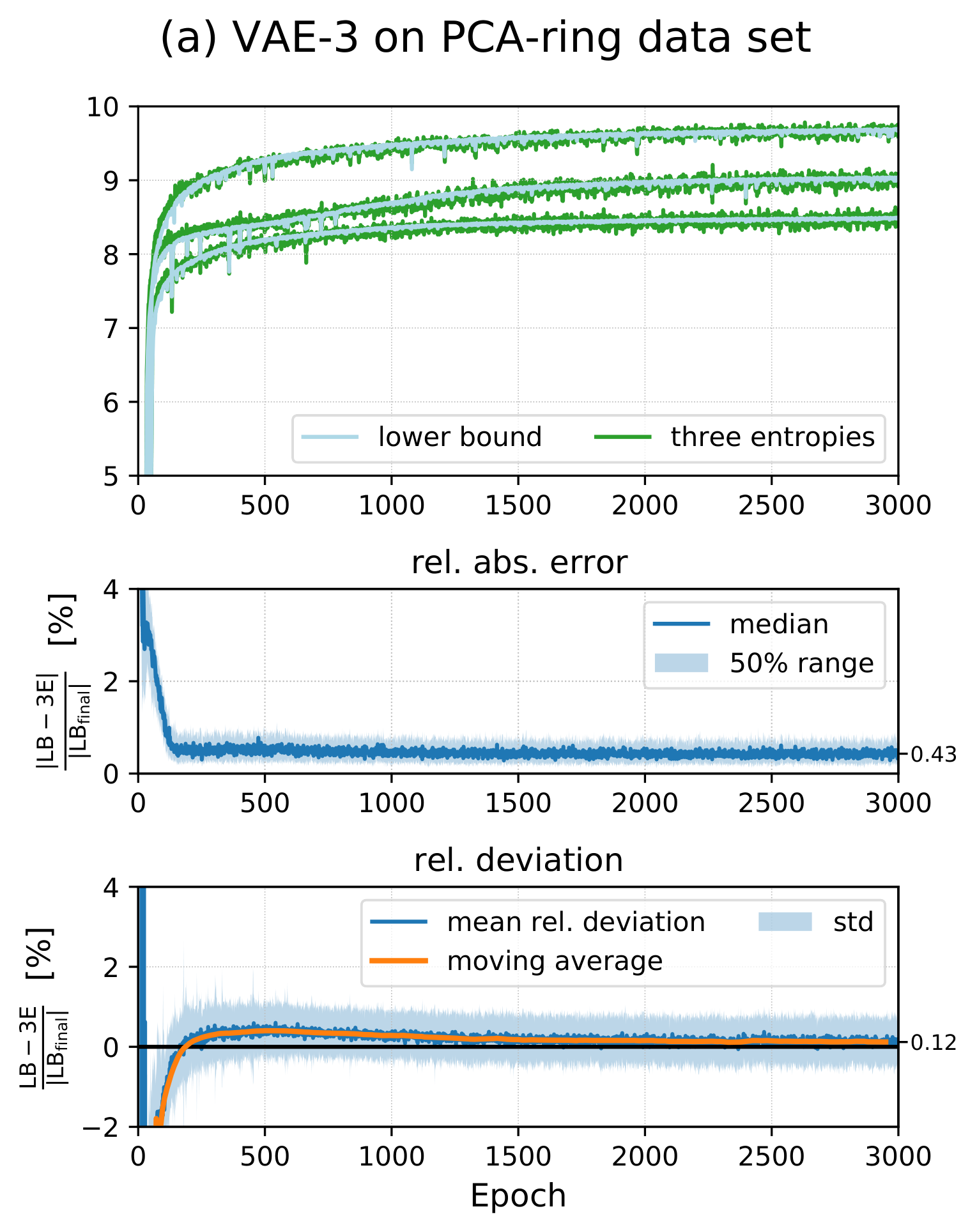}  
\includegraphics[width=0.33\linewidth]{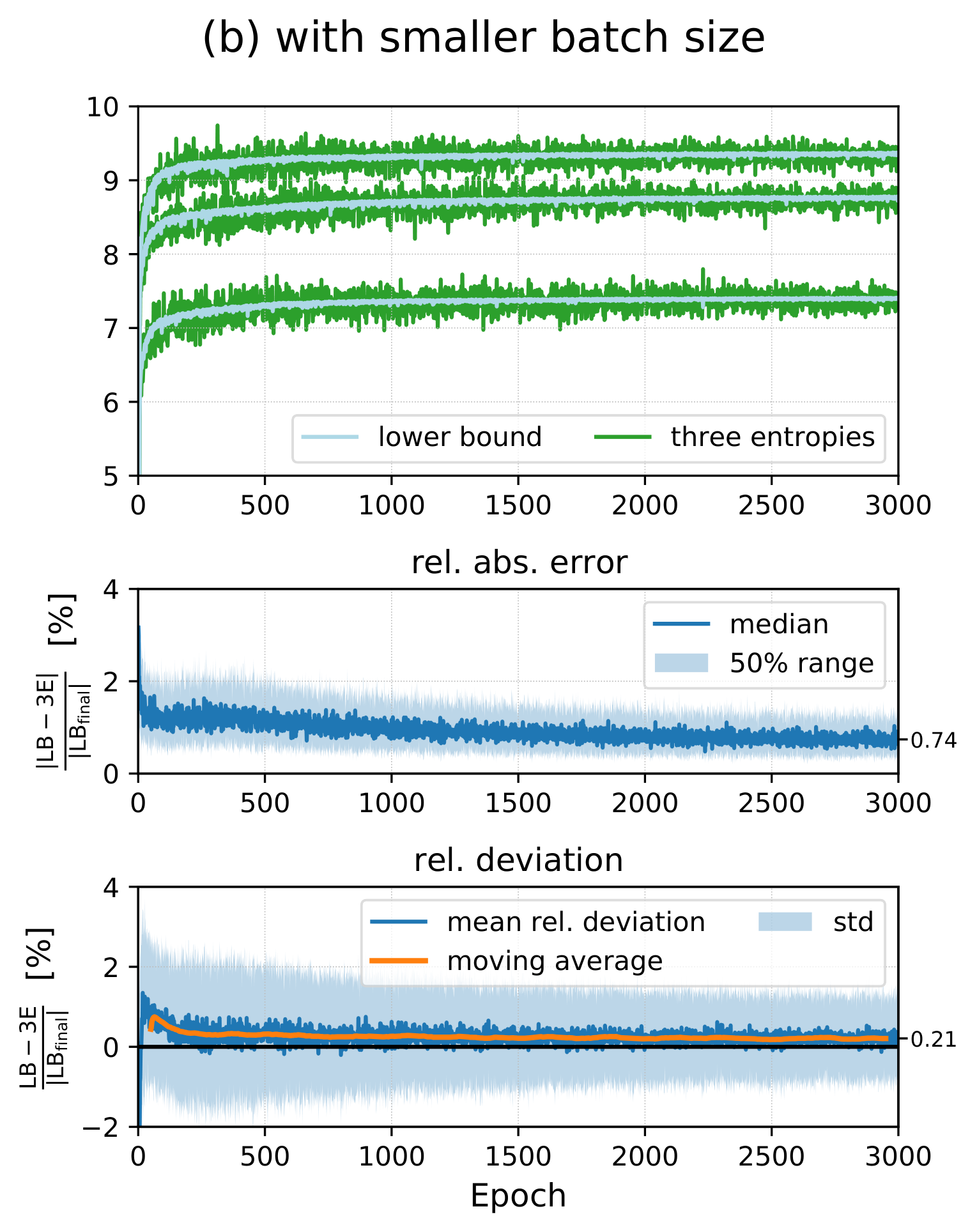}
\includegraphics[width=0.33\linewidth]{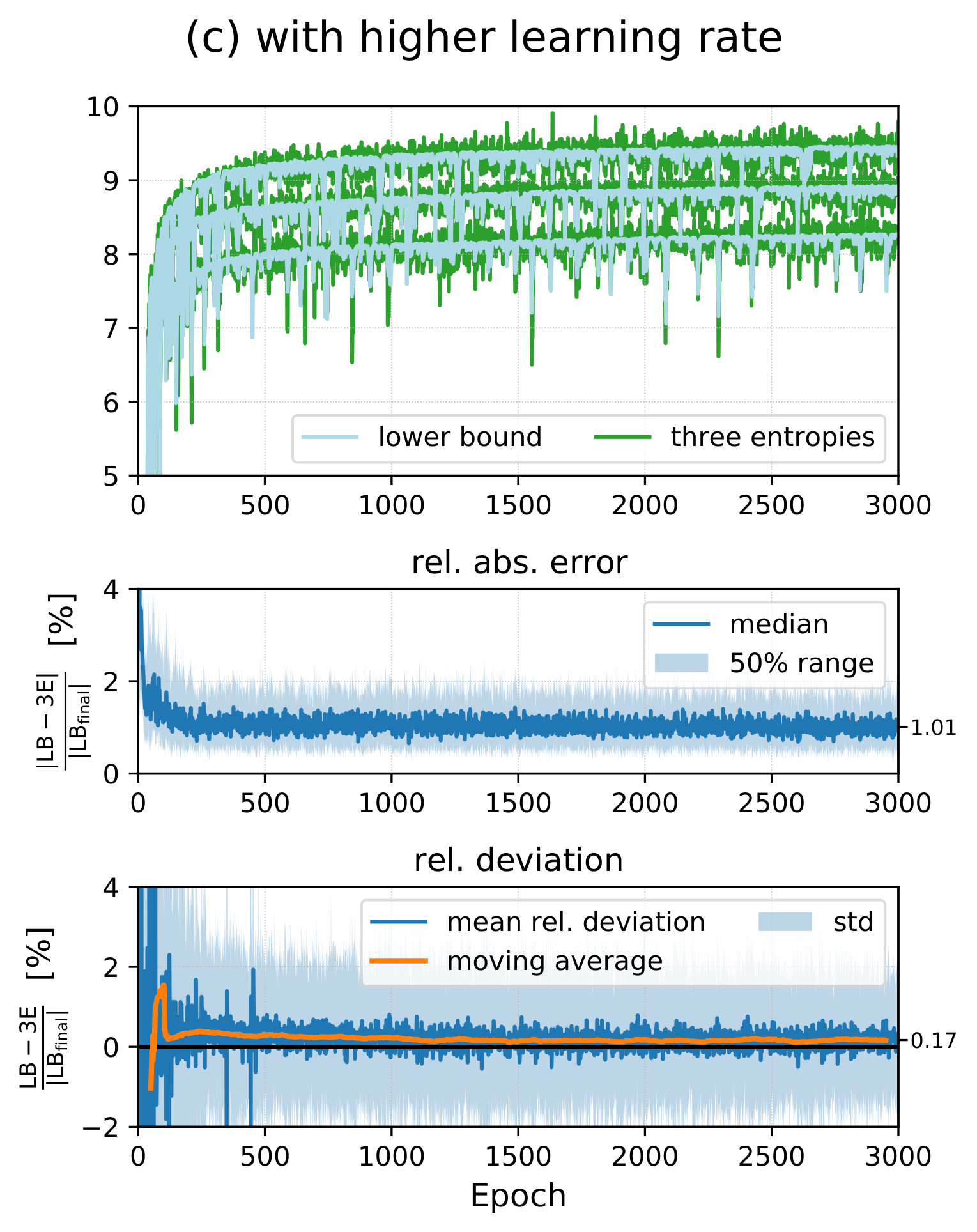}
\vspace{-6pt}
\caption{VAE--3 on PCA-ring data sets. The top two plots show experiments in the same way as \cref{fig:Verification}:
The top plot shows three independent runs on new randomly generated PCA-ring data each, while the middle plot shows the median of the relative absolute error between lower bound and three entropies over 100~such independent runs.
Additionally we show the mean relative deviation (i.e., without taking absolute values) between lower bound an three entropies of these runs as bottom plots, together with the moving averages over 100~epochs.
(a)~shows experiment in the same setting as \cref{fig:Verification}(c) over longer training time. (b) shows the same experiments with a batch size of 100 (compared to a batch size of 2000 in the other plots). (c) shows the same experiments with a learning rate of 0.005 (compared to a learning rate of 0.001 in the other plots).}
\label{fig:noise}
\vspace{-6pt}
\end{figure*}

The optimization of VAEs is stochastic due to finite learning rates, finite batch sizes and approximations of integrals over $\mathbf{z}$ using sampling. A stationary point is consequently never fully converged to. Instead the parameters will finally stochastically fluctuate around a stationary point. In \cref{fig:Verification}(c) we have (for VAE--3) numerically quantified the relative difference between the variational lower bound and Eqn.\,(\ref{PropThree}) of \cref{theo:VAE-3}. As can be observed, and as stated in the main text, the values for the standard variational bound (\cref{EqnELBO,PropThree}) match very well in the region close to the stationary point (i.e., the region to which stochastic learning finally converges to).
By evaluating over 100~runs, we observed an average {\em absolute} error of below 0.5\% (i.e., $0.005$) between the variational bound and Eqn.\,(\ref{PropThree}). 

We can also further study the effect of stochasticity by using smaller batch sizes (\cref{fig:noise}(b) or higher learning rates (\cref{fig:noise}c).
In both cases, we obtain a higher stochasticity of the final fluctuations around the stationary point. As a consequence, the average absolute error becomes larger (\cref{fig:noise}, middle row) but is still smaller-equal than $1\%$ (i.e., $0.01$). The change of the error with changing stochasticity is better observed for the absolute relative error than for the relative deviation
(i.e., if we define the relative deviation as the relative absolute error but without taking magnitudes in the numerator, see \cref{fig:noise}, bottom row).
However, for completeness, we also provide the relative deviation which is (as expected) smaller than the error (\cref{fig:noise}, bottom row). 
This means that the difference between variational lower bound and Eqn.\,(\ref{PropThree}) can be positive as well as negative close to the stationary point. Averaging cancels out the differences in large parts, which is the reason for the relative deviation being significantly smaller than the relative absolute error.

In summary, the numerical experiments provide (A)~consistency with the theoretical result of \cref{theo:VAE-3} (and the other Theorems), and (B)~they show that the three entropies results of %
\cref{theo:VAE-1,theo:VAE-3}
can provide very accurate estimations of the variational bound in practice.
As demonstrated in \cref{fig:ELBOEstimation,fig:ELBOEstimation_App} the three entropy expression is even able to significantly reduce the variance in ELBO estimation for trained VAE--1 models.
Note in this respect that the computation of final values of lower bounds always provides valuable information. Final lower bounds can be used as approximations to the true log-likelihood in many settings, and could be used for comparisons between different runs and/or different VAEs. Typical such comparisons can be applied for model selection, for instance. But also when the lower bounds are not good approximations of the log-likelihood, knowing their values at convergence is very useful to analyze learning: runs with high final values for variational lower bounds but low values for (held-out) log-likelihood indicate overfitting, for instance. %
The experiments of \cref{fig:Verification,fig:noise} show that the values of variational lower bounds can in practice easily be estimated with high accuracy. In the case of VAEs in the form of VAE--1, estimation with high accuracy is even possible using a closed-form expressions (see \cref{theo:VAE-1}) which in turn enable us to reduce the variance in ELBO estimation by at least a factor of 10 (\cref{fig:ELBOEstimation,fig:ELBOEstimation_App} and the accompanying discussion). 

\subsection{Example Implementation}
\label{app:Code}
Listing~\ref{lst:Code} shows an example \texttt{PyTorch} implementation of the linear VAE as well as functions to compute the three entropies for the linear VAE, as well as VAE--1 and VAE--3, either directly or based on \texttt{torch.distributions}. For the full code see the link provided in the beginning of \cref{app:Experiments}.

\definecolor{codegreen}{rgb}{0,0.6,0}
\definecolor{codegray}{rgb}{0.5,0.5,0.5}
\definecolor{codepurple}{rgb}{0.58,0,0.82}
\definecolor{backcolour}{rgb}{0.95,0.95,0.93}

\lstdefinestyle{mystyle}{
    backgroundcolor=\color{backcolour},   
    commentstyle=\color{codegreen},
    keywordstyle=\color{magenta},
    numberstyle=\tiny\color{codegray},
    stringstyle=\color{codepurple},
    basicstyle=\ttfamily\footnotesize,
    breakatwhitespace=false,         
    breaklines=true,                 
    captionpos=t,                    
    keepspaces=true,                 
    numbers=left,
    xleftmargin=2em,
    showspaces=false,                
    showstringspaces=false,
    showtabs=false,                  
    tabsize=2,
    frame=lines,
    language=Python,
}

\lstset{style=mystyle}
\lstset{caption={Example Implementations}}
\lstset{label={lst:Code}}
\begin{lstlisting}
import torch
from torch.nn as import Linear, Module, Parameter
from torch.distributions import Normal, kl_divergence
import numpy as np


class LinearVAE(Module):
    def __init__(self, H, D, num_samples=100):
        super(LinearVAE, self).__init__()
        self.H = H
        self.D = D
        self.num_samples = num_samples

        self.encoder = Linear(D, H)
        self.decoder = Linear(H, D)
        self.noise_std = Parameter(torch.Tensor([0.]))
        self.q_z_std = Parameter(torch.zeros(H))
        self.p_z = Normal(torch.Tensor([0.]), torch.Tensor([1.]))

    def forward(self, x):
        sigma = softplus(self.noise_std)
        tau = softplus(self.q_z_std)
        N, D = x.shape

        z_params = self.encoder(x)
        q_z = Normal(z_params, tau)

        z_samples = q_z.rsample([self.num_samples])
        p_x_z_mean = self.decoder(z_samples)
        p_x_z = Normal(p_x_z_mean, sigma)

        lower_bound = p_x_z.log_prob(x).mean(0).sum() \
                      - kl_divergence(q_z, self.p_z).sum()
        three_entropies_linear = self.three_entropies_linear(N, D, sigma, tau)
        return lower_bound, three_entropies_linear

    @staticmethod
    def three_entropies_linear(N, D, sigma, tau):
        """Three Entropies (accumulated) for linear VAE, see Corollary 1.
        Args:
            N (int) : batch size
            D (int) : output dimensionality
            sigma (torch.tensor, size=(1,)) : decoder standard deviation
            tau (torch.tensor, size=(H,)    : encoder standard deviation
        """
        return N*(-D/2*(torch.log(2*pi)+1)-D*torch.log(sigma)+torch.log(tau).sum())


class VAE1(Module):
		
		[...]		

    @staticmethod
    def three_entropies_nonlinear(N, D, sigma, tau):
        """Three Entropies (accumulated) for non-linear VAE (VAE-1), see Theorem 2.
        Args:
            N (int) : batch size
            D (int) : output dimensionality
            sigma (torch.tensor, size=(1,)) : decoder standard deviation
            tau (torch.tensor, size=(N, H)) : encoder standard deviation
        """
        return N*D*(-1/2*(torch.log(2*pi)+1)-torch.log(sigma)) + torch.log(tau).sum()

class VAE3(Module):
		
		[...]
		
		@staticmethod
    def three_entropies_sigmaz(N, D, sigma, tau):
        """Three Entropies (accumulated) for sigma(z)-VAE (VAE-3), see Theorem 3.
        Args:
            N (int) : batch size
            D (int) : output dimensionality
            sigma (torch.tensor, size=(num_samples, N, D)) : decoder std
            tau (torch.tensor, size=(N, H)) : encoder standard deviation
        """
        return -N*D/2*(torch.log(2*pi)+1) - torch.log(sigma).mean(0).sum() \
				       + torch.log(tau).sum()
           
        
def ELBO_three_entropies(vae, x, x_recon, enc_mu, enc_logvar, dec_logvar,
                         device):
    """
    Calculate ELBO and the sum of three entropies (H_sum); see Theorem 2 & 3
    :param vae: VAE instance (either VAE-1, i.e., dec_logar = log(sigma^2) 
                                  or VAE-3, i.e., dec_logar = DNN(z))
    :param x: input data (usually a batch)
    :param x_recon: reconstructed data (by VAE)
    :param enc_mu: Encoder means
    :param enc_logvar: Encoder log-variances
    :param dec_logvar: Decoder log-variance(s)
    :return: ELBO, H_sum
    """
    # Encoder Distribution q(z|x)
    q_z_x = Normal(enc_mu, (0.5 * enc_logvar).exp())	
    
    # Decoder Distribution p(x|z) with learned variance
    p_x_z = Normal(x, (0.5 * dec_logvar).exp())
        
    # Prior Distribution p(z) (standard normal prior)
    p_z = Normal(torch.zeros(vae.H).to(device), torch.ones(vae.H).to(device))
    
    # Calculate the ELBO (mean of batch, sum over data dims d/latent dims h)
    log_prob = p_x_z.log_prob(x_recon).mean(0).sum() 
    kl_div = kl_divergence(q_z_x, p_z).mean(0).sum()
    
    ELBO = log_prob - kl_div 
    
    # Calculate the Three Entropies
    H_prior = p_z.entropy().sum() # sum over latents
    H_enc = q_z_x.entropy().mean(0).sum() # mean of batch, sum over latents
    H_dec = p_x_z.entropy().mean(0).sum() # mean of batch, sum over data dims
    
    H_sum = - H_dec - H_prior + H_enc
    
    return ELBO, H_sum
\end{lstlisting}

\begin{figure*}[tb]
\begin{adjustbox}{varwidth=\textwidth, fbox, center}
	\hfill
	\includegraphics[width=.48\linewidth]{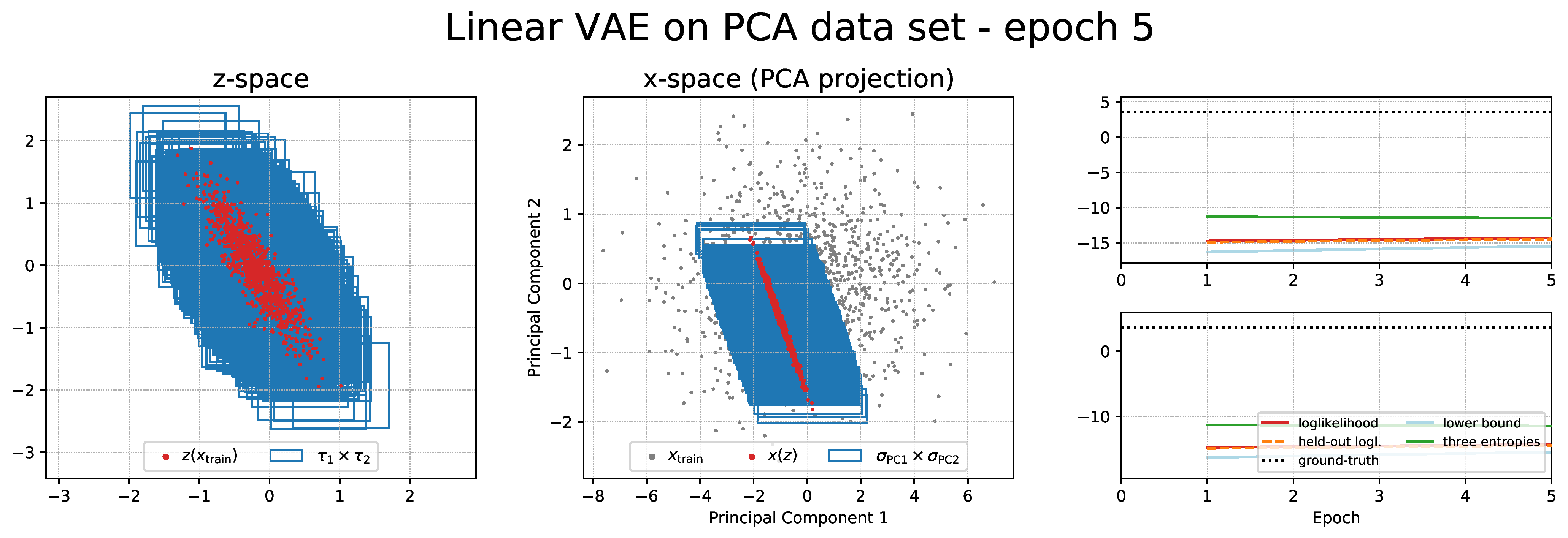}
	\hfill
	\includegraphics[width=.48\linewidth]{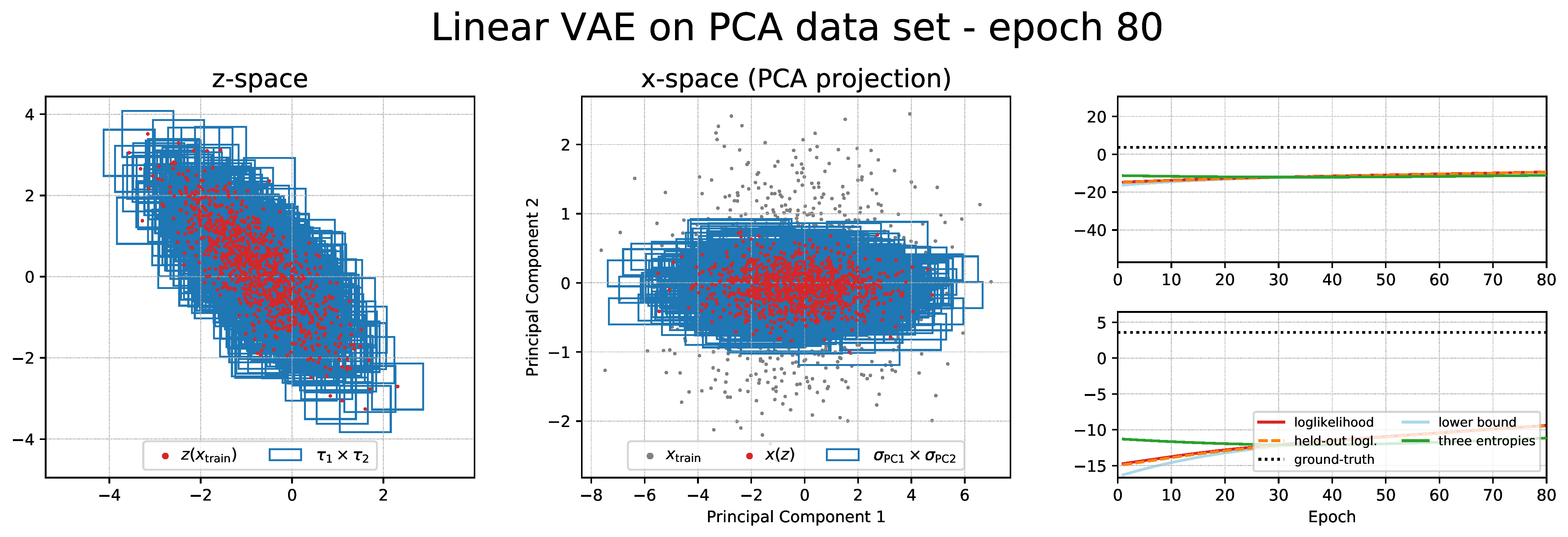}
	\hfill

	\vspace{10pt}

	\hfill
	\includegraphics[width=.48\linewidth]{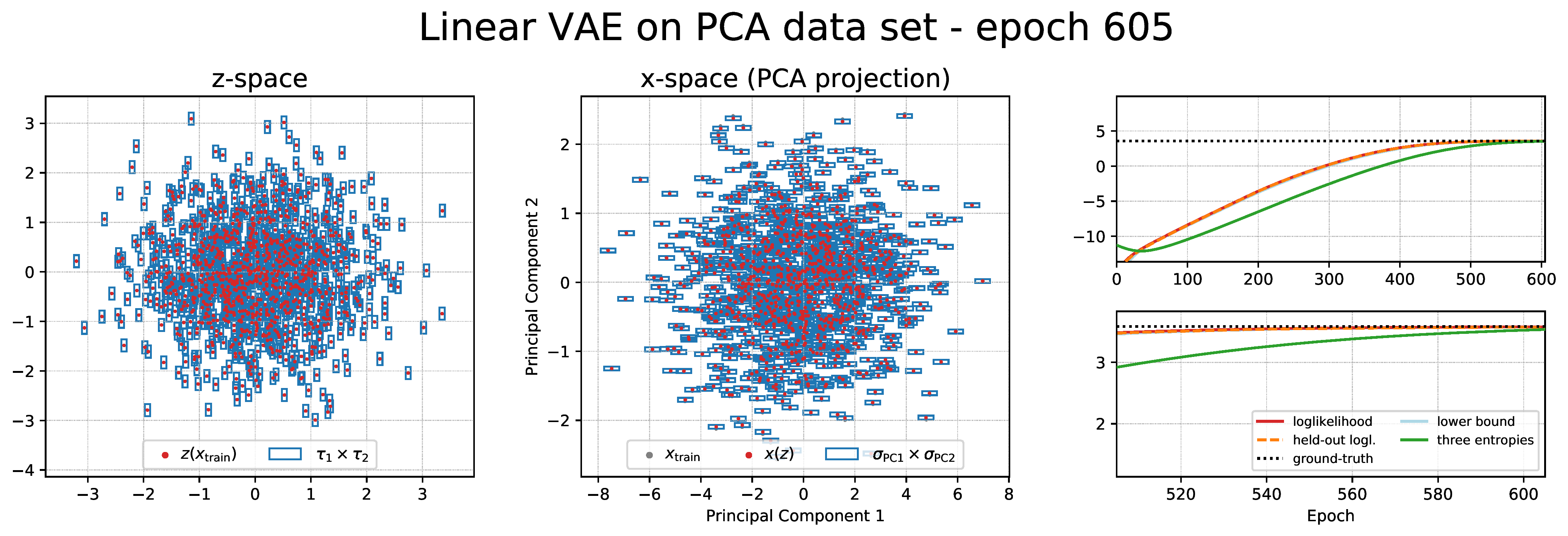}
	\hfill
	\includegraphics[width=.48\linewidth]{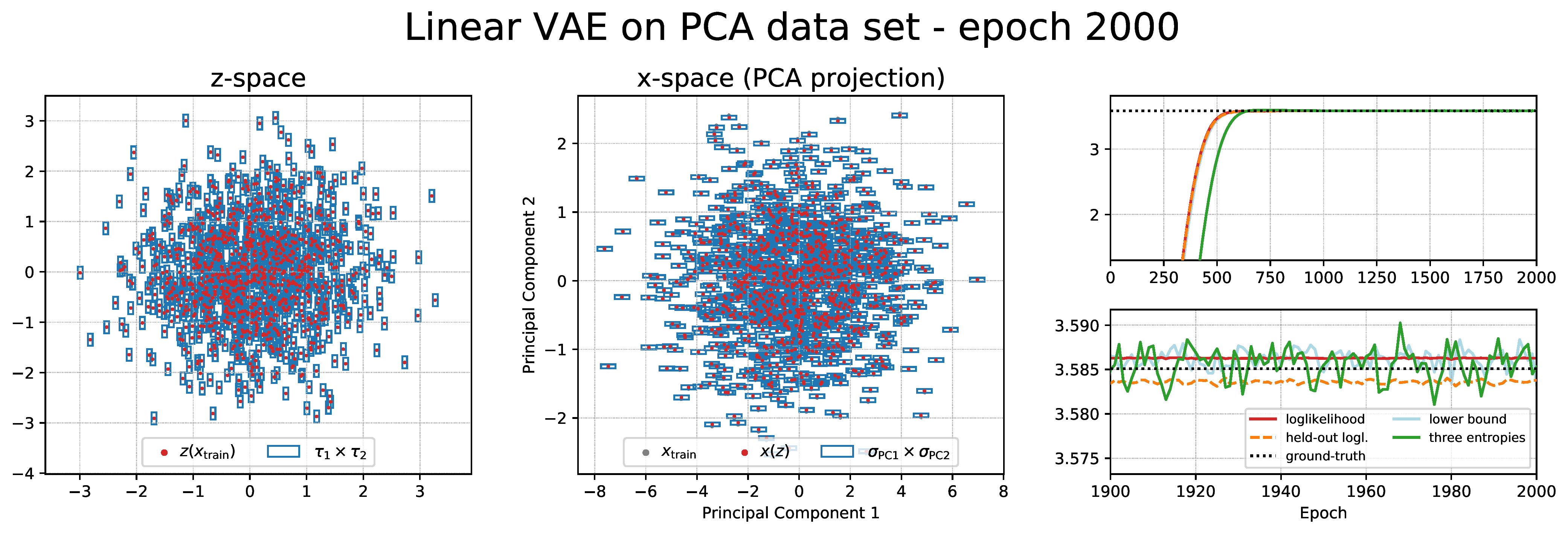}
	\hfill
	
	(a) Linear VAE on PCA data set
\end{adjustbox}
\newline
\vspace{5pt}
\newline	
\begin{adjustbox}{varwidth=\textwidth, fbox, center}
	\hfill
	\includegraphics[width=.48\linewidth]{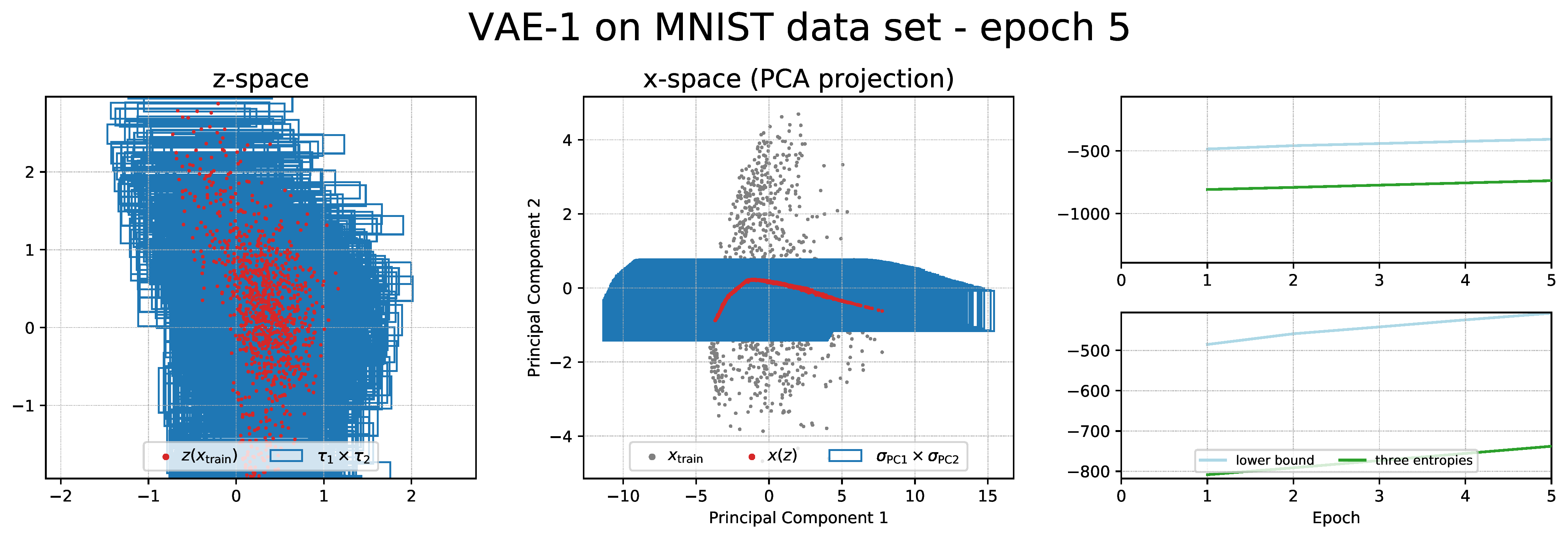}  
	\hfill
	\includegraphics[width=.48\linewidth]{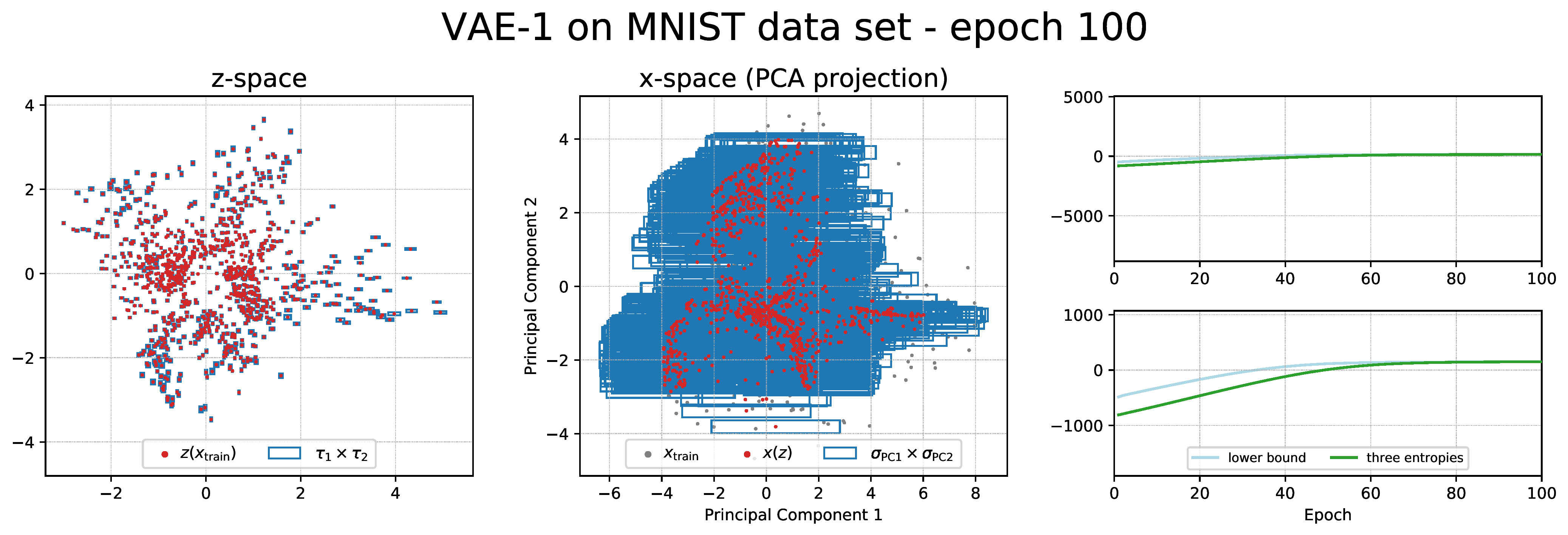}
	\hfill

	\vspace{10pt}
	
	\hfill
	\includegraphics[width=.48\linewidth]{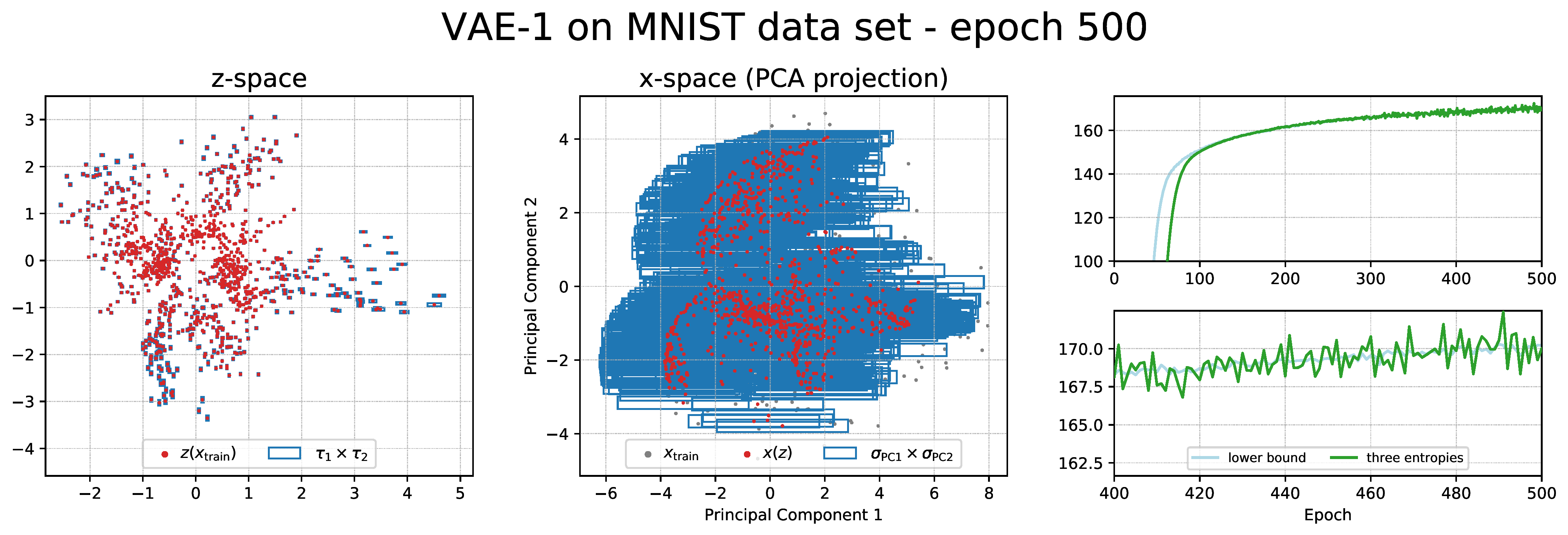}
	\hfill
	\includegraphics[width=.48\linewidth]{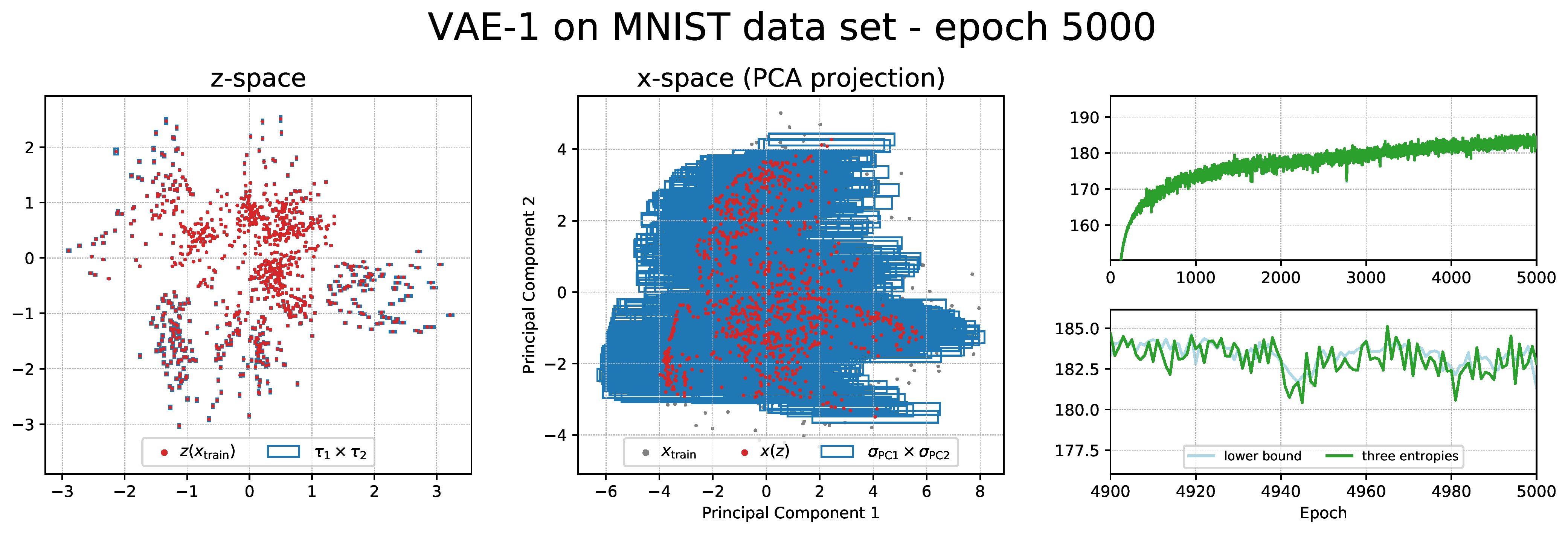}
	\hfill
	
	(b) VAE--1 on MNIST data set
\end{adjustbox}
\newline
\vspace{5pt}
\newline	
\begin{adjustbox}{varwidth=\textwidth, fbox, center}
	\hfill
	\includegraphics[width=.48\linewidth]{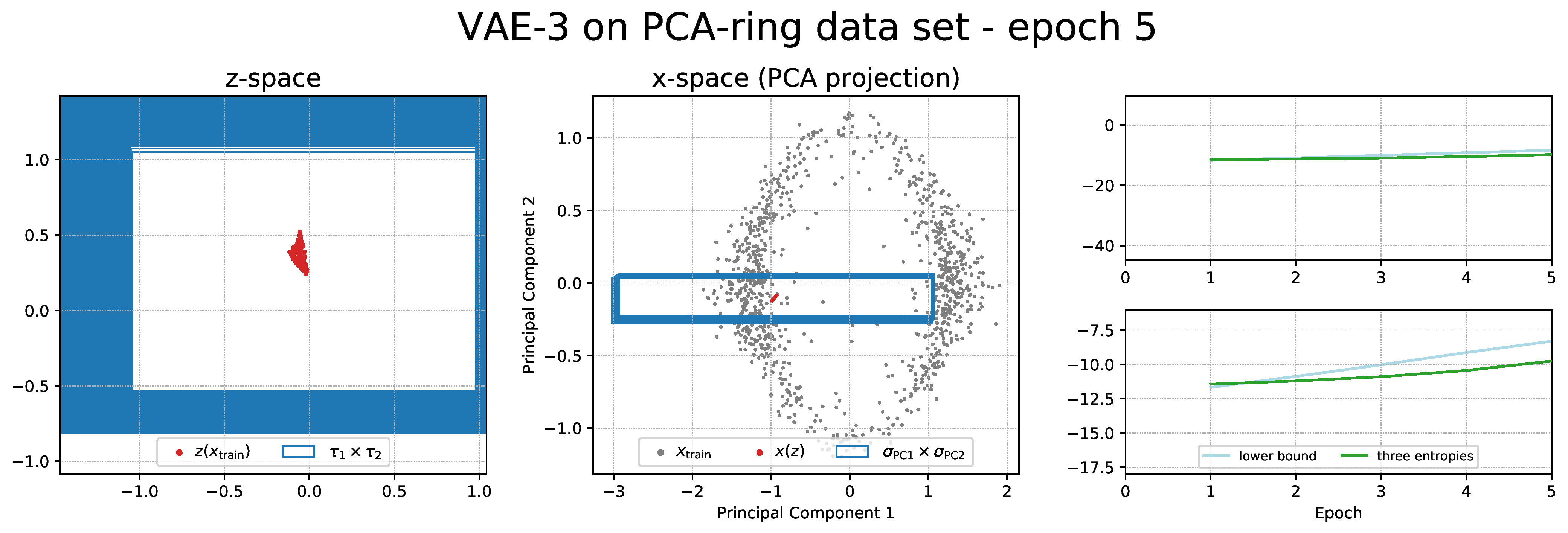}  
	\hfill
	\includegraphics[width=.48\linewidth]{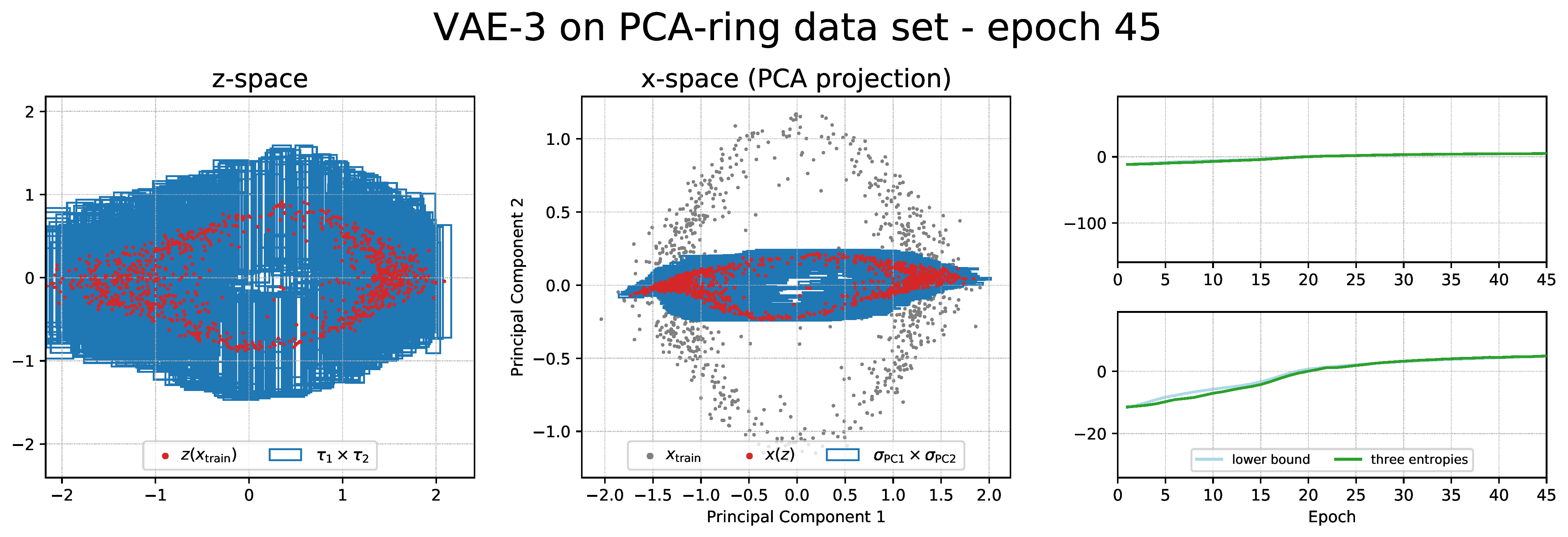}
	\hfill

	\vspace{10pt}

	\hfill
	\includegraphics[width=.48\linewidth]{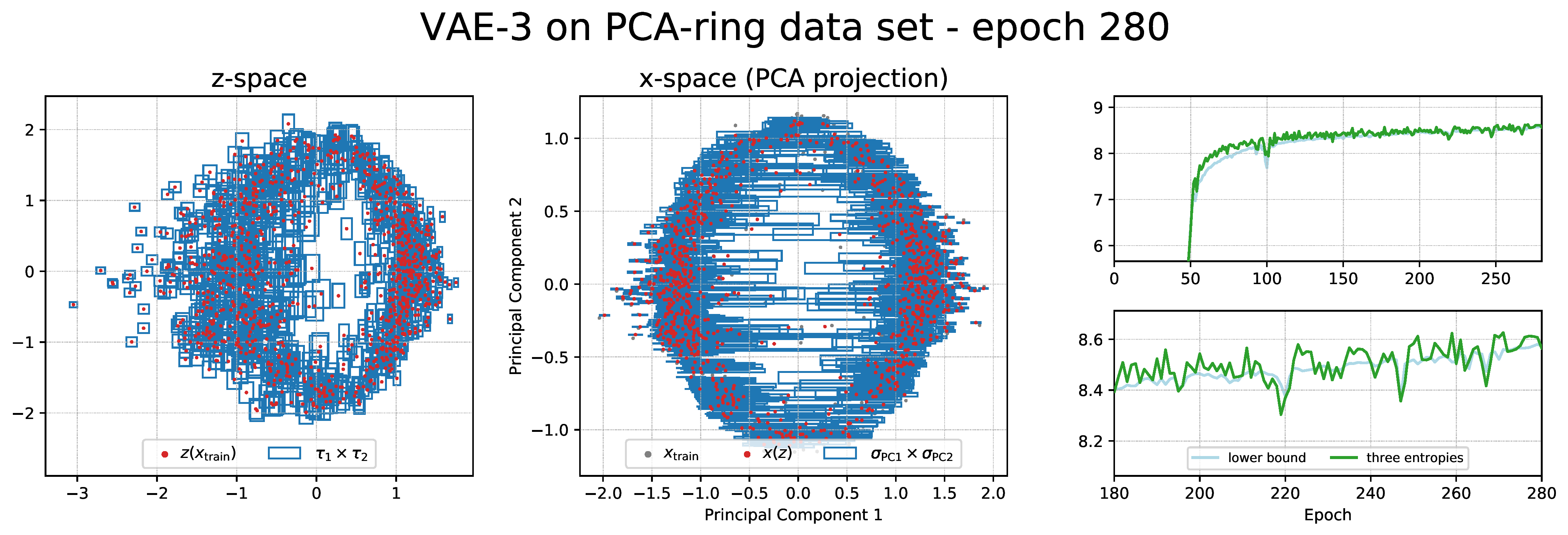}
	\hfill
	\includegraphics[width=.48\linewidth]{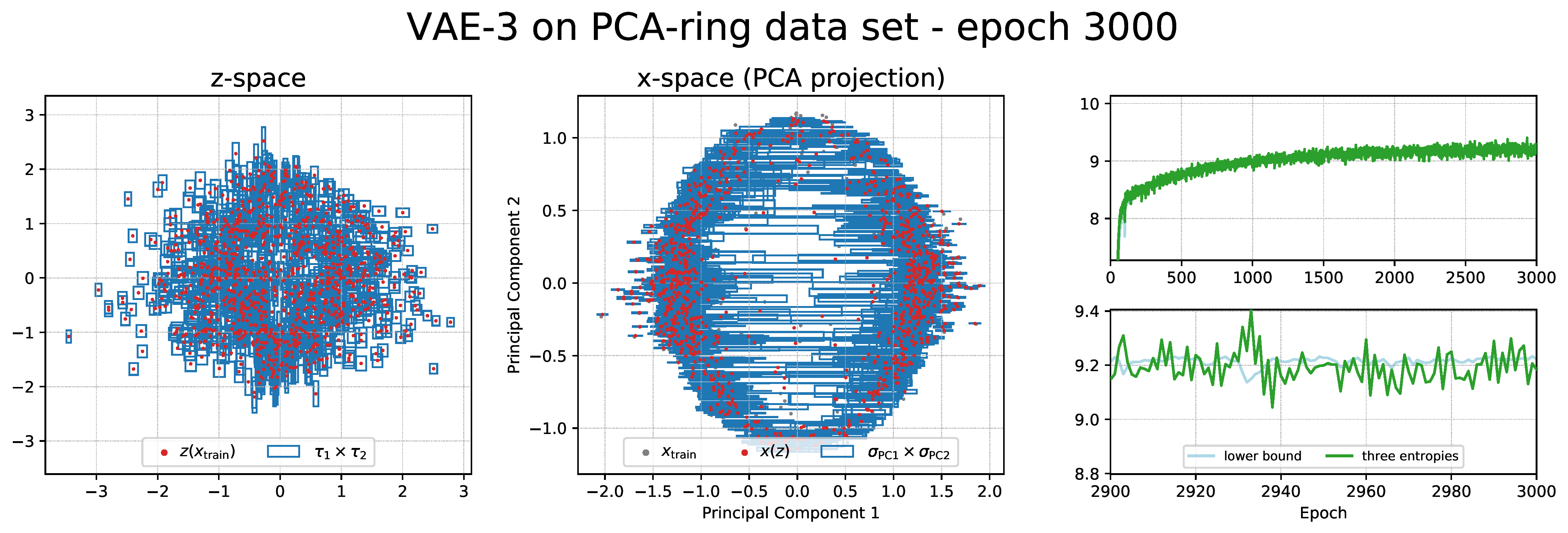}
	\hfill
	
	(c) VAE--3 on PCA-ring data set

\end{adjustbox}

\caption{The training state of the VAEs of \cref{fig:Verification} is shown at four different points during training for each VAE on their respective data sets.
The first plot of each segment shows 1000~$\mathbf{z}$-samples from the encoder in the 2-dimensional z-space with the encoder standard deviation $\tau_h$ displayed as rectangle of width $2\tau_1$ and height $2\tau_2$.
The second plot shows projections of the training data and of the 1000~$\mathbf{x}(\mathbf{z})$-reconstructions of the decoder to the first two PCA components of the training data, as well as the PCA projections of the decoder standard deviation as rectangle of width $2\sigma_\mathrm{PC1}$ and height $2\sigma_\mathrm{PC2}$. The last two plots show the sampled lower bound and the three entropies per data point, with the lower plots showing zoomed-in versions of the upper plots. For (a) the training, held-out and ground-truth log-likelihoods are shown additionally.\vspace{-1.58pt}}
\label{fig:visualization}
\end{figure*}

\end{document}